\documentclass[final,5p,times,twocolumn]{elsarticle}
\usepackage{amsmath}
\usepackage{amssymb} 

\usepackage{tikz}
\usetikzlibrary{arrows,matrix,positioning,patterns}
\usetikzlibrary{calc}
\usepackage{pgfplots} 
\usetikzlibrary{external}
\usetikzlibrary{angles}
\usepackage{hyperref}
\usetikzlibrary{plotmarks}

\newdefinition{rmk}{Remark}
\newproof{pf}{Proof}
\newproof{pot}{Proof of Theorem \ref{thm2}}

\usepackage[numbers]{natbib}
\usepackage[ruled, vlined, linesnumbered]{algorithm2e}

\usepackage{amsthm}

\newtheorem{problem}{Problem}

\usepackage{amsmath}
\usepackage{amssymb} 

\usepackage{eurosym}
\usepackage{colortbl}

\usetikzlibrary{arrows.meta,calc,decorations.markings,math,arrows.meta}

\usepackage[labelformat=simple]{subcaption}
\DeclareCaptionLabelSeparator{periodspace}{.\quad}
\usepackage{color,framed}
\usepackage[utf8]{inputenc} 
\usepackage{tabularx}

\usepackage{rotating}
\usepackage{graphicx}
\usepackage{lscape}
\allowdisplaybreaks 
\usepackage{longtable}

\usepackage[utf8]{inputenc}
\usepackage{pgfplots}
\usepgfplotslibrary{groupplots} 
\pgfplotsset{compat=1.3}

\makeatletter
\def\ps@pprintTitle{%
 \let\@oddhead\@empty
 \let\@evenhead\@empty
 \def\@oddfoot{}%
 \let\@evenfoot\@oddfoot}
\makeatother

\begin{document}

\begin{frontmatter}
%
%
\title{Nominal Evaluation Of Automatic Multi-Sections Control Potential In Comparison To\\ A Simpler One- Or Two-Sections Alternative With Predictive Spray Switching}
\author{Mogens Plessen\corref{cor1}}
\cortext[cor1]{MP is with Findklein GmbH, Switzerland, \texttt{mgplessen@gmail.com}}

%
%
%
%

\begin{abstract}
Automatic Section Control (ASC) is a long-standing trend for spraying in agriculture. It promises to minimise spray overlap areas. The core idea is to (i) switch off spray nozzles on areas that have already been sprayed, and (ii) to dynamically adjust nozzle flow rates along the boom bar that holds the spray nozzles when velocities of boom sections vary during turn maneuvers. ASC is not possible without sensors, in particular for accurate positioning data. Spraying and the movement of modern wide boom bars are highly dynamic processes. In addition, many uncertainty factors have an effect such as cross wind drift, boom height, nozzle clogging in open-field conditions, and so forth. In view of this complexity, the natural question arises if a simpler alternative exist. Therefore, an Automatic Multi-Sections Control method is compared to a proposed simpler one- or two-sections alternative that uses predictive spray switching. The comparison is provided under nominal conditions. Agricultural spraying is intrinsically linked to area coverage path planning and spray switching logic. Combinations of two area coverage path planning and switching logics as well as three sections-setups are compared. The three sections-setups differ by controlling 48 sections, 2 sections or controlling all nozzles uniformly with the same control signal as one single section. Methods are evaluated on 10 diverse real-world field examples, including non-convex field contours, freeform mainfield lanes and multiple obstacle areas. An economic cost analysis is provided to compare the methods. A preferred method is suggested that (i) minimises area coverage pathlength, (ii) offers intermediate overlap, (iii) is suitable for manual driving by following a pre-planned predictive spray switching logic for an area coverage path plan, and (iv) and in contrast to ASC can be implemented sensor-free and therefore at low cost. Surprisingly strong economic arguments are found to not recommend ASC for small farms.
\end{abstract}
\end{frontmatter}


\section{Introduction\label{sec_intro}}

\begin{table}
\centering
\begin{tabular}{|ll|}
\hline
\multicolumn{2}{|c|}{MAIN NOMENCLATURE}\\
\multicolumn{2}{|l|}{Symbols}\\
$A_\text{field}$ & Field size, (ha).\\
$A_\text{total}$ & Total farm size, (ha).\\
$C_\text{chemical}$ & Cost for pesticides, (EUR/$\ell$).\\
$C_\text{water}$ & Cost for water for spray mixture, (EUR/$\ell$).\\
$\Delta K_\text{ASC}$ & Purchase price difference for ASC, (EUR).\\
$L$ & Pathlength for machinery, (m).\\
$N_\text{runs}^\text{field}$ & Number of field runs per crop cycle, (-).\\
$N_\text{years}^\star$ & Number of years until profitability for ASC, (-).\\
$S_\text{field}^\text{ref}$ & Spray volume reference for a given field, ($\ell$).\\
$S$ & Spray volume required for a given field, ($\ell$).\\
$W$ & Machinery working width, (m).\\
$w_\text{nozzle}$ & Inter-nozzle spacing along the boom bar, (m).\\[3pt]
\multicolumn{2}{|l|}{Abbreviations}\\
ASC & Automatic Section Control.\\
\hline
\end{tabular}
\end{table}

Within an agricultural arable farming context, spraying applications play a major role in the process of open-space cereal crop cultivation of grains like wheat, rapeseed, barley and the like. Spraying applications include (i) spraying of herbicides, pesticides and the like for plant protection, but can alternatively (ii) also  refer to spraying of fertilising means, or in general (iii) to applications where input means are sprayed onto a work area through a set of nozzles. Throughout a crop cycle, multiple field runs are required to apply different sprays.  This underlines the importance of an efficient spraying technique for the crop cycle.

Spraying in open-space agriculture is typically realised by a moving machinery, such as ground-based vehicles and more recently also aerial vehicles, which follow an area coverage path plan while spray is simultaneously applied to crops through nozzles. The nozzles are attached to a boom bar aligned along a working width and carried by the machinery with the purpose of applying spray along the working width \cite{portman1979calibrating, smith2000broadcast}.

Typically, \emph{uniform} spray application over the entire field area is desired. Nevertheless, it is mentioned that there exist alternative scenarios such as \emph{spot spraying} or \emph{banding}, where only specific spots or bands of the field area are meant to be sprayed \cite{hafeez2023implementation,plessen2025path,hassen2013effect}. However, the topic for the remainder of this paper is uniform spraying.


Agricultural spraying is intrinsically linked to area coverage path planning. In order to achieve uniform spraying and subject to the constraint of a given working width, an area coverage path plan is desired that avoids area coverage gaps and overlaps, while minimising area coverage pathlength. 

Agricultural spraying is also intrinsically linked to the development of logics for spray switching, i.e., for the on and off switching of nozzles. Besides optimised area coverage path planning, a second strategy for efficient and resource-saving spraying is to suitably switch on and off individual nozzles for areas already sprayed, and to vary nozzle flow rates of individual nozzles along turn maneuvers to compensate for different traveling velocities of different sections of the boom bar. The name for this is \emph{Automatic Section Control} (ASC) or \emph{Automatic Boom Section Control} \cite{luck2010potential}. ASC is a long-standing trend for spray applications in agriculture \cite{han2001modification}.

\begin{figure}
\captionsetup[subfigure]{labelformat=empty}
\centering
  \includegraphics[width=0.99\linewidth]{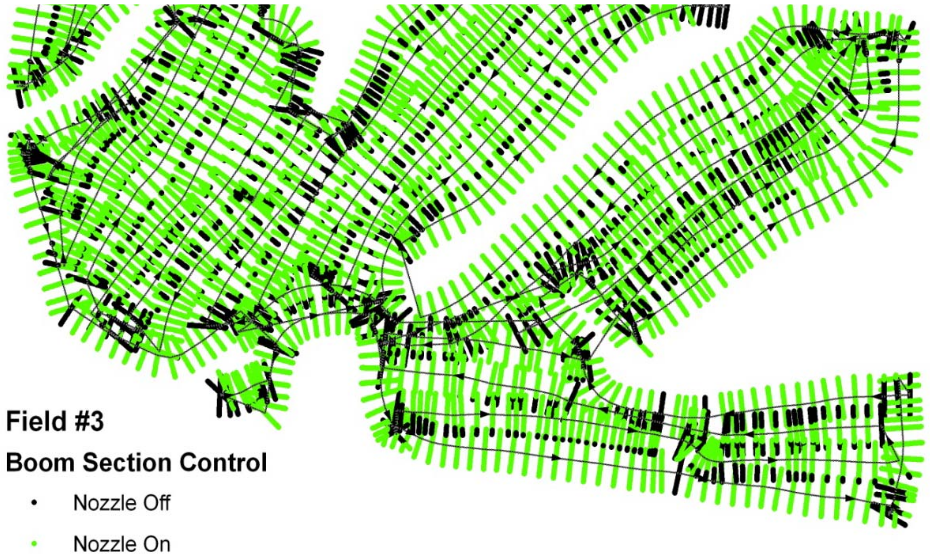}
\caption{Illustration of switching results stemming from a map-based Automatic Section Control.  The figure is from \cite{luck2010generating}. Four comments are made: (i) The gaps between green areas are due to the fact that boom section states (nozzle on- or off-states) are sampled discretely. (ii) Multiple overlapping green areas can still be observed. (iii) The discrete nature of the displayed results (large gaps between two sampled green areas) implies that delays between switching changes (from nozzle on- to an off-state and vice versa) naturally occur. (iv) It can further be observed that adjacent path segments (mainfield lanes) do not have a constant distance to each other. This results in excessive overlapping potential. The figure is given for visualisation of 4 typical difficulties associated with ASC.}
\label{fig_Luck2010}
\end{figure}

The research of this paper is motivated by the observation of the complexity of the spraying process and ASC, especially for open-space agriculture. The spraying process is affected by a remarkable amount of parameters. These include nozzle type, spray fan angle, spray pressure, boom height, nozzle spray overlap, nozzle spacing, nozzle clogging, machinery traveling speed changes, cross wind for spray drift and more \cite{johnson1996sprayer,holterman1997modelling,wolf2009best,hassen2013effect,balsari2017field,
mangus2017analyzing,grisso2019nozzles,cui2019development,burgers2021comparison,
vong2021development,fabula2021nozzle,wang2023evaluation,hussain2025enhancing}. For unmanned aerial spraying \cite{carreno2022numerical}, in contrast to traditional spraying with tractors carrying or trailing spraying machinery and sprayers operating close to the ground, dynamic effects are further amplified. Furthermore, for ASC on top of these factors accurate localisation data and velocity estimates are required along a wide boom bar, whose movement is highly dynamic and also affected by mechanical structure vibrations and inertia. This further adds to the complexity of the spraying process. See also Fig. \ref{fig_Luck2010}.

This raises the natural question: "\emph{In view of the plethora of complexities of the spraying process, is ASC quantitatively worth it and does an alternative exist?}"


To approach this question, this paper (i) focuses on nominal conditions, and (ii) tries to answer first the following question: "\emph{What is the best-case performance an optimal ASC-solution under nominal conditions can achieve in comparison to an alternative practical method?}". Thus, the analysis of this paper searches for an \emph{performance upper bound} of ASC under nominal conditions. Real-world savings potential will be degraded by aforementioned uncertainty factors.

For comparison, an alternative practical method is suggested that varies from a common ASC-solution along 2 independent vectors. First, the \emph{predictive spray switching} method from \cite{plessen2025predictive} is employed. It is here extended to the non-convex general case with non-convex field areas and multiple obstacle areas. This method employs a predictive switching logic tailored to a specific path pattern. This is in contrast to the \emph{reactive} switching logic that is common for ASC, which is based on \emph{Boustrophedon}-based path planning (Ancient Greek for "like the ox turns") with a meandering path pattern. 

\begin{figure*}
\centering
%
%
\begin{subfigure}[t]{.33\linewidth}
  \centering
  \includegraphics[width=.9999\linewidth]{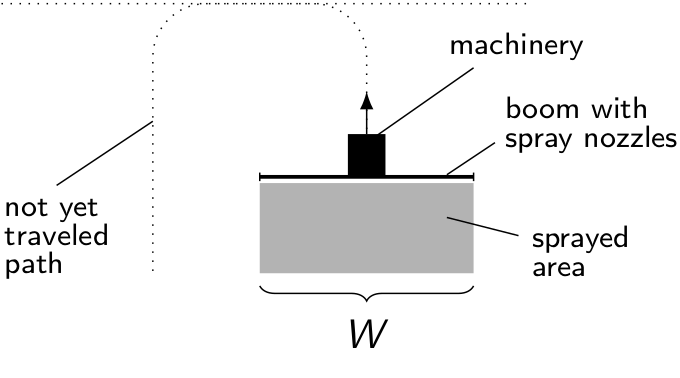}\\[-1pt]
\caption{One section.}
  \label{fig_f1}
\end{subfigure}
\begin{subfigure}[t]{.33\linewidth}
  \centering
  \includegraphics[width=.9999\linewidth]{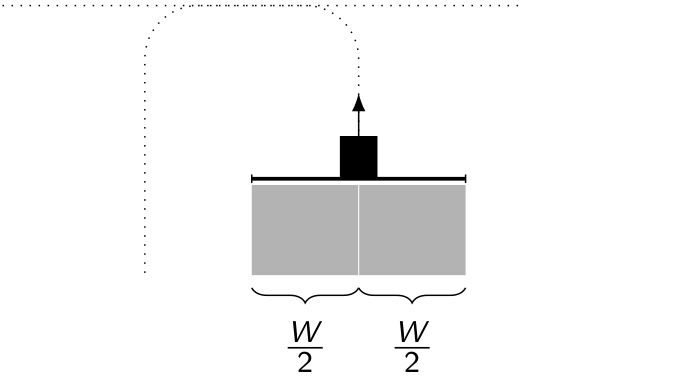}\\[-1pt]
\caption{Two sections.}
  \label{fig_f2}
\end{subfigure}
\begin{subfigure}[t]{.33\linewidth}
  \centering
  \includegraphics[width=.9999\linewidth]{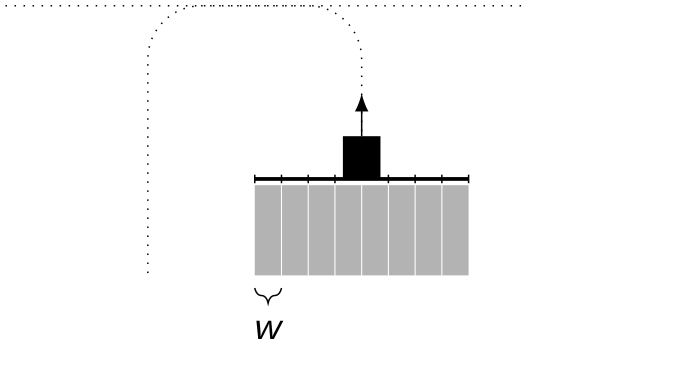}\\[-1pt]
\caption{Multiple sections.}
  \label{fig_f3}
\end{subfigure}
\caption{Distinction between three different sections control setups. The machinery working width is denoted by $W$. For the case of multiple sections, $w$ denotes the section width, which is typically a fraction of the machinery working width. A numerical example is $W=24$m and $w=0.5$m, which results in 48 sections.}
\label{fig_3setups}
\end{figure*}

Second, three different setups for the \emph{number of sections} are considered: 1, 2 or multiple sections. See Fig. \ref{fig_3setups} for illustration. In all three cases always the same number of nozzles is controlled. However, nozzles are controlled either individually or uniformly. For the 1-section setup all nozzles along the boom bar uniformly receive the same switch-on or switch-off signal. In contrast, for the multiple sections-setup different sections can receive different switch-on or switch-off signals.

Thus, in total 6 different experimental setups are compared. Evaluation metrics are (i) pathlength, (ii) spray volume consumption in litres (a proxy for the overlap area, but more accurate since also accounting for nozzle flow rate variations for turn compensation), and (iii) overlap area visualisation. In addition, a detailed economic cost discussion is provided.

Materials and methods, results, discussion and the conclusion are described in Sec. \ref{sec_probformulation}-\ref{sec_conclusion}.

\section{Materials and Methods\label{sec_probformulation}}

This section lists uncertainties associated with the spraying process, before discussing four hierarchical planning levels for agricultural area coverage, and finally discussing two different alternatives for each of the first three planning levels. 

The problem addressed in this paper is as follows:

\begin{problem}\label{problem1}
What is the performance benefit of Automatic Section Control (ASC) under nominal conditions? More specifically, (i) suggest a practical alternative that can also be realised by manual on- and off-switching of at most 2 boom-sections along the boom bar, and (ii) analyse what best-case savings ASC can provide in comparison to the alternative under idealistic nominal conditions.
\end{problem}

\subsection{Complexity of the Spraying Process}

\begin{table}
\centering
\begin{tabular}{|c|l|l|}
\hline 
\rowcolor[gray]{0.8} & & \\[-8pt]
\rowcolor[gray]{0.8} Nr. & Factors affecting spraying results & \multicolumn{1}{c|}{References}  \\[2pt]
\hline
& & \\[-8pt]
1 & Machinery traveling speed changes & \cite{wolf2009best}\\
2 & Varying traveling speeds of different  & \cite{grisso2019nozzles}\\
  & boom sections along the boom bar   & \\
  & during turn maneuvers, see Fig. \ref{fig_complexity1} & \\
3 & Oscillations and inertia of mechanical   & \cite{cui2019development, grisso2019nozzles}\\
 & boom structure during turn maneuvers & \\
4 & Boom height and vertical boom bar  & \cite{burgers2021comparison,cui2019development,balsari2017field}\\
  & oscillations, $H$ in Fig. \ref{fig_complexity2} & \\
5 & Nozzle type selection & \cite{hassen2013effect,balsari2017field,johnson1996sprayer}\\
6 & Nozzle spray overlap of adjacent nozzles & \cite{mangus2017analyzing}\\
7 & Nozzle spacing, $w_\text{nozzle}$ in Fig. \ref{fig_complexity2} & \cite{grisso2019nozzles,hassen2013effect}\\
8 & Nozzle clogging & \cite{vong2021development,grisso2019nozzles}\\ 
9 & Water volume and spray composition & \cite{grisso2019nozzles}\\ 
10 & Spray pressure or nozzle flow rate & \cite{hassen2013effect,wolf2009best,fabula2021nozzle}\\
11 & Spray droplet size & \cite{hassen2013effect,mangus2017analyzing,wolf2009best}\\
12 & Spray fan angle, $\gamma$ in Fig. \ref{fig_complexity2} & \cite{hassen2013effect,mangus2017analyzing}\\
13 & Angled nozzles along the vertical axis   & \cite{grisso2019nozzles}\\
 & ('z-axis') such that adjacent spray    & \\
 &  patterns do not intersect, $\eta$ in Fig. \ref{fig_complexity2} & \\
14 & Non-perpendicular spray angles towards  & \cite{grisso2019nozzles}\\
 & field ground for 3D topography & \\
15 & Section width selection, & \cite{grisso2019nozzles} \\
   & $w_\text{section}$ in Fig. \ref{fig_complexity3} & \\   
16 & Cross wind and spray drift & \cite{holterman1997modelling, wolf2009best, wang2023evaluation}\\
17 & Occupancy grid hyperparameters & \cite{ISO11783}\\
18 & For ASC requirement of at least 1   & \cite{luck2010potential,hussain2025enhancing}\\ 
   & accurate localisation sensor & \\
   & (e.g. RTK-GPS) & \\
19 & Pulse-Width-Modulation (PWM) for & \cite{fabula2021nozzle,han2001modification} \\
   & individual nozzle flow rate control   & \\ 
20 & Reactive spraying (i.e. with delays) unless  & \cite{plessen2025predictive}\\ 
   & spraying is directly coupled to automated  & \\
   & vehicle steering \& velocity control  & \\[2pt]   
\hline
\end{tabular}
\caption{Factors that influence the spraying process. The list is provided to underline the high complexity of the spraying process in open-space agriculture. Variations from a calibrated reference setting for each of the listed factors make an accurate ASC-execution more difficult.}
\label{tab_influences}
\end{table}

\begin{figure}
\captionsetup[subfigure]{labelformat=empty}
\centering
  \includegraphics[width=.6\linewidth]{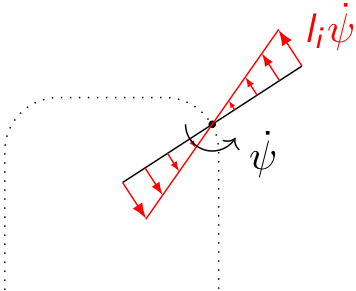}
\caption{Complexities of the spraying process: a sketch illustrating different velocities along different sections of the boom bar during turn maneuvers.}
\label{fig_complexity1}
\end{figure}

\begin{figure}
\captionsetup[subfigure]{labelformat=empty}
\centering
  \includegraphics[width=.99\linewidth]{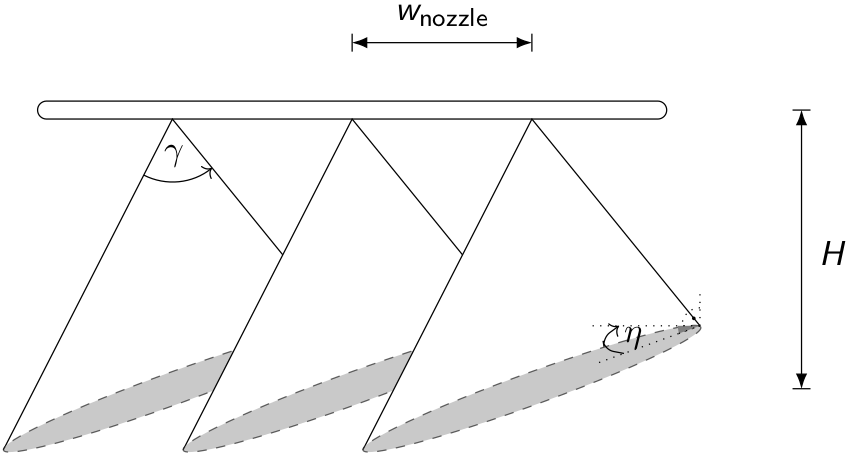}
\caption{Complexities of the spraying process: sketch illustrating (i) boom height $H$, (ii) inter-nozzle spacing $w_\text{nozzle}$, (iii) spray angle $\gamma$, (iv) angle $\eta$ for angled nozzles along the vertical axis ('z-axis') such  that adjacent spray patterns do not intersect, and (v) typically ellipsoid spray patterns for flat-fan nozzles.}
\label{fig_complexity2}
\end{figure}

\begin{figure}
\captionsetup[subfigure]{labelformat=empty}
\centering
  \includegraphics[width=.99\linewidth]{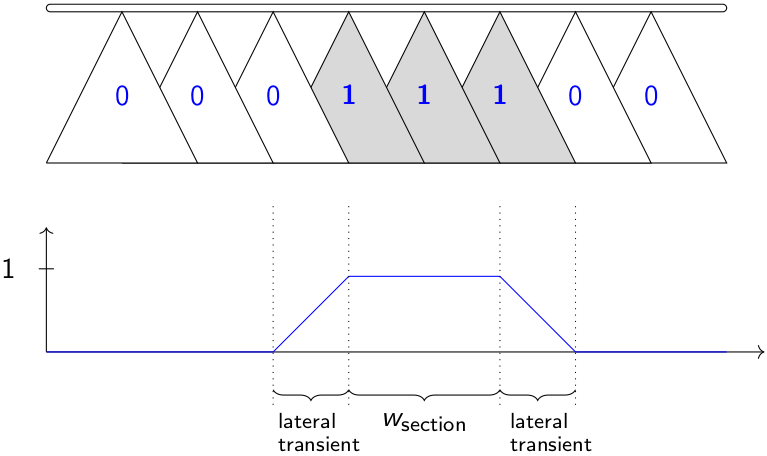}
\caption{Complexities of the spraying process: a sketch illustrating (i) typical overlap of nozzle sprays to attain uniform spray coverage over the boom bar ('broadcast spraying'), and (ii) illustration of lateral transients resulting from the overlap when only subsets of adjacent nozzles ('sections') are actuated.}
\label{fig_complexity3}
\end{figure}

The optimisation of the spraying process is driven by two contradicting objectives: \emph{spraying overlap avoidance} and \emph{spraying gap avoidance}. These contradicting objectives are the root-cause for the complexity of the spraying process.

Table \ref{tab_influences} lists factors that influence uncertainty of the spraying process, particularly in open-space agriculture. A list is provided for clarity and to underline complexity. 
 
For unmanned aerial spraying \cite{carreno2022numerical}, in contrast to traditional spraying with tractors carrying or trailing spraying machinery and sprayers operating close to the ground, dynamic effects are further amplified. Some complexities are further illustrated in Fig. \ref{fig_complexity1}, \ref{fig_complexity2} and \ref{fig_complexity3}.

\subsection{Four Hierarchical Planning Levels\label{subsec_4hierarchicallevels}}

\begin{figure}
\centering%
\begin{tikzpicture}
\draw[fill=white,draw=blue,ultra thick,dotted] (-1,2.25) rectangle (8.25,15.75);
\draw[fill=white] (0,11.5) rectangle (8,15.5);
\node (c) at (4,13.5) [scale=1,color=black,align=left] {
\includegraphics[width=.85\linewidth]{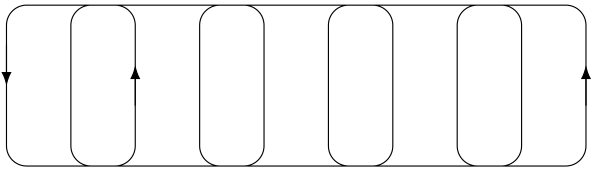}
};
\node (c) at (-0.5,13.5) [scale=1,color=black,align=left] {(i)};
\draw[black,-{Latex[scale=1.0]}] (4, 11.5) -- (4, 11);
\draw[fill=white] (0,7) rectangle (8,11);
\node (c) at (4,9) [scale=1,color=black,align=left] {
\includegraphics[width=.37\linewidth]{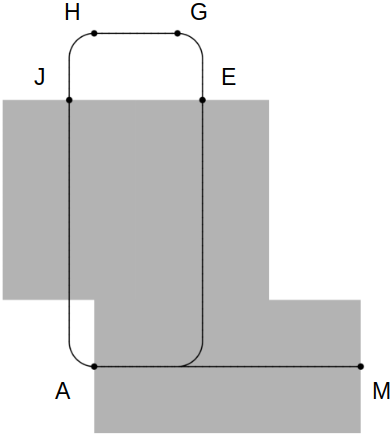}
};
\node (c) at (-0.5,9) [scale=1,color=black,align=left] {(ii)};
\draw[black,-{Latex[scale=1.0]}] (4, 7) -- (4, 6.5);
\draw[fill=white] (0,2.5) rectangle (8,6.5);
\node (c) at (4,4.5) [scale=1,color=black,align=left] {
\includegraphics[width=.65\linewidth]{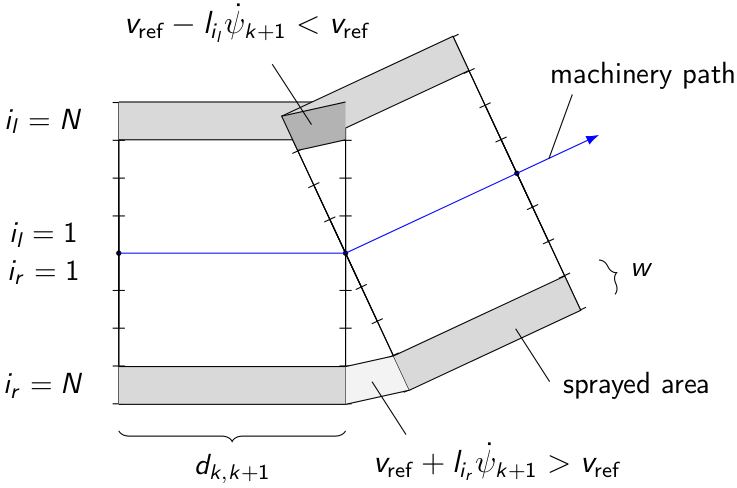}
};
\node (c) at (-0.5,4.5) [scale=1,color=black,align=left] {(iii)};
\draw[black,-{Latex[scale=1.0]}] (4, 2.5) -- (4, 2);
\draw[fill=white] (0,-2) rectangle (8,2);
\node (c) at (4.5,0) [scale=1,color=black,align=left] {
\includegraphics[width=.74\linewidth]{f5.png}
};
\node (c) at (-0.5,0) [scale=1,color=black,align=left] {(iv)};
\end{tikzpicture}
\caption{Distinction between 4 hierarchical planning levels for area coverage: (i) path planning-level, (ii) block switching-level, (iii) sections switching-level, and  (iv) a nozzle control-level (realisation of individual nozzle actuation commands at the embedded machine level). The topic of this paper are the first three levels (i)-(iii).}
\label{fig_4levels}
\end{figure}

\begin{figure}
\centering%
\begin{tikzpicture}
%
\draw[fill=white] (0,13.5) rectangle (8,17.5);
\node (c) at (3.7,15.5) [scale=1,color=black,align=left] {
\includegraphics[width=.72\linewidth]{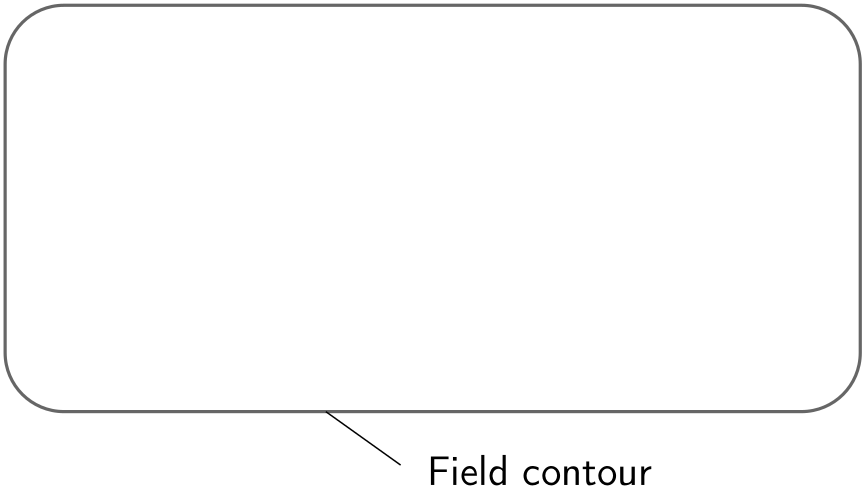}
};
%
\draw[black,-{Latex[scale=1.0]}] (4, 13.5) -- (4, 13);
\draw[fill=white] (0,9) rectangle (8,13);
\node (c) at (3.7,11) [scale=1,color=black,align=left] {
\includegraphics[width=.72\linewidth]{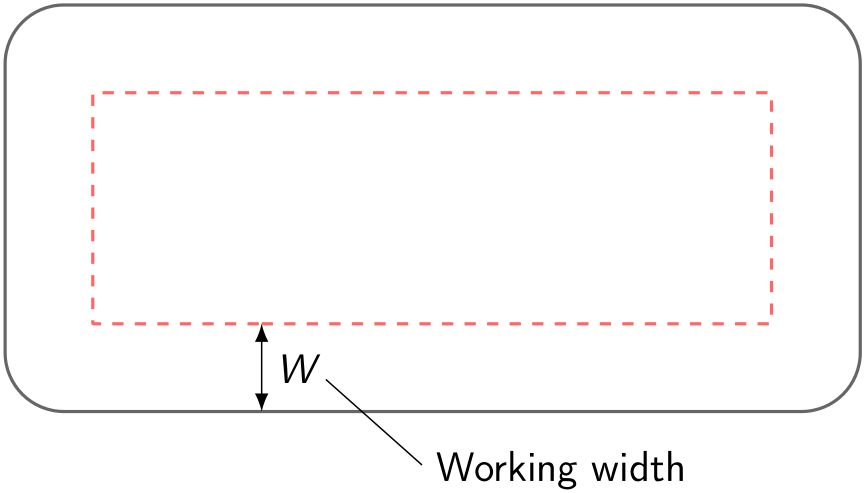}
};
%
\draw[black,-{Latex[scale=1.0]}] (4, 9) -- (4, 8.5);
\draw[fill=white] (0,2.5) rectangle (8,8.5);
\node (c) at (4.1,5.5) [scale=1,color=black,align=left] {
\includegraphics[width=.86\linewidth]{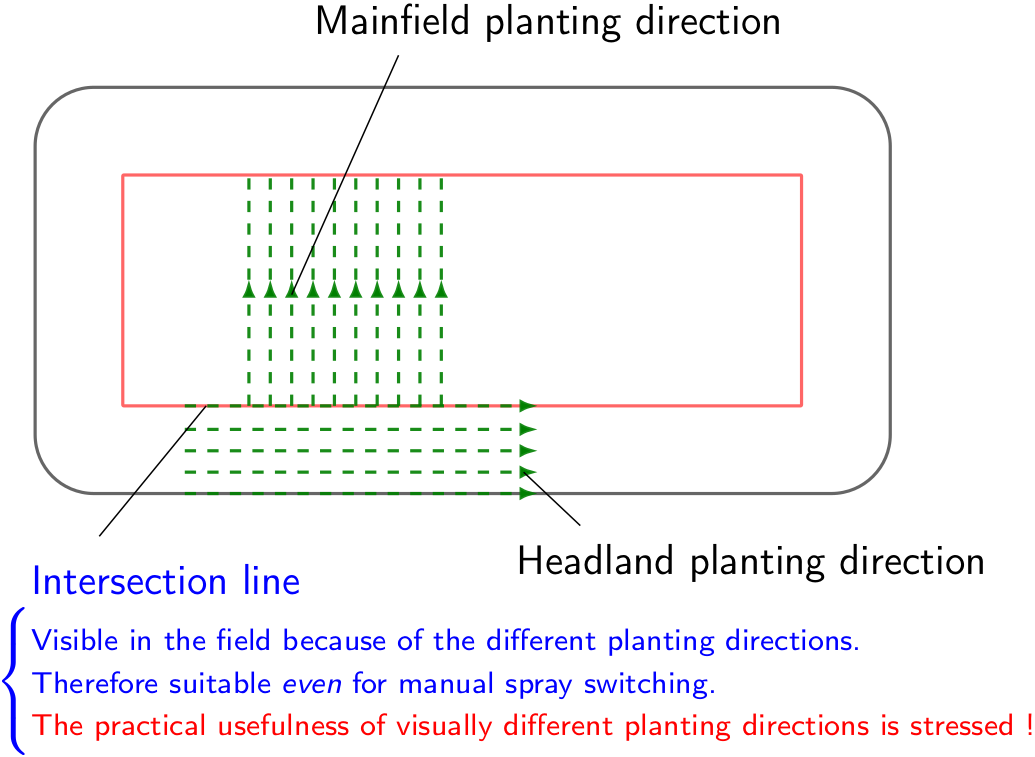}
};
%
%
\draw[black,-{Latex[scale=1.0]}] (4, 2.5) -- (4, 2.1) -- (2, 1.6) -- (2, 1.3);
\draw[black,-{Latex[scale=1.0]}] (4, 2.5) -- (4, 2.1) -- (6, 1.6) -- (6, 1.3);
\draw[fill=white] (0,-2.55) rectangle (3.9,1.3);
\node (c) at (2.0,-0.2) [scale=1,color=black,align=left] {
\includegraphics[width=.425\linewidth]{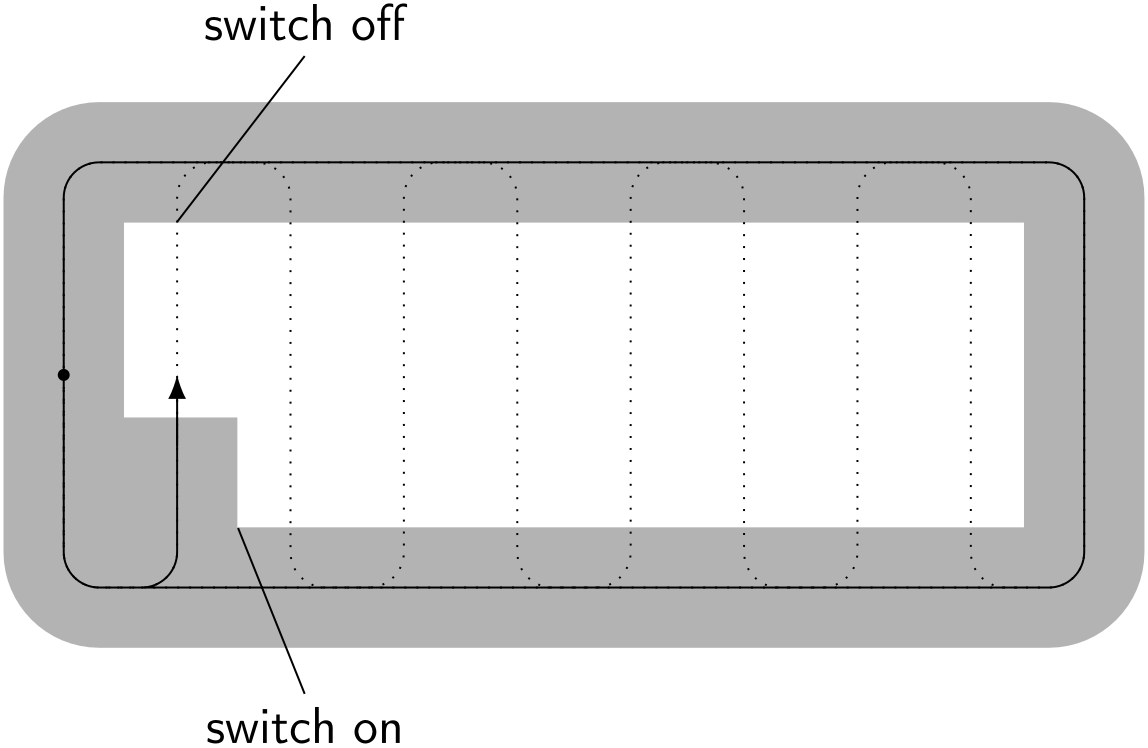}
};
\node (c) at (2,-2.2) [scale=1,color=black,align=left] {Boustrophedon};
\draw[fill=white] (4.1,-2.55) rectangle (8,1.2);
\node (c) at (6,-0.24) [scale=1,color=black,align=left] {
\includegraphics[width=.28\linewidth]{8.png}
};
\node (c) at (6,-2.2) [scale=1,color=black,align=left] {Alternative};
\node (c) at (2,2) [scale=0.9,color=black,align=left] {Method 1};
\node (c) at (6,2) [scale=0.9,color=black,align=left] {Method 2};
\end{tikzpicture}
\caption{Illustration of steps from field contour data to area coverage. The practical usefulness of different planting directions in headland and mainfield area is highlighted. The intersection line between the two areas is useful for triggering spray switching changes (i.e., on- and off-switching) and is useful for both, (i) the \textsf{Boustrophedon}-based and (ii) the \textsf{Alternative} planning method. An extension to the case with obstacle areas (tree islands) inside the field is straightforward by the introduction of obstacle area headland areas with planting areas different from the mainfield area. The headland and mainfield planting directions in above figure are sketched at a $90^\circ$-angle to each other. Any other angle is also possible as long as an intersection line results that can be employed as a visual cue for switching command changes.}
\label{fig_2plantingdirections}
\end{figure}

For area spraying it can be differentiated between 4 hierarchical planning levels: (i) an \emph{area coverage path planning}-level where for a given field area it is distinguished between a mainfield area (see Fig. \ref{fig_2plantingdirections}) and a headland area used for turning, before fitting a lane-grid consisting of a headland path and mainfield lanes, and determining an area covering path based on this lane-grid \cite{tan2021comprehensive,jayalakshmi2025comprehensive,plessen2019optimal}, (ii) a high-level spray switching level here denoted as \emph{block switching}-level, which differentiates between switching logics at the transitions between mainfield lanes and headland path segments \cite{plessen2025predictive}, (iii) a \emph{sections switching}-level which differentiates between switching-states along different sections along the boom bar, and (iv) a \emph{nozzle control}-level which covers all details about individual nozzle control such as e.g. nozzle flow rate at the machine level \cite{fabula2021nozzle}. For illustration see Fig. \ref{fig_4levels}. The focus of this paper is on the first three hierarchical levels (i)-(iii). These three levels will be described in detail in the next three subsections.

Note that the spraying process can be further facilitated by the design of the seeding process, which precedes the spraying process. Two influences are mentioned. First, the orientation of straight mainfield lanes or alternatively freeform mainfield lanes can facilitate turning or reduce the required number of transitions between mainfield lanes and headland path \cite{plessen2021freeform}. Second, the \emph{intersection line} between mainfield area and headland area can be used as visual cue to determine suitable spray on/off-switching locations. See Fig. \ref{fig_2plantingdirections}.

\subsection{Area Coverage Path Planning Level \label{subsec_areacovgpp}}

\begin{figure*}
\centering
\begin{subfigure}[t]{.49\linewidth}
  \centering
\begin{subfigure}[t]{.9999\linewidth}
  \centering
  \includegraphics[width=.89\linewidth]{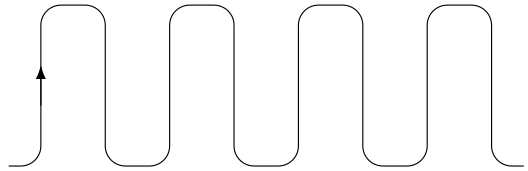}\\[-1pt]
\caption{First path pattern: Boustrophedon.\\[10pt]}
  \label{fig_Boustro1}
\end{subfigure}
\begin{subfigure}[t]{.9999\linewidth}
  \centering
  \includegraphics[width=.89\linewidth]{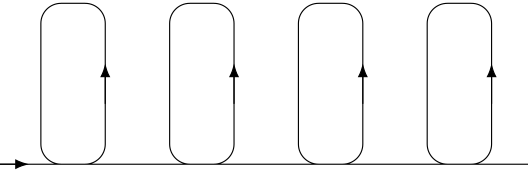}\\[-1pt]
\caption{Second path pattern: Alternative.}
  \label{fig_Alter1}
\end{subfigure}
\end{subfigure}
\begin{subfigure}[t]{.49\linewidth}
  \centering
\begin{subfigure}[t]{.9999\linewidth}
  \centering
  \includegraphics[width=.99\linewidth]{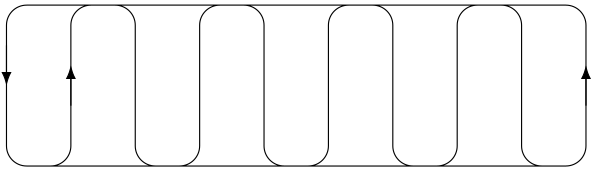}\\[-1pt]
\caption{Boustrophedon path pattern with headland path.\\[10pt]}
  \label{fig_Boustro2}
\end{subfigure}
\begin{subfigure}[t]{.9999\linewidth}
  \centering
  \includegraphics[width=.99\linewidth]{fig_circwheadl.png}\\[-1pt]
\caption{Alternative path pattern with headland path.}
  \label{fig_Alter2}
\end{subfigure}
\end{subfigure}
\caption{First hierarchical level according to Fig. \ref{fig_4levels}: area coverage path planning. Comparison of two path patterns that can be used for area coverage (see \cite{plessen2025predictive}). These two area coverage path planning-setups are evaluated in numerical experiments. }
\label{fig_2pathpatterns}
\end{figure*}

Two different area coverage path planning strategies are compared. For convexly-shaped fields these can be used throughout. For the non-convex cases a graph optimisation problem is solved that adopts the respective strategies whenever possible. 

\begin{enumerate}

\item \textsf{Boustrophedon}: Boustrophedon paths, whose name is derived from Ancient Greek for "like the ox turns", represents the standard pattern for area coverage applications and agricultural real-world praxis. It is characterised by a full headland coverage before mainfield lanes are traversed in a meandering pattern. See Fig. \ref{fig_Boustro1} and \ref{fig_Boustro2}. 

\item \textsf{Alternative}: The method from \cite{plessen2019optimal} is used for area coverage path planning. For convexly shaped field areas the method can be reduced to a recurring path pattern that is discussed in detail in \cite{plessen2018partial} and shown in Fig. \ref{fig_Alter1} and \ref{fig_Alter2}. Headland path edges and headland-to-mainfield lane transitions can be smoothed with the method from \cite{plessen2025smoothing}.

\end{enumerate}

\subsection{Block Switching Level\label{subsec_blockswitchinglevel}}

\begin{figure}
\centering
\begin{subfigure}[t]{.9999\linewidth}
  \centering
  \includegraphics[width=.94\linewidth]{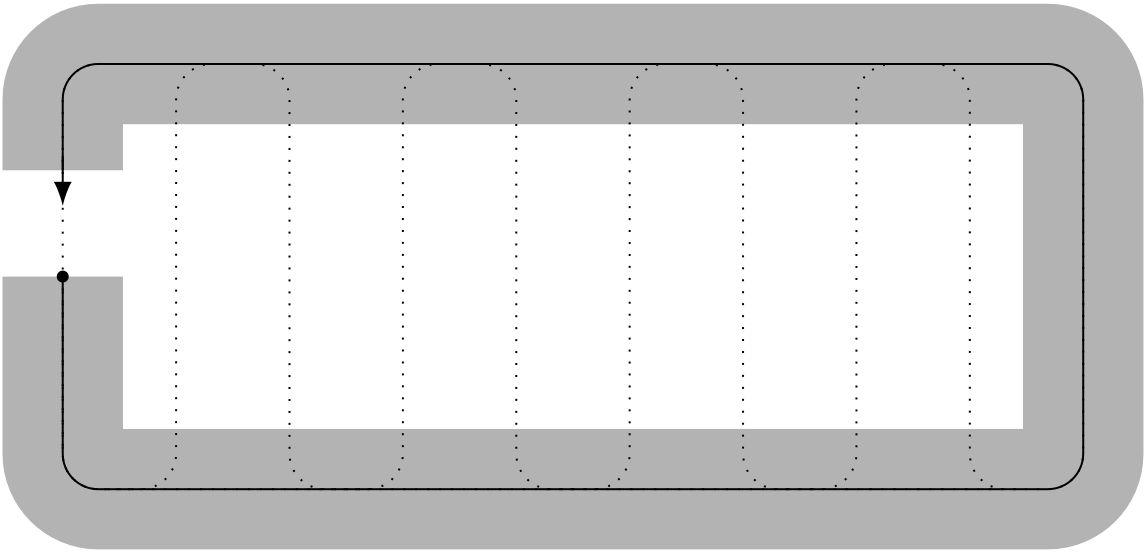}\\[-1pt]
\caption{Initial full headland path coverage.\\[10pt]}
  \label{fig_Boustro_step1}
\end{subfigure}
\begin{subfigure}[t]{.9999\linewidth}
  \centering
  \includegraphics[width=.94\linewidth]{fig_S_p2.png}\\[1pt]
\caption{Reactive switching along mainfield lanes.\\[10pt]}
  \label{fig_Boustro_step2}
\end{subfigure}
\begin{subfigure}[t]{.9999\linewidth}
  \centering
  \includegraphics[width=.94\linewidth]{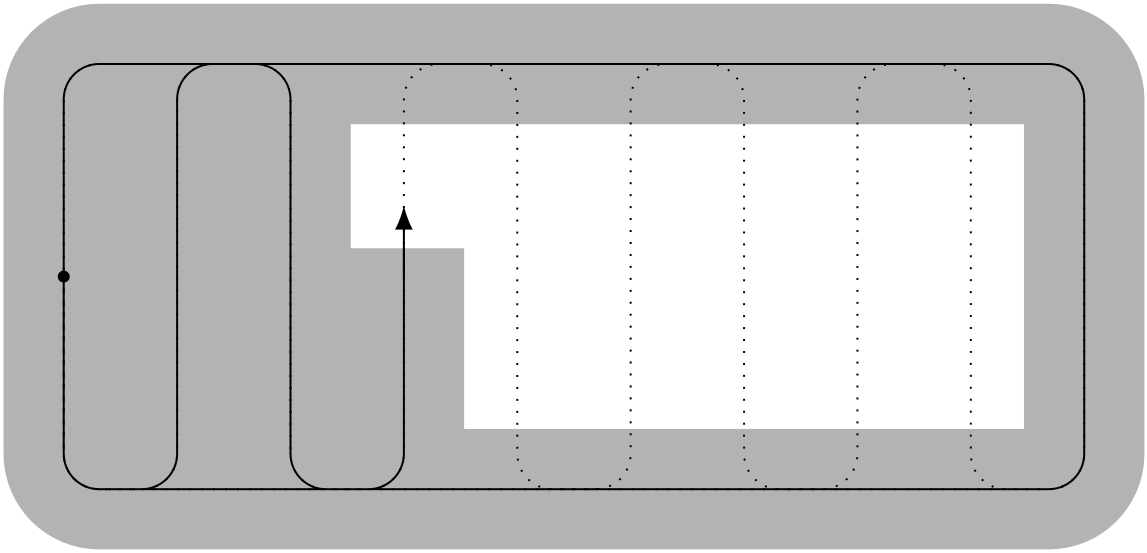}\\[-1pt]
\caption{Reactive switching along mainfield lanes.\\[10pt]}
  \label{fig_Boustro_step3}
\end{subfigure}
\begin{subfigure}[t]{.9999\linewidth}
  \centering
  \includegraphics[width=.94\linewidth]{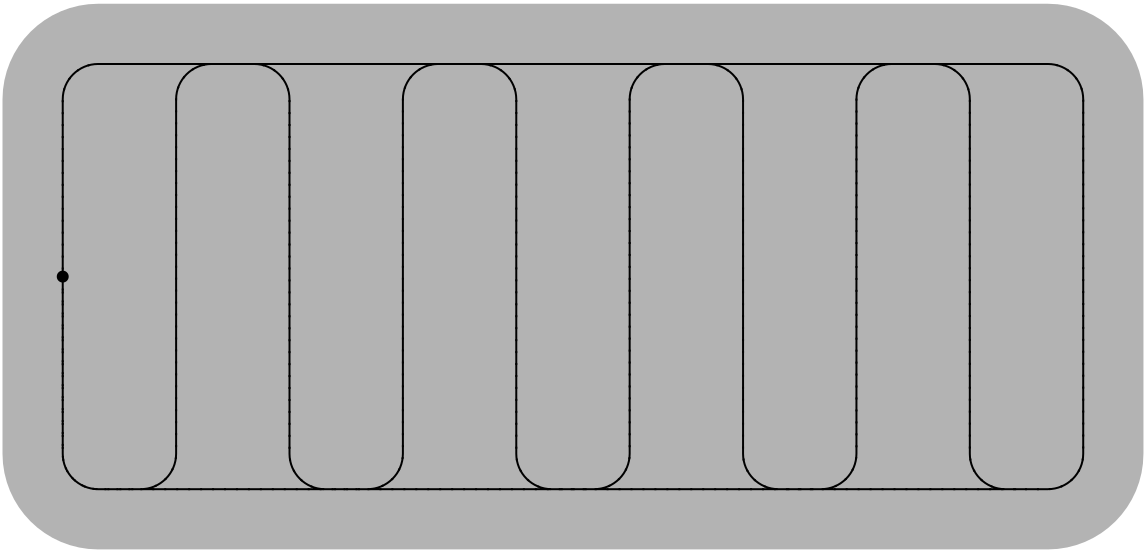}\\[-1pt]
\caption{Result after full traversal of the path plan.}
  \label{fig_Boustro_step4}
\end{subfigure}
\caption{Block Switching Method $\textsf{M}_1$: State-of-the-art reactive switching logic for area coverage based on the Boustrophedon-path pattern in combination with an initial full headland path traversal. The field entrance and starting point of the path is indicated by the black dot.}
\label{fig_Boustro_4steps}
\end{figure}


\begin{figure}
\centering
\begin{subfigure}[t]{.49\linewidth}
  \centering
  \includegraphics[width=.96\linewidth]{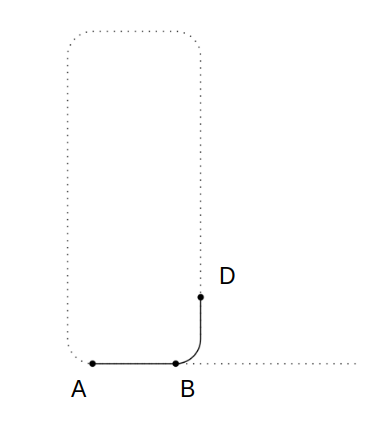}\\[-1pt]
\caption{\\[10pt]}
  \label{fig_step1_switchinglogic}
\end{subfigure}
~\begin{subfigure}[t]{.49\linewidth}
  \centering
\begin{tikzpicture}
\node (c) at (0,0) [scale=1,color=black,align=left] {
\includegraphics[width=.96\linewidth]{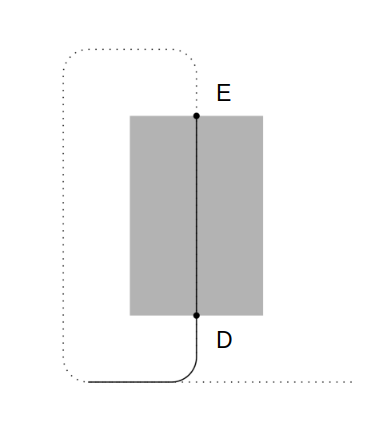}
};
\node (c) at (1.46,-1) [scale=0.9,color=black,align=left] {switch on};
\node (c) at (1.47,1.16) [scale=0.9,color=black,align=left] {switch off};
\end{tikzpicture}
  \caption{\\[10pt]}
  \label{fig_step2_switchinglogic}
\end{subfigure}
\begin{subfigure}[t]{.49\linewidth}
  \centering
  \includegraphics[width=.96\linewidth]{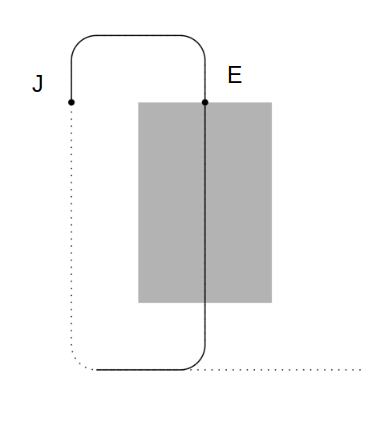}\\[-1pt]
\caption{\\[10pt]}
  \label{fig_step3_switchinglogic}
\end{subfigure}
\begin{subfigure}[t]{.49\linewidth}
  \centering
  \includegraphics[width=.96\linewidth]{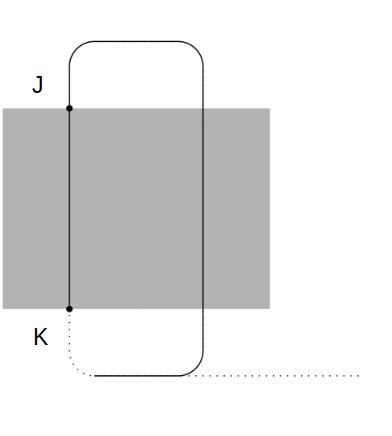}\\[-1pt]
\caption{\\[10pt]}
  \label{fig_step4_switchinglogic}
\end{subfigure}
\begin{subfigure}[t]{.49\linewidth}
  \centering
  \includegraphics[width=.96\linewidth]{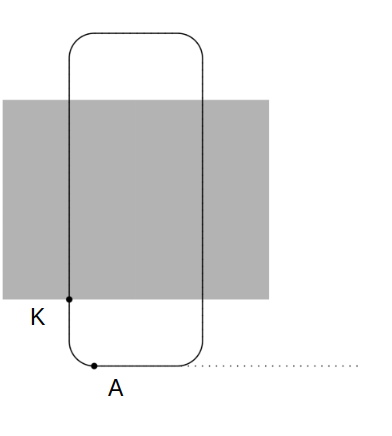}\\[-1pt]
\caption{\\[10pt]}
  \label{fig_step5_switchinglogic}
\end{subfigure}
\begin{subfigure}[t]{.49\linewidth}
  \centering
  \includegraphics[width=.96\linewidth]{8.png}\\[-1pt]
\caption{\\[10pt]}
  \label{fig_step6_switchinglogic}
\end{subfigure}
\caption{Block Switching Method $\textsf{M}_2$: Illustration of the predictive switching logic (a)-(f) according to the method from \cite{plessen2025predictive}. The switching logic exploits the structure of a specific path pattern. The concatenation of this path pattern and the concatenation of corresponding switching logics permits area coverage. Gray areas indicate sprayed area. Along transitions A-D, E-J and K-A it is switched off. Along transitions D-E, J-K and A-M it is switched on.}
\label{fig_6steps_switchinglogic}
\end{figure}

\begin{figure}
\captionsetup[subfigure]{labelformat=empty}
\centering
  \includegraphics[width=.82\linewidth]{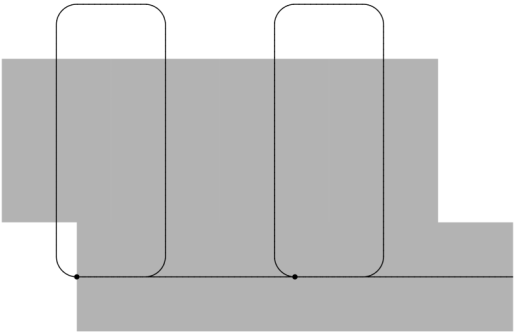}
\caption{Block Switching Method $\textsf{M}_2$: Concatenation of two path planning patterns. This figure illustrates how the concatenation of multiple path patterns enables spraying of a larger area \cite{plessen2025predictive}.}
\label{fig_concat2patterns}
\end{figure}

\begin{figure}
\centering
\begin{subfigure}[t]{.9999\linewidth}
  \centering
  \includegraphics[width=.99\linewidth]{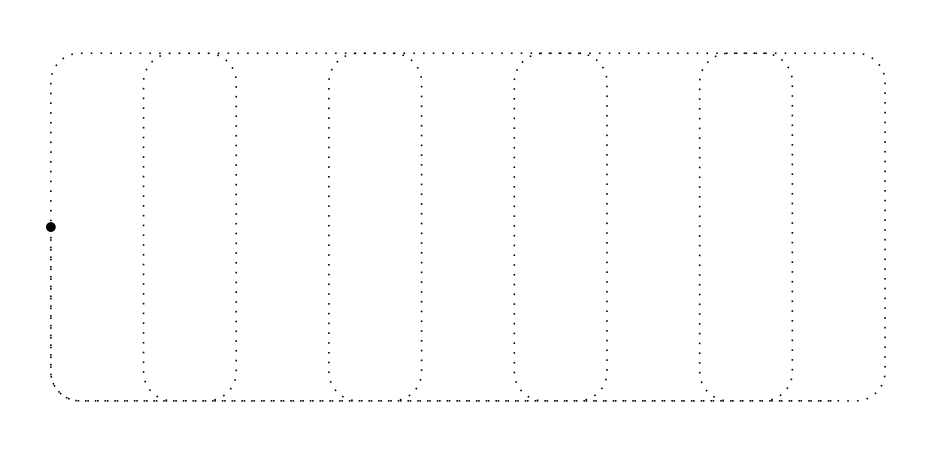}\\[-1pt]
\caption{Planned area coverage path.\\[10pt]}
  \label{fig_prob1}
\end{subfigure}
\begin{subfigure}[t]{.9999\linewidth}
  \centering
  \includegraphics[width=.99\linewidth]{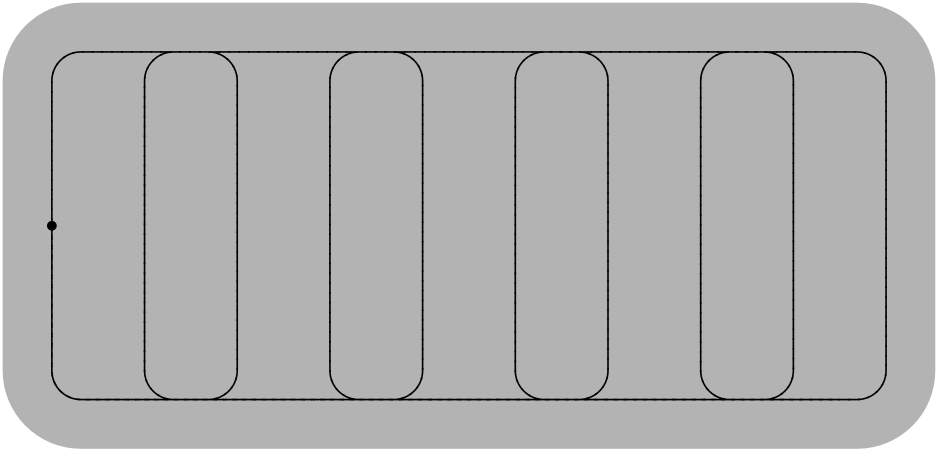}\\[-1pt]
\caption{Spraying result after traversal of the planned area coverage path and application of proposed switching logic.}
  \label{fig_prob2}
\end{subfigure}
\caption{Area Coverage Path Planning Method $\textsf{M}_2$: Visualisation of a full area coverage example resulting from the concatenation of multiple path patterns, and the application of proposed switching logic in Fig. \ref{fig_6steps_switchinglogic} for each path pattern.}
\label{fig_fullareacovg}
\end{figure}

According to the hierarchical structure of Fig. \ref{fig_4levels}, a block swiching logic is applied \emph{on top} of a given area coverage path plan according to the methods from the previous Sec. \ref{subsec_areacovgpp}. 

\begin{enumerate} 

\item \textsf{Boustrophedon}: The associated switching logic \emph{on top} of this area coverage path planning method is illustrated in Fig. \ref{fig_Boustro_4steps}.

\item \textsf{Alternative}: The associated switching logic is summarised in Fig. \ref{fig_6steps_switchinglogic}, \ref{fig_concat2patterns} and \ref{fig_fullareacovg}. The logic is introduced in detail in \cite{plessen2025predictive}.

\end{enumerate}

In the following, the \textsf{Boustrophedon}-based and \textsf{Alternative} area coverage path planning methods in combination with their corresponding block switching logic shall be abbreviated by method $\textsf{M}_1$ and $\textsf{M}_2$, respectively.

\subsection{Sections Switching Level\label{subsec_sectionsswitchinglevel}}

\begin{figure}
\captionsetup[subfigure]{labelformat=empty}
\centering
  \includegraphics[width=.99\linewidth]{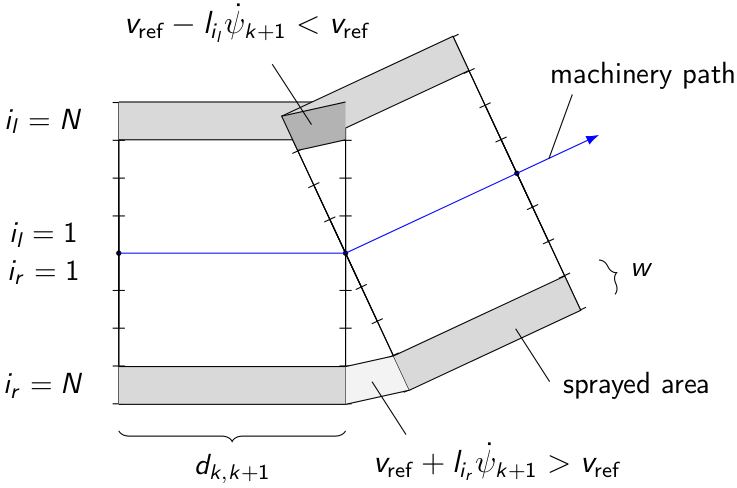}
\caption{Third hierarchical planning level according to Fig. \ref{fig_4levels}: sections switching-level. Illustration of the discretisation method used for evaluation in numerical experiments.  }
\label{fig_sectionsswitchinglevel}
\end{figure}

According to the hierarchical structure of Fig. \ref{fig_4levels}, on the sections switching level (i) a decision regarding section width, and (ii) regarding discretisation modeling has to be made. There are 2 factors that influence our discretisation modeling of spray section areas at each sampling distance along a path.

The 2 influencing factors are: (i) Spray patterns that result from the dominant standard \emph{flat-fan} nozzle type are ellipsoid \cite{grisso2019nozzles}. (ii) The simplifying assumption is made that the boom bar is orthogonally oriented along the area coverage path plan at every sampling point. See Fig \ref{fig_complexity2} for visualisation of ellipsoidal spray patterns. See Fig. \ref{fig_sectionsswitchinglevel} for illustration of assumption (ii).

For the remainder of this paper the discretisation method in Fig. \ref{fig_sectionsswitchinglevel} is employed. It is distinguished between two types of polygons that each consists of 4 points. The first polygon type has an area of $A_{k,k+1,}=d_{k,k+1}w$, and is constructed by two sampling spaces along the machinery area coverage path and a given section width $w>0$. 

The second polygon type is constructed at heading changes along the area coverage path and constructed by connecting boundary points of two polygons at the previous and next sampling space. At those heading changes the traveling velocity along different sections of the boom bar varies according to
\begin{equation}
v_{i}=v_\text{ref} + l_{i}\dot{\psi}_{k+1},~\forall i\in\{i_r,i_l\},~\forall i_r,i_l\in\{1,\dots,N\},\label{eq_vi}
\end{equation} 
where $N$ different sections are assumed both at the left and right hand-side of the machinery path; see Fig. \ref{fig_sectionsswitchinglevel} for illustration. Depending on a left-turn or a right-turn the traveling velocity of a boom section may be larger or smaller than the machinery reference velocity $v_\text{ref}$. Negative $v_{i}<0$ are possible, which imply 'backwards motion' of section $i$ that can occur during turn maneuvers and at low reference velocity $v_\text{ref}\approx 0$.

Spray volumes are assumed to be uniform over the polygon areas for this discretisation scheme.

A section area width $w$ has to be selected. This is not a straightforward choice since lateral transients result from spray overlap (see Fig. \ref{fig_complexity3}). The smallest possible selection is
\begin{equation}
w=w_\text{nozzle},
\end{equation} 
where $w_\text{nozzle}$ indicates the inter-nozzle spacing as illustrated in Fig. \ref{fig_complexity2}. For numerical experiments below this setting is used since it offers the finest discretisation spacing possible. 

Nevertheless, the simplification of assuming constant spray along this section despite an actual lateral transient according to Fig. \ref{fig_complexity3} is explicitly stressed. In Fig. \ref{fig_complexity3}, three nozzles are illustratively treated as one section of width $w_\text{section}$ with 100\% overlap, where the lateral spray distance on the ground is twice the nozzle spacing \cite{johnson1996sprayer}.

We assume the following linear relationship \cite{johnson1996sprayer,grisso2019nozzles} between controllable section flow rates, $f_\text{section}^i$ ($\ell$/s) for $N$ sections $i\in\{1,\dots,N\}$ on both the left- and right-hand side of the machinery path according to Fig. \ref{fig_complexity2}, and actual spray volume $s_\text{volume}^{\text{actual},i}$ ($\ell$/ha),
\begin{equation}
s_\text{volume}^{\text{actual},i} = \frac{f_\text{section}^i}{v_i w},~\forall i\in\{i_r,i_l\},~\forall i_r,i_l\in\{1,\dots,N\},\label{eq_svolactuali}
\end{equation}
where $w$ denotes the section width and $v_i\in\mathbb{R}$ denotes the traveling  speed of the centroid of section $i$ according to \eqref{eq_vi}. Throughout visualisations in this paper and in line with Fig. \ref{fig_sectionsswitchinglevel} and \eqref{eq_svolactuali}, smaller and larger $s_\text{volume}^{\text{actual},i}$ are differentiated by brighter and darker gray colors, respectively.

Ideally, for uniform spraying the applied spray volume is constant for all non-overlapping section areas with
\begin{equation}
s_\text{volume}^{\text{actual},i} = s_\text{volume}^{\text{ref}},~\forall i\in\{i_r,i_l\},~\forall i_r,i_l\in\{1,\dots,N\},\label{eq_svolref}
\end{equation}
where $s_\text{volume}^{\text{ref}}>0$ denotes a desired reference spray volume. Reformulating \eqref{eq_svolactuali} with reference \eqref{eq_svolref} one obtains a nominal nozzle flow rate control signal,
\begin{equation}
f_\text{section}^i  = s_\text{volume}^{\text{ref}} v_i w,~\forall i\in\{i_r,i_l\},~\forall i_r,i_l\in\{1,\dots,N\}.\label{eq_fsectioni}
\end{equation}
This nominal signal, which implies that higher flow rates are required for faster section velocities to maintain a constant reference spray volume, then serves as reference to low level controllers that actuate Pulse-Width-Modulated (PWM) solenoids to control the nozzle flow rate \cite{fabula2021nozzle,han2001modification}.

Starting from basis equation \eqref{eq_fsectioni}, multiple customisations and refinements are possible. First, one may switch off for negative $v_i < 0$, which implies ’backwards motion’, which may occur during turn maneuvers and low reference velocity $v_\text{ref} \approx 0$. Thus, $f_\text{section}^i=0$ if $v_i < 0$.

\begin{figure*}
\centering
\begin{subfigure}[t]{.245\linewidth}
  \centering
  \includegraphics[width=.9999\linewidth]{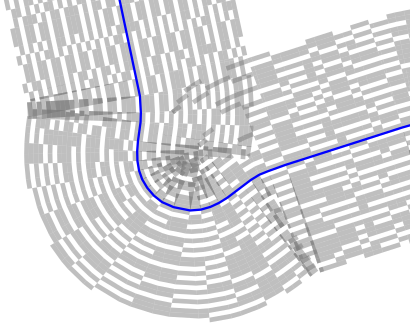}\\[-1pt]
\caption{Occupancy grid: $d_\text{G}=0.5$m.}
  \label{fig_occup1}
\end{subfigure}
\begin{subfigure}[t]{.245\linewidth}
  \centering
  \includegraphics[width=.9999\linewidth]{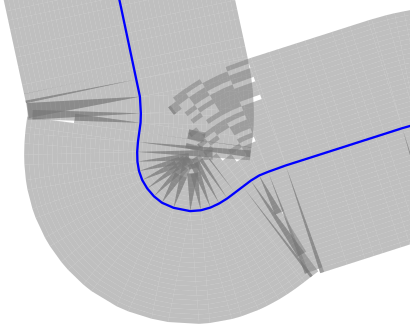}\\[-1pt]
\caption{Occupancy grid: $d_\text{G}=0.25$m.}
  \label{fig_occup2}
\end{subfigure}
\begin{subfigure}[t]{.245\linewidth}
  \centering
  \includegraphics[width=.9999\linewidth]{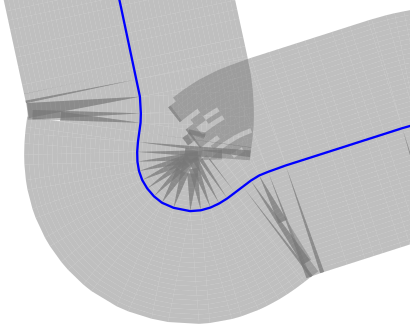}\\[-1pt]
\caption{Occupancy grid: $d_\text{G}=0.125$m.}
  \label{fig_occup3}
\end{subfigure}
\begin{subfigure}[t]{.245\linewidth}
  \centering
  \includegraphics[width=.9999\linewidth]{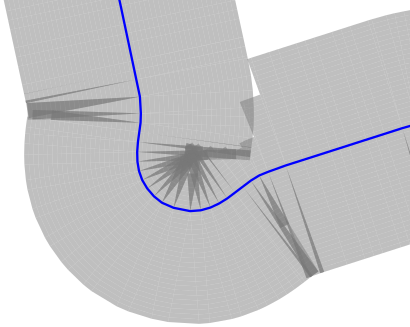}\\[-1pt]
\caption{Polygon-approach (ours).}
  \label{fig_occup4}
\end{subfigure}
\caption{The machinery path to be followed from the top-left to the right is in blue. Sprayed area is in gray. Illustration of the disadvantage of an \emph{occupancy grid}-based approach to determine if a small area has already been sprayed before or not (see also Fig. \ref{fig_Luck2010}). The effect of different hyperparameter choices for $d_\text{G}$, a threshold that determines if the centroid of a small area candidate is close to the centroid of an already sprayed area via the Euclidean norm, is visualised. In (a) there are gaps if $d_\text{G}$ is too large. In (b)-(c) there are overlaps if $d_\text{G}$ is too small. In contrast, in Fig. \ref{fig_occup4} an alternative non-occupancy grid based approach is visualised. This is approach does not require a threshold hyperparameter and yields both less gaps as well as less overlapping.}
\label{fig_occupgrid}
\end{figure*}

Second, one may switch off, $f_\text{section}^i=0$, if there is an overlap with an already sprayed area or with an area that will be sprayed later within an predictive spraying scheme. The standard approach for this is an \emph{occupancy grid}-based approach \cite{ISO11783, luck2010generating}. The advantage of this approach is computational simplicity. The disadvantage is that at least 1 hyperparameter needs to be set, which results in either spray gaps or spray overlaps, as Fig. \ref{fig_occupgrid} demonstrates. Therefore, an alternative \emph{polygon-based} approach is employed throughout the remainder of this paper: Here, (i) a polygon is recursively constructed from the area that has been sprayed so far, before (ii) at area coverage runtime for a potential spray section cell candidate it is evaluated whether it is element of the polygon representing already sprayed area or not. If the centroid of a cell candidate is not yet element of the polygon the cell is sprayed and the polygon is augmented. For computational efficiency and to keep a constant size, the polygon is recursively updated by considering only the last 10 sampling spaces. This polygon-approach is only applied along the headland path area. 

Third, multiple nozzles may be uniformly controlled. For example, for the 1-sections case of Fig. \ref{fig_3setups}, one may set $f_\text{section}^i  = s_\text{volume}^\text{ref}v_\text{ref}w,~\forall i\in\{i_r,i_l\},~\forall i_r,i_l\in\{1,\dots,N\}$.

Fourth and as a detail, for nominal numerical experiments in Sec. \ref{sec_results}, equation \eqref{eq_fsectioni} is recalculated at each sampling space along the area coverage path.  No rate constraints on $f_\text{section}^i$ are considered. This assumption is in line with the nominal experimental setup and best-case evaluation. Rate-constraints are regarded as PWM-associated uncertainties in Table \ref{tab_influences}.

\begin{figure*}
\centering
\begin{subfigure}[t]{.33\linewidth}
  \centering
  \includegraphics[width=.9999\linewidth]{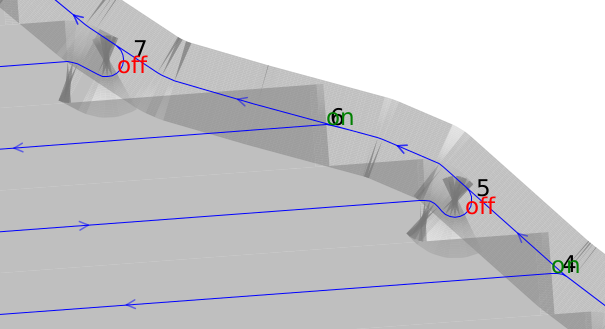}\\[-1pt]
\caption{One section.}
  \label{fig_schrag1}
\end{subfigure}
\begin{subfigure}[t]{.33\linewidth}
  \centering
  \includegraphics[width=.9999\linewidth]{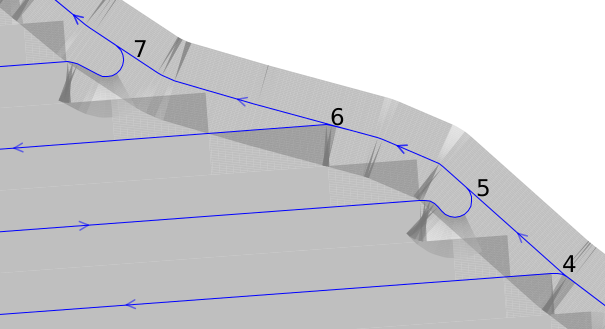}\\[-1pt]
\caption{Two sections}
  \label{fig_schrag2}
\end{subfigure}
\begin{subfigure}[t]{.33\linewidth}
  \centering
  \includegraphics[width=.9999\linewidth]{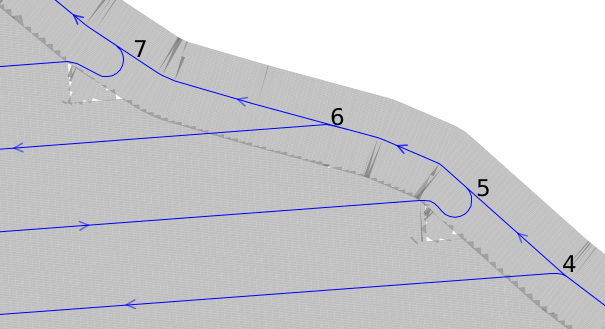}\\[-1pt]
\caption{Multiple sections.}
  \label{fig_schrag3}
\end{subfigure}
\caption{The effect of mainfield lane endings not being rectangular with respect to the headland path is visualised for the 3 different section-setups according to Fig. \ref{fig_3setups}. See in particular the differences around the transition annotated with index number 6: In Fig. \ref{fig_schrag1} all nozzles are simultaneously switched on or off. In contrast, in Fig. \ref{fig_schrag3} each nozzle is switched individually.}
\label{fig_schrag}
\end{figure*}

\subsection{Strategy 1: Automatic Section Control}

The objective of ASC is to minimise spray overlap areas. The core idea is to (i) switch off spray nozzles on areas that have already been sprayed, and (ii) to dynamically adjust nozzle
flow rates along the boom bar that holds the spray nozzles when velocities of boom sections vary during turn maneuvers. Several comments are made. 

First, in contrast to an occupancy grid approach that was mentioned in Sec. \ref{subsec_sectionsswitchinglevel} and that grids a field area into cells (i.e. discretised sub-areas of the field), here a different approach is taken. This is done to generate high-precision results and to avoid having to select discretisation-hyperparameters (see Fig. \ref{fig_occupgrid}). Here the approach is used that (i) recursively extends a polygon that is constructed from the sprayed area so far, before (ii) evaluating at area coverage runtime if a potential spray section area is element of the polygon area or not.

Second, \emph{inter-mainfield lane} priorities are assigned based on the accumulated heading changes throughout the traversal of each mainfield lane. Mainfield lanes with less accumulated heading changes are preferred since these permit simpler spraying. In cases of larger machinery turning radii and smaller operating widths, scenarios may in general occur at transitions between mainfield lanes and headland paths where sprayed areas associated with two adjacent mainfield lane traversals potentially overlap. In such scenarios the mainfield lane with less accumulated heading changes is granted priority for spraying.

Finally, two additional observations are made: (i) ASC must run automatedly since a large number of individually controlled nozzles (e.g. 48 nozzles for a 24m wide boom bar and 0.5m nozzle spacing) cannot be efficiently switched manually by a human operator at typical fast machinery traveling speeds. (ii) At least one localisation sensor (e.g. for accurate RTK-GPS) is required to enable ASC-automation. Ideally, additional sensors are available to as accurately as possible measure the traveling velocity of individual nozzles (instead of estimating them from geometrical calculations subject to uncertainties such as mechanical oscillations that may occur for a wide boom bar during turn maneuvers). Thus, in general ASC is highly dependent on accurate sensor measurements and automation.

\subsection{Strategy 2: GPS-free, Maximal 2 Sections, Visual Cues\label{subsec_strategy2}}

In order to develop a strategy that is simpler than ASC and that may also be implemented manually, two considerations are made. First, a maximum of 1 or 2 sections are considered; see Fig. \ref{fig_3setups}. Note that even if only 1 or 2 sections are used for the spray logic this does not alter the number of nozzles along the boom bar. Multiple nozzles (either all or half) simply follow the same control signal at a given sampling time. Second, to avoid the need for GPS-sensors for localisation a visual cues-approach is taken. Therefore, it is differentiated between headland and mainfield \emph{planting directions}. Then, the visual difference of these two directions in the field (an 'intersection line' of the two areas) can be used to trigger spray switching changes, i.e., to trigger on- and off-switching only when all sections of the boom bar have crossed the intersection line. See Fig. \ref{fig_2plantingdirections} for illustration. Note that this setup can be used as basis for both the \textsf{Boustrophedon}-based and the \textsf{Alternative} area coverage path planning method. So-called \emph{pre-emergence markers} (marker discs that are trailed in the earth), which represent a mechanical solution and can be employed sensor-free, can help to build the intersection line during the seeding process and to maintain a constant distance in-between mainfield lanes.

See Fig. \ref{fig_schrag} for the effect of employing only 1 or 2 sections in comparison to a multi-sections approach.

\section{Results\label{sec_results}}

\subsection{Comparison of 6 Experimental Setups}

To address Problem \ref{problem1} the two area coverage path planning strategies from Sec. \ref{subsec_areacovgpp}, $\textsf{M}_1$ and $\textsf{M}_2$, are compared with 3 different sections control setups in Fig. \ref{fig_3setups}. Thus, overall 6 different setups are compared. The corresponding spray volumes are denoted $S_{\textsf{M}_k}^{j},~\forall k\in\{1,2\},~\forall j\in\{1,2,48\}$. The area coverage pathlengths for $\textsf{M}_1$ and $\textsf{M}_2$ shall be denoted by $L_{\textsf{M}_1}$ and $L_{\textsf{M}_2}$. Note that the pathlengths are the same for all 3 sections setups.

\subsection{10 Nominal Real-World Field Examples}

\begin{figure*}
\centering
\begin{subfigure}[t]{.195\linewidth}
  \centering
  \includegraphics[width=.99\linewidth]{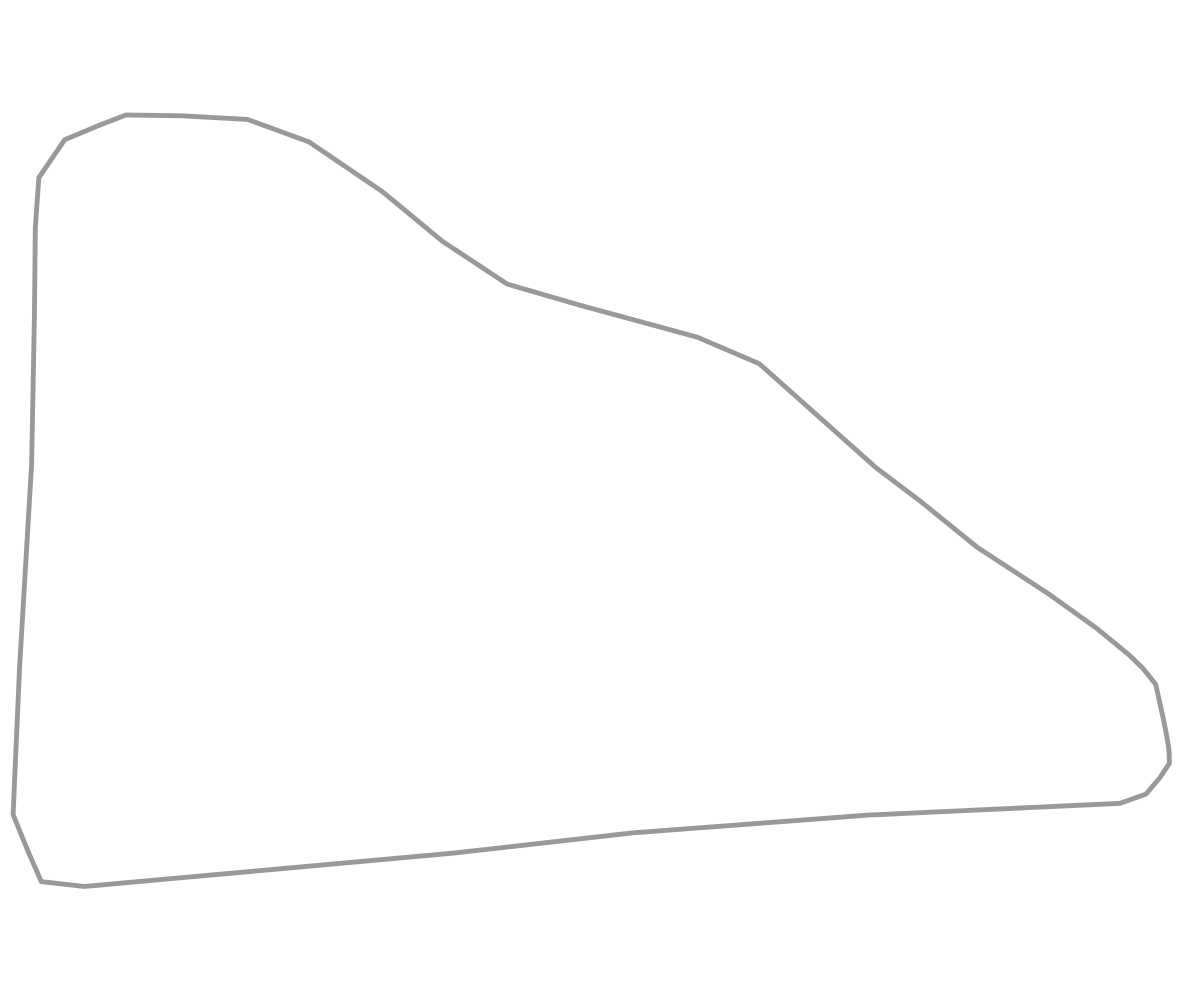}\\[-1pt]
  \label{fig_ef14}
\end{subfigure}
\begin{subfigure}[t]{.195\linewidth}
  \centering
  \includegraphics[width=.99\linewidth]{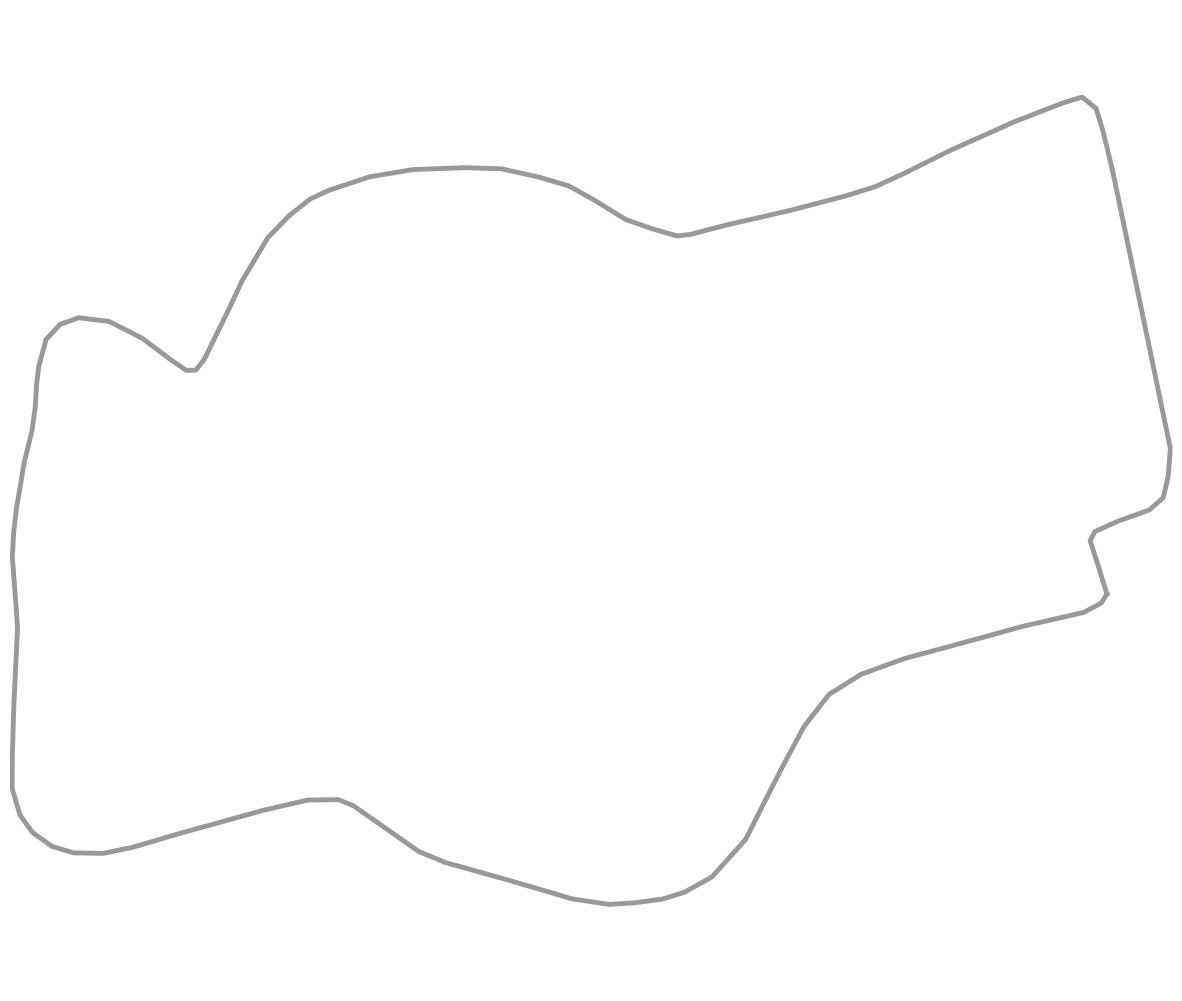}\\[-1pt]
  \label{fig_ef17}
\end{subfigure}
\begin{subfigure}[t]{.195\linewidth}
  \centering
  \includegraphics[width=.99\linewidth]{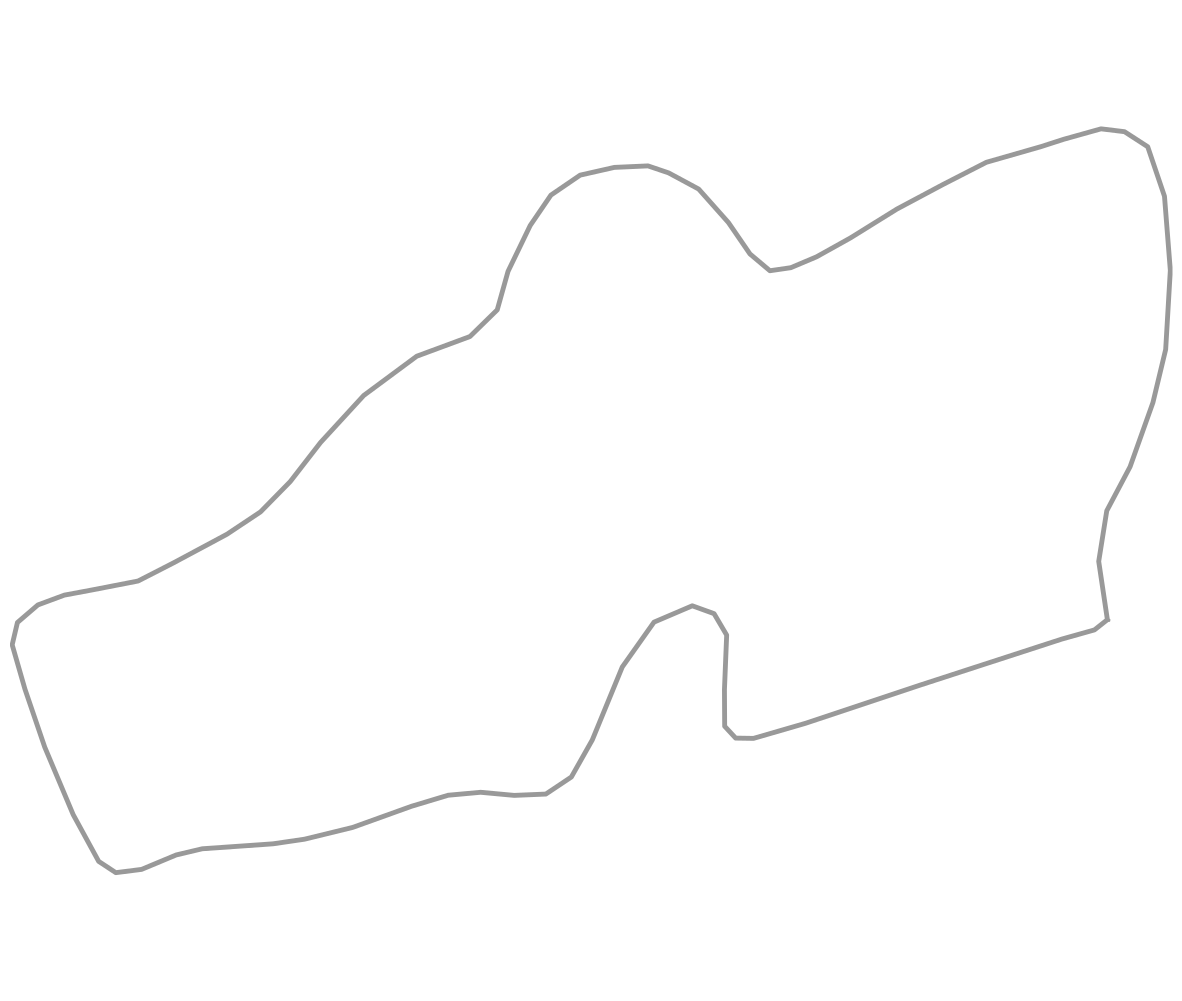}\\[-1pt]
  \label{fig_ef16}
\end{subfigure}
\begin{subfigure}[t]{.195\linewidth}
  \centering
  \includegraphics[width=.99\linewidth]{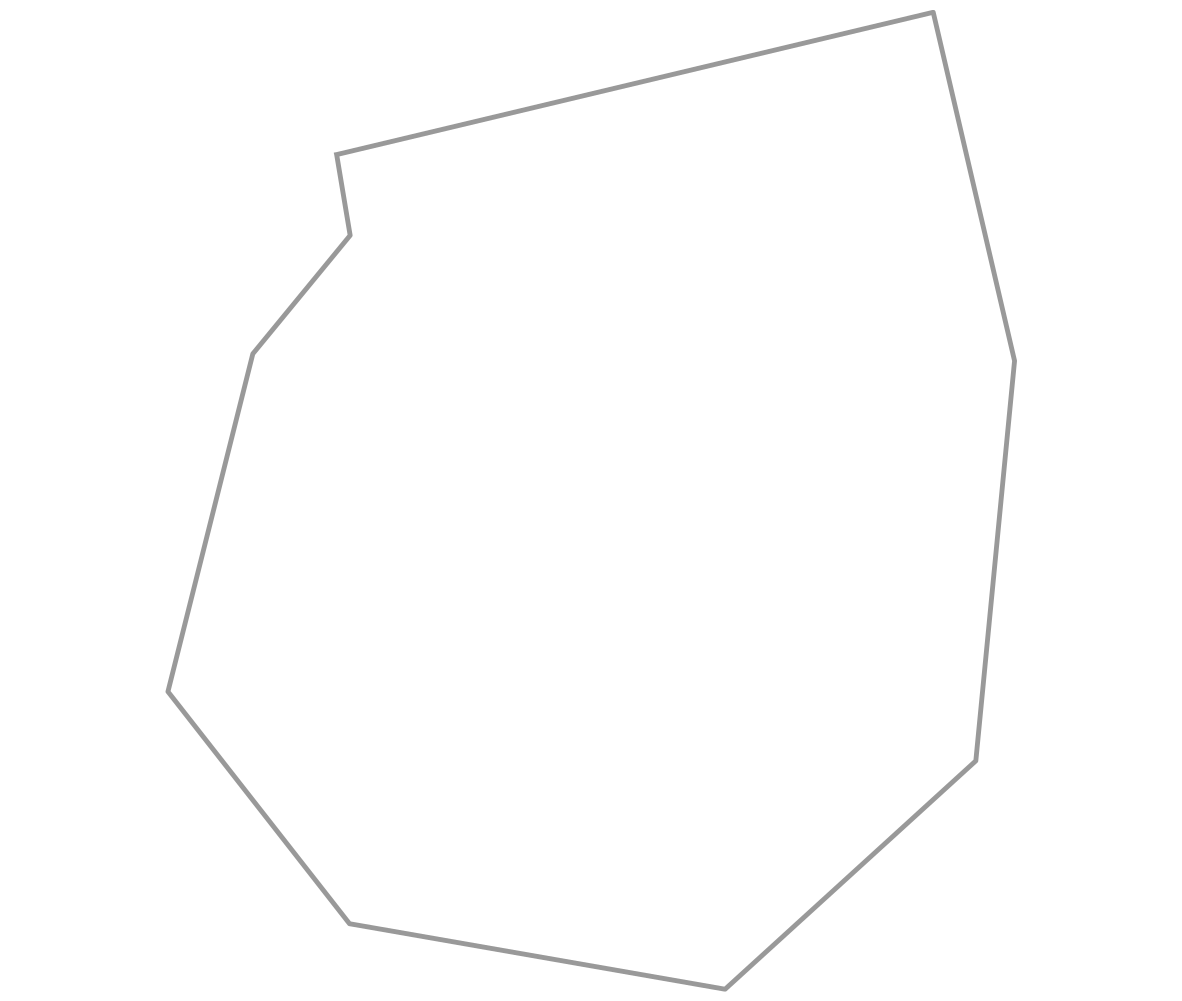}\\[-1pt]
  \label{fig_ef6}
\end{subfigure}
\begin{subfigure}[t]{.195\linewidth}
  \centering
  \includegraphics[width=.99\linewidth]{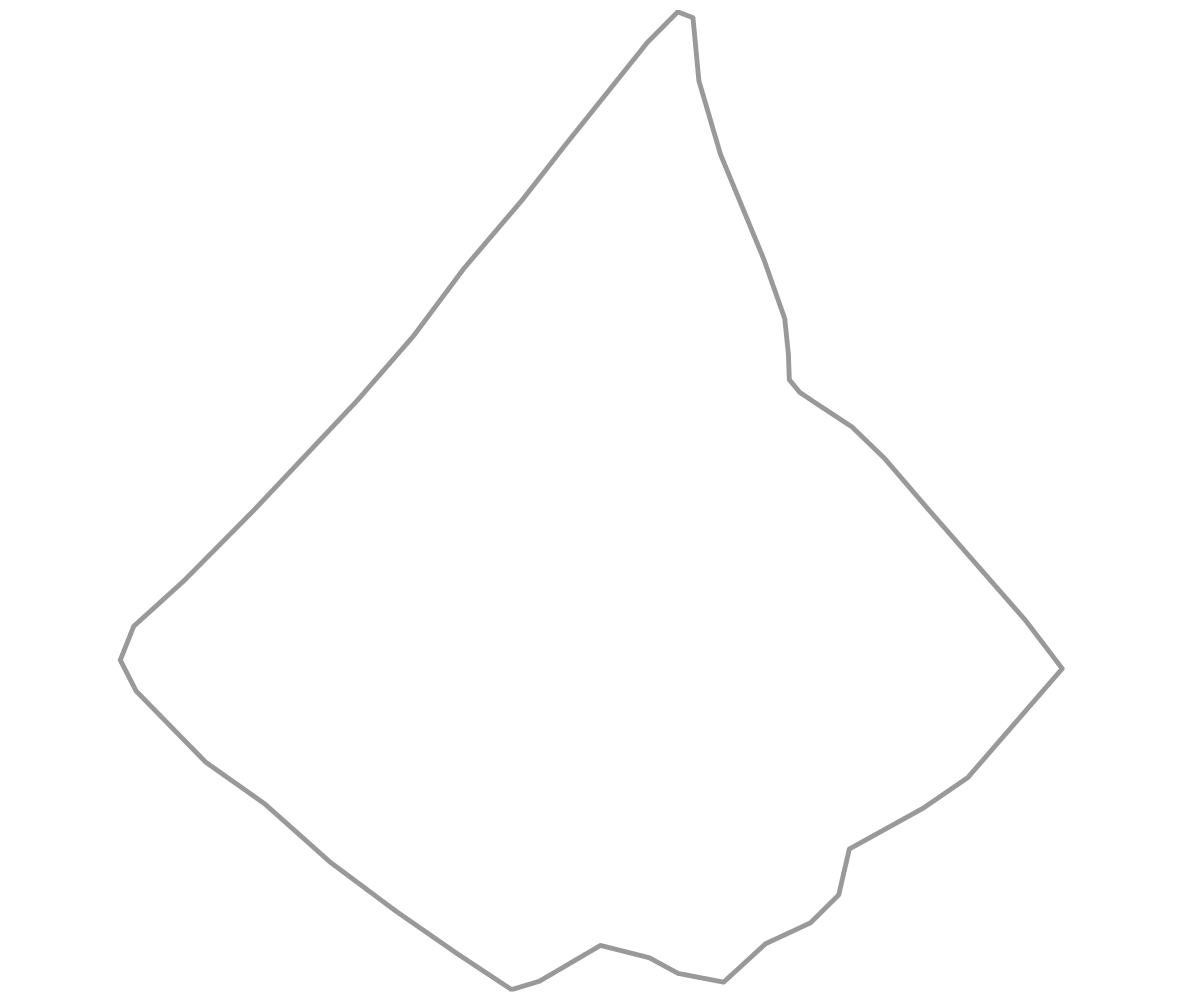}\\[-1pt]
  \label{fig_ef7}
\end{subfigure}

\begin{subfigure}[t]{.195\linewidth}
  \centering
  \includegraphics[width=.99\linewidth]{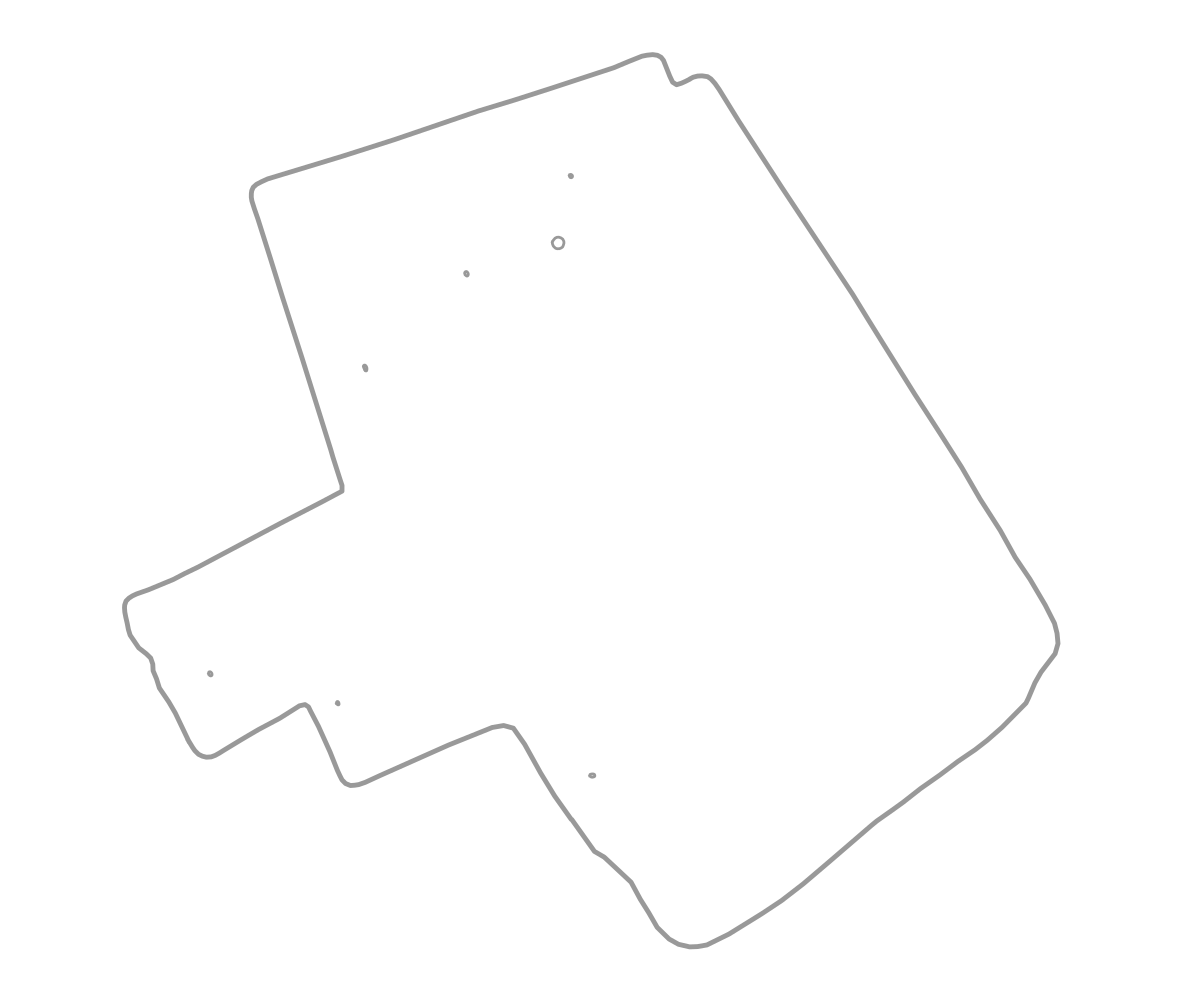}\\[-1pt]
  \label{fig_ef10}
\end{subfigure}
\begin{subfigure}[t]{.195\linewidth}
  \centering
  \includegraphics[width=.99\linewidth]{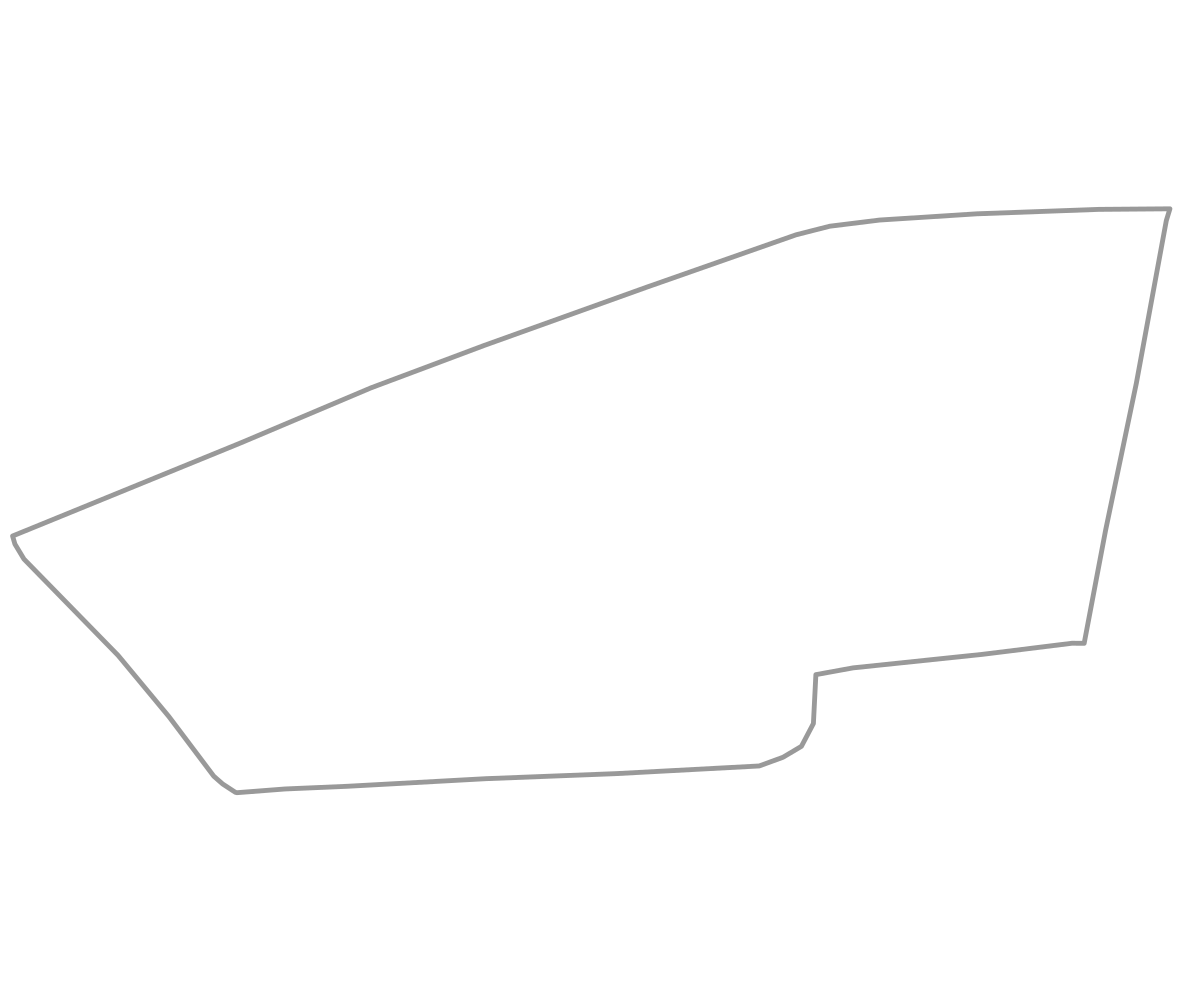}\\[-1pt]
  \label{fig_ef5}
\end{subfigure}
\begin{subfigure}[t]{.195\linewidth}
  \centering
  \includegraphics[width=.99\linewidth]{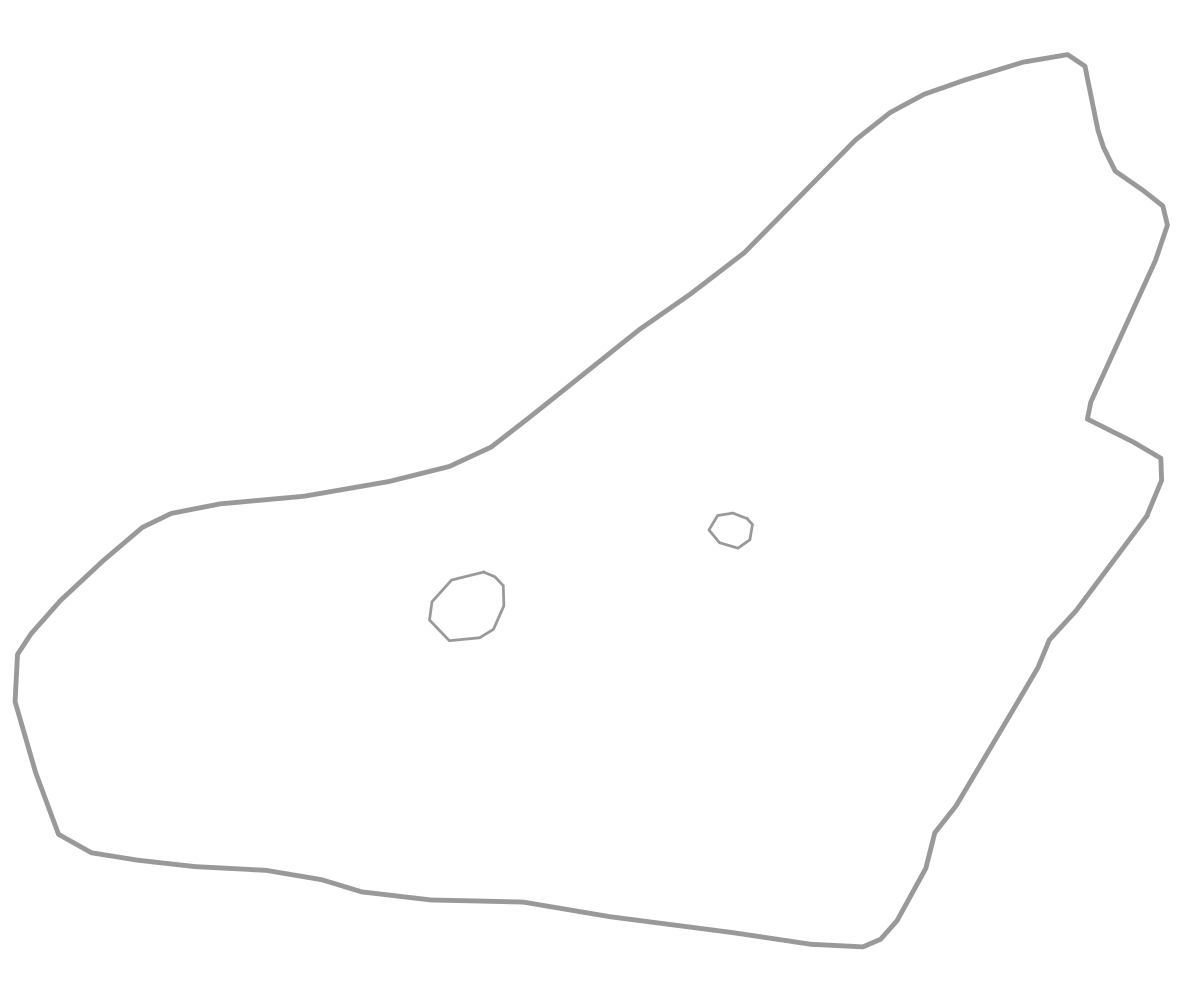}\\[-1pt]
  \label{fig_ef8}
\end{subfigure}
\begin{subfigure}[t]{.195\linewidth}
  \centering
  \includegraphics[width=.99\linewidth]{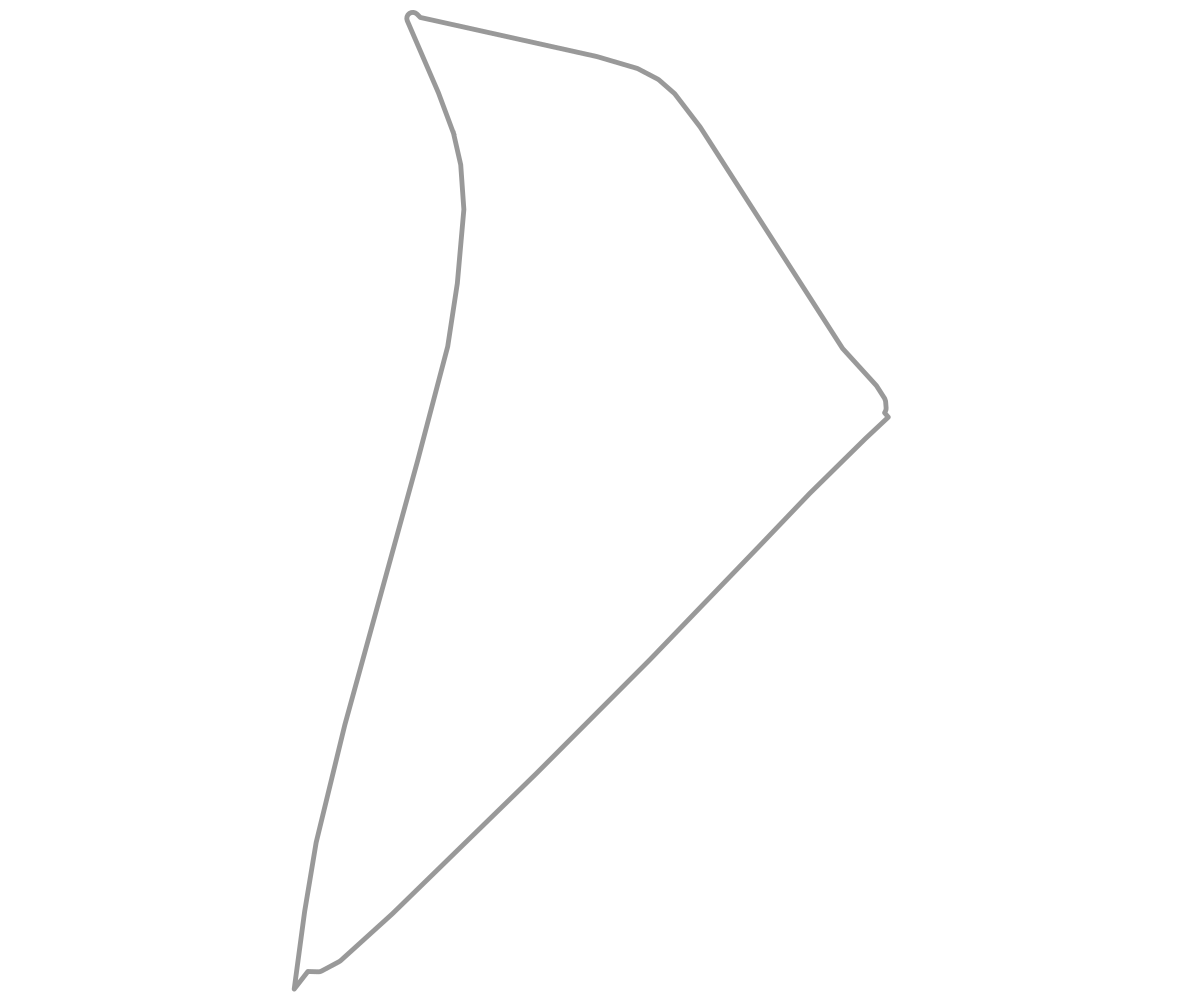}\\[-1pt]
  \label{fig_ef1}
\end{subfigure}
\begin{subfigure}[t]{.195\linewidth}
  \centering
  \includegraphics[width=.99\linewidth]{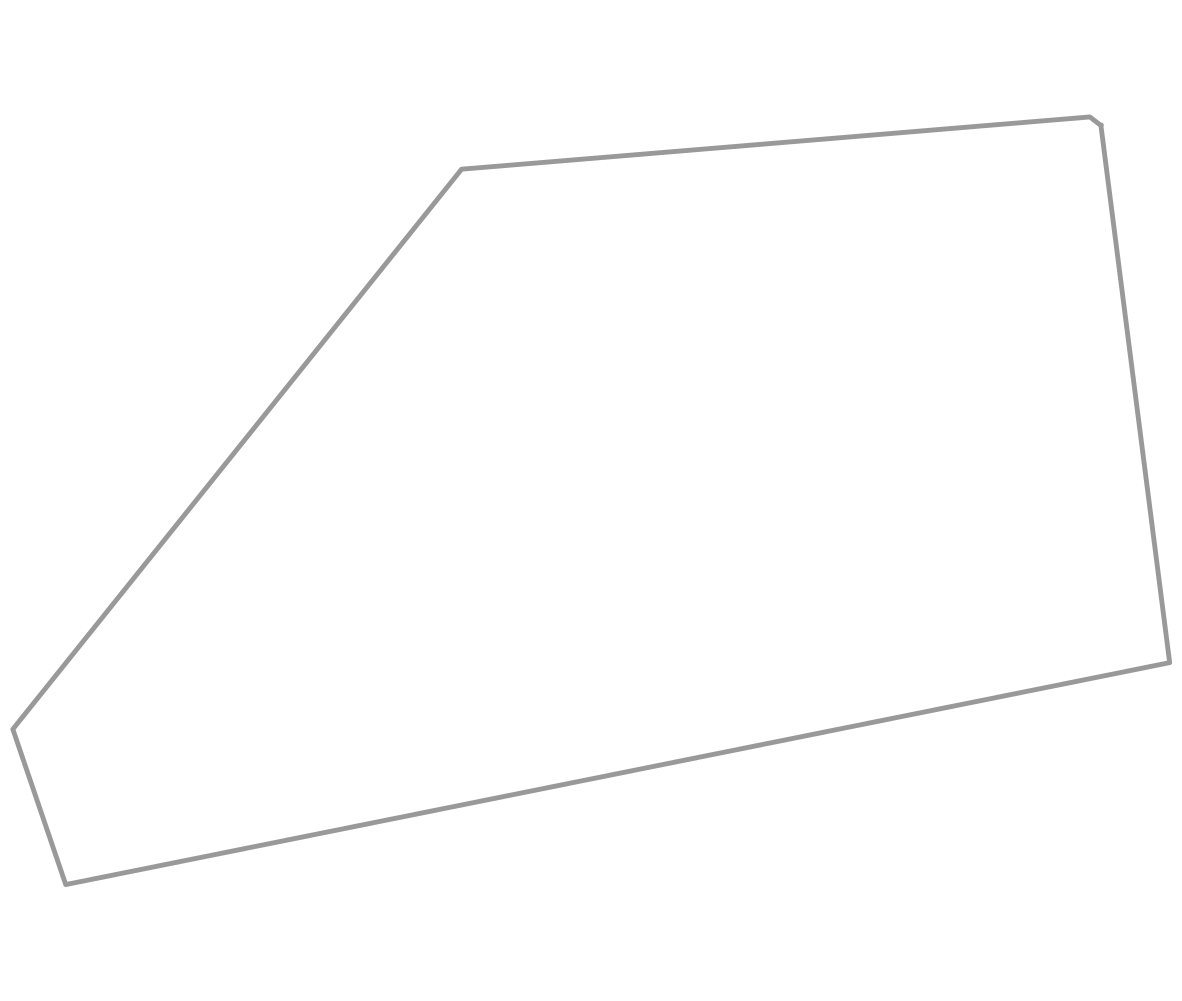}\\[-1pt]
  \label{fig_ef11}
\end{subfigure}
\caption{Visualisation of input data for 10 fields: coordinates of field contours and obstacle areas if applicable. The first 5 fields are ordered in the upper row from left to right. The remaining 5 fields are ordered in the second row from left to right.}
\label{fig_inputdata}
\end{figure*}


Parameters are used uniformly throughout all experiments. Vehicle dynamics are as in \cite{plessen2025smoothing} with a minimum turning radius of 5m. A reference spray volume of  $s_\text{volume}^\text{ref}=5$gpa (gallons per acre) is assumed, which corresponds to 46.78$\ell$/ha. An inter-nozzle spacing of $w_\text{nozzle}=0.5$m is assumed. For the variable rate case and working width $W=24$m this results in a maximum of 48 individually controllable sections. For the case of controlling multiple adjacent nozzles with the same signal (e.g. 4 nozzles resulting in $w_\text{section}=2$m), the number of sections can in general be reduced up to controlling all nozzles as a single section. If the number individually controllable sections is reduced then also the performance of ASC is degraded, ultimately approaching the performance of the 1-section method. 

For the generation of freeform mainfield lanes, for optimisation of area coverage path planning, and for the smoothing of headland path edges and headland-to-mainfield lane transitions the methods from \cite{plessen2019optimal}, \cite{plessen2021freeform} and \cite{plessen2025smoothing} were used with path discretisation spacing of 1m, respectively. 

\begin{figure*}
\centering
\begin{subfigure}[t]{.33\linewidth}
  \centering
  \includegraphics[width=.99\linewidth]{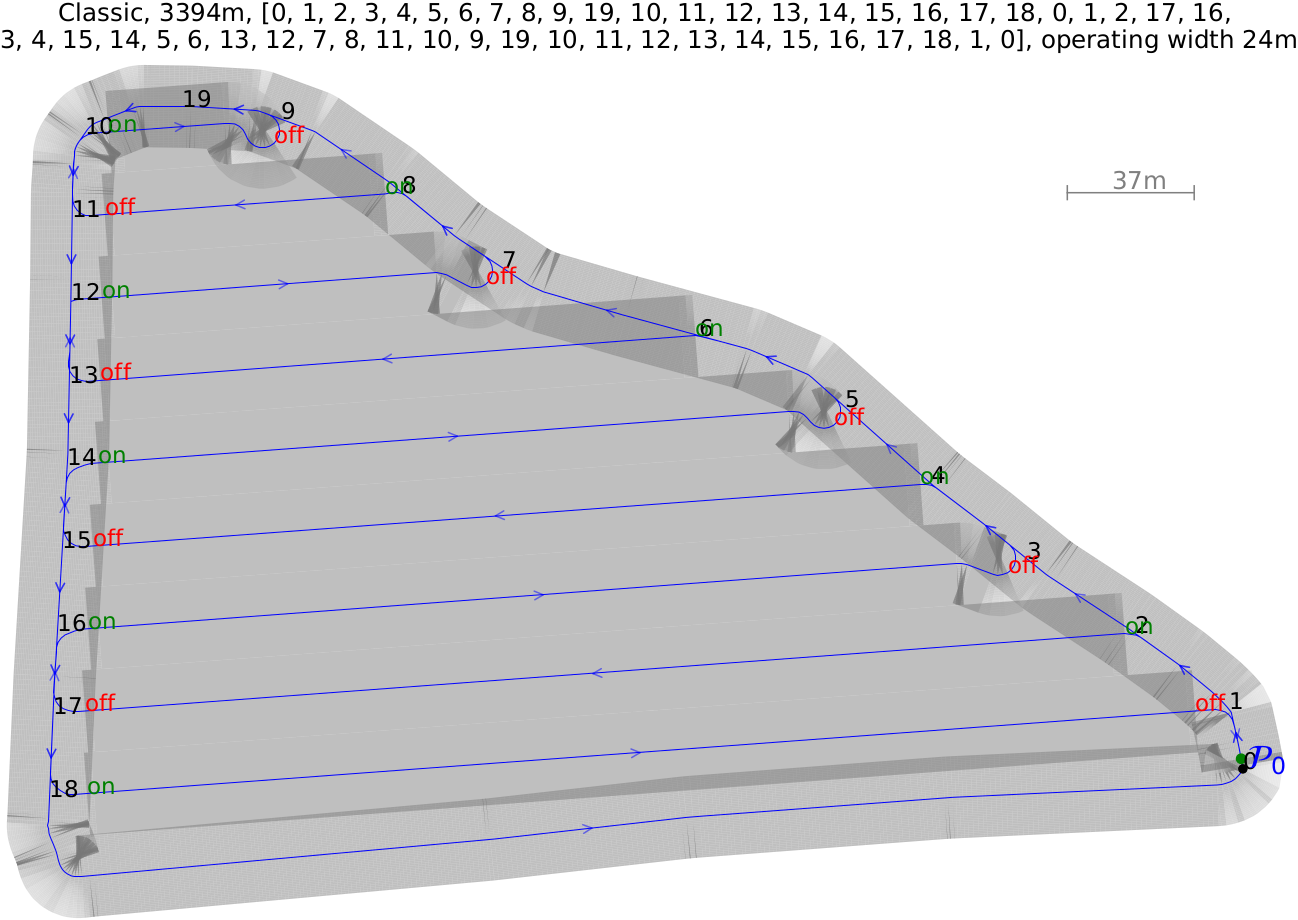}\\[-1pt]
\caption{$S_{\textsf{M}_1}^{1}$ \\[5pt]}
  \label{fig_f14_SC0}
\end{subfigure}
\begin{subfigure}[t]{.33\linewidth}
  \centering
  \includegraphics[width=.99\linewidth]{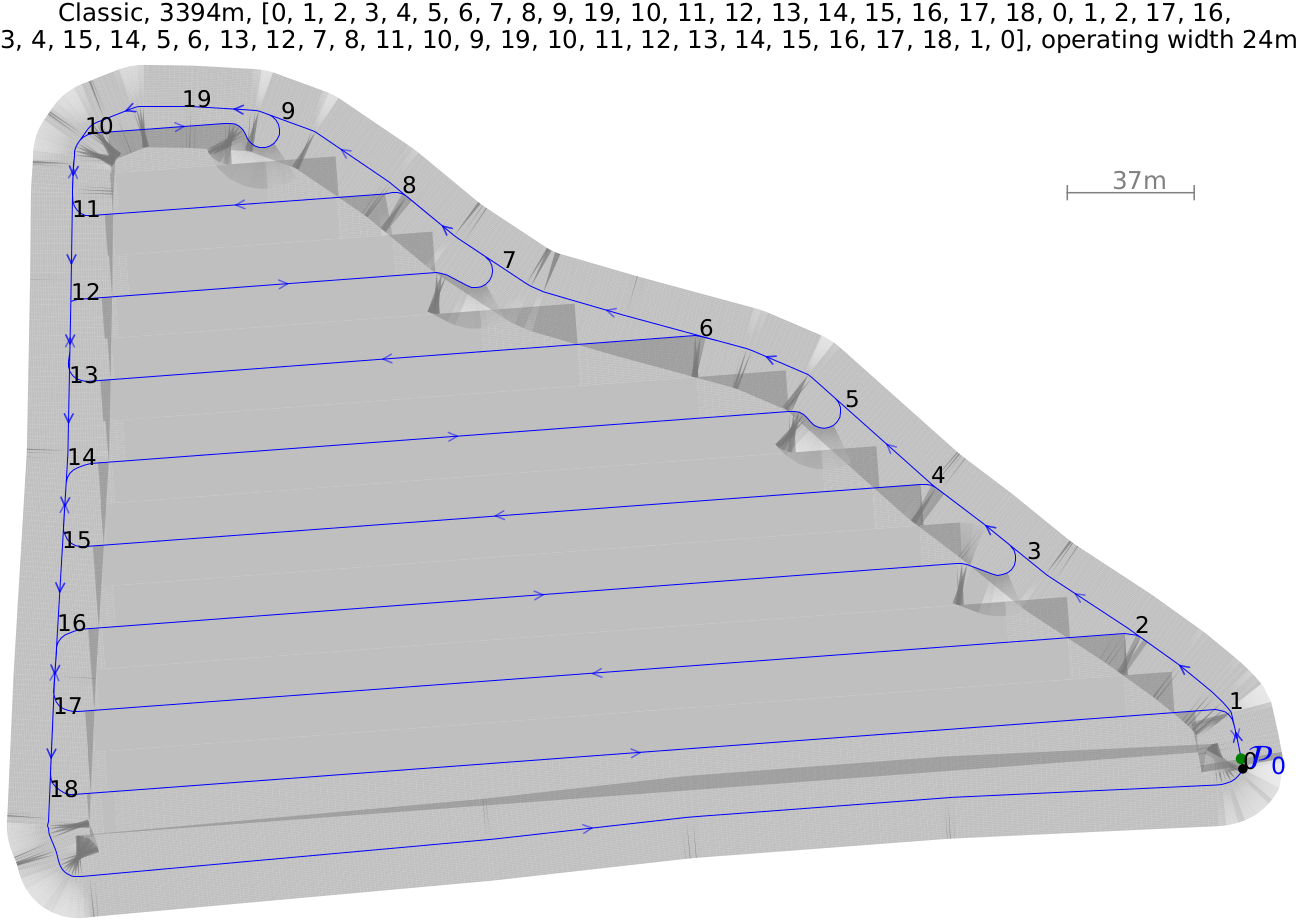}\\[-1pt]
\caption{$S_{\textsf{M}_1}^{2}$\\[5pt]}
  \label{fig_f14_SC1}
\end{subfigure}
\begin{subfigure}[t]{.33\linewidth}
  \centering
  \includegraphics[width=.99\linewidth]{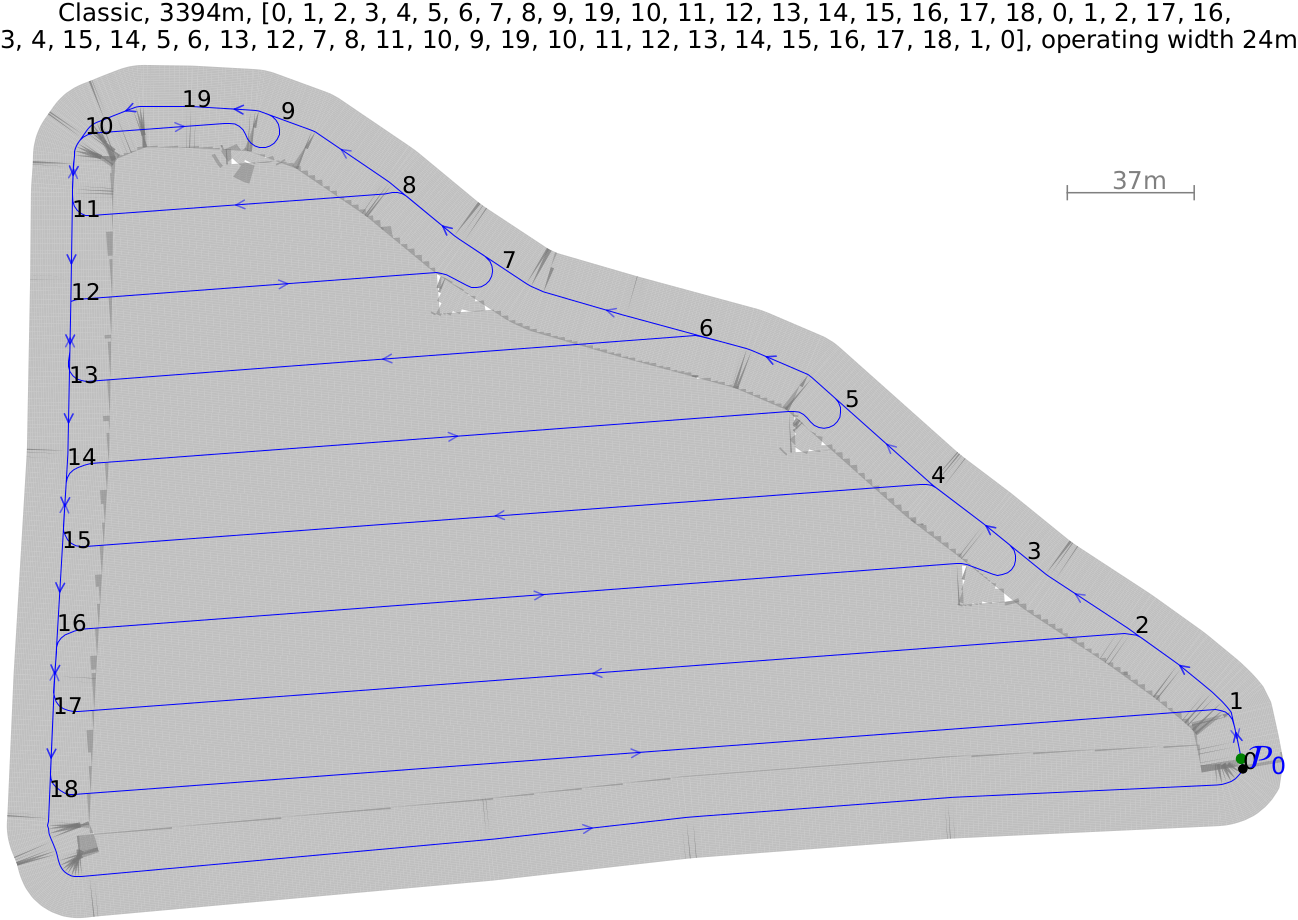}\\[-1pt]
\caption{$S_{\textsf{M}_1}^{48}$\\[5pt]}
  \label{fig_f14_SC2}
\end{subfigure}
\begin{subfigure}[t]{.33\linewidth}
  \centering
  \includegraphics[width=.99\linewidth]{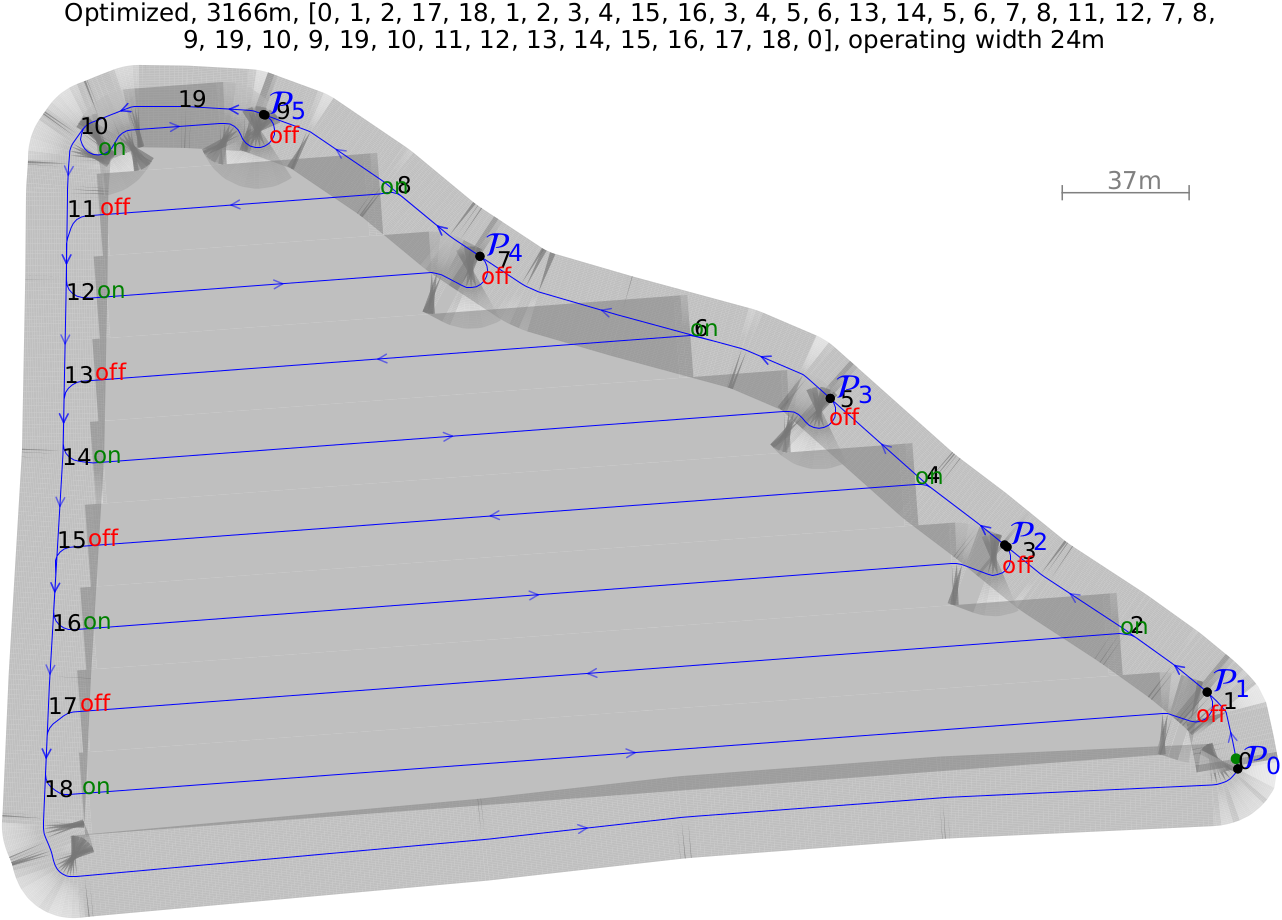}\\[-1pt]
\caption{$S_{\textsf{M}_2}^{1}$}
  \label{fig_f14_SO0}
\end{subfigure}
\begin{subfigure}[t]{.33\linewidth}
  \centering
  \includegraphics[width=.99\linewidth]{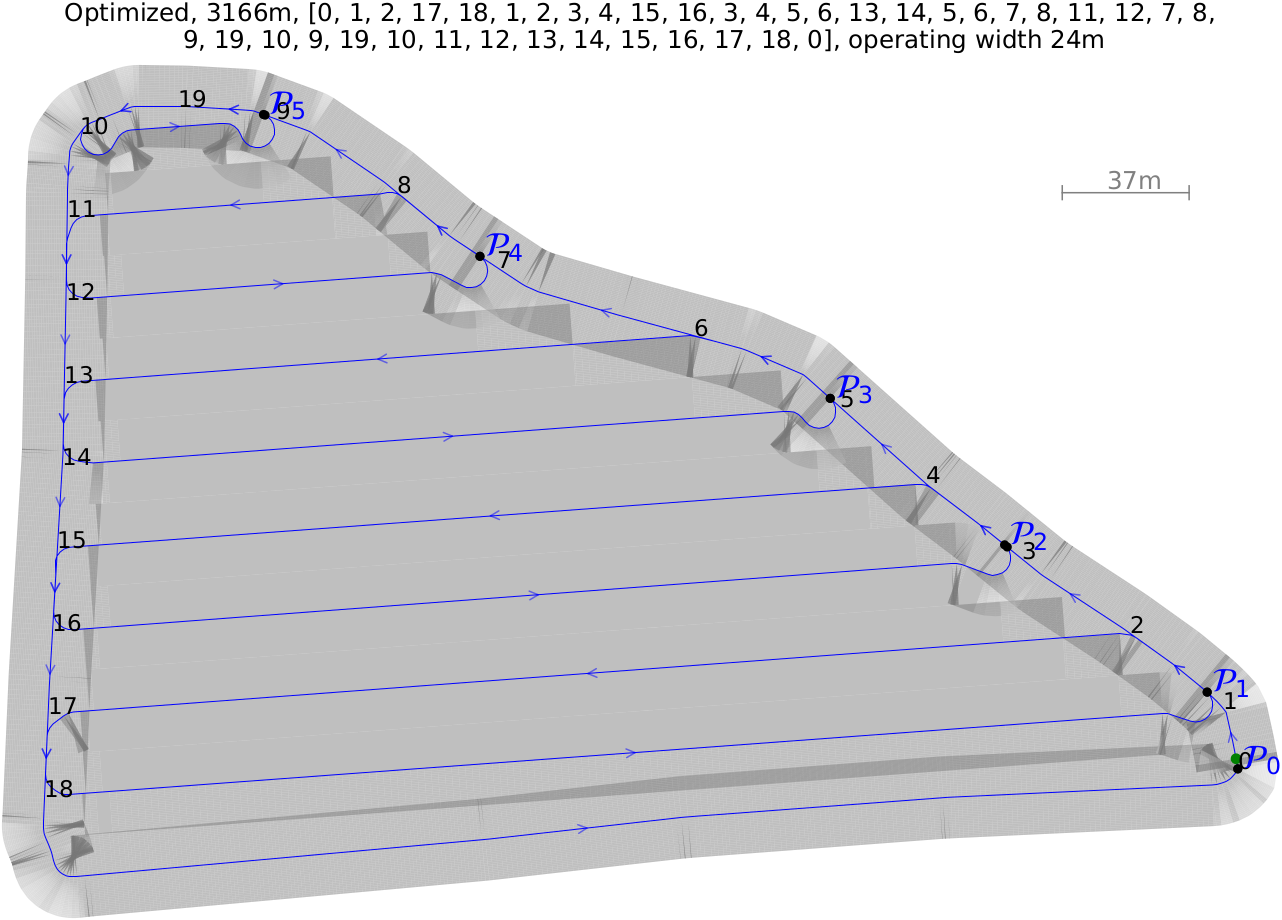}\\[-1pt]
\caption{$S_{\textsf{M}_2}^{2}$}
  \label{fig_f14_SO1}
\end{subfigure}
\begin{subfigure}[t]{.33\linewidth}
  \centering
  \includegraphics[width=.99\linewidth]{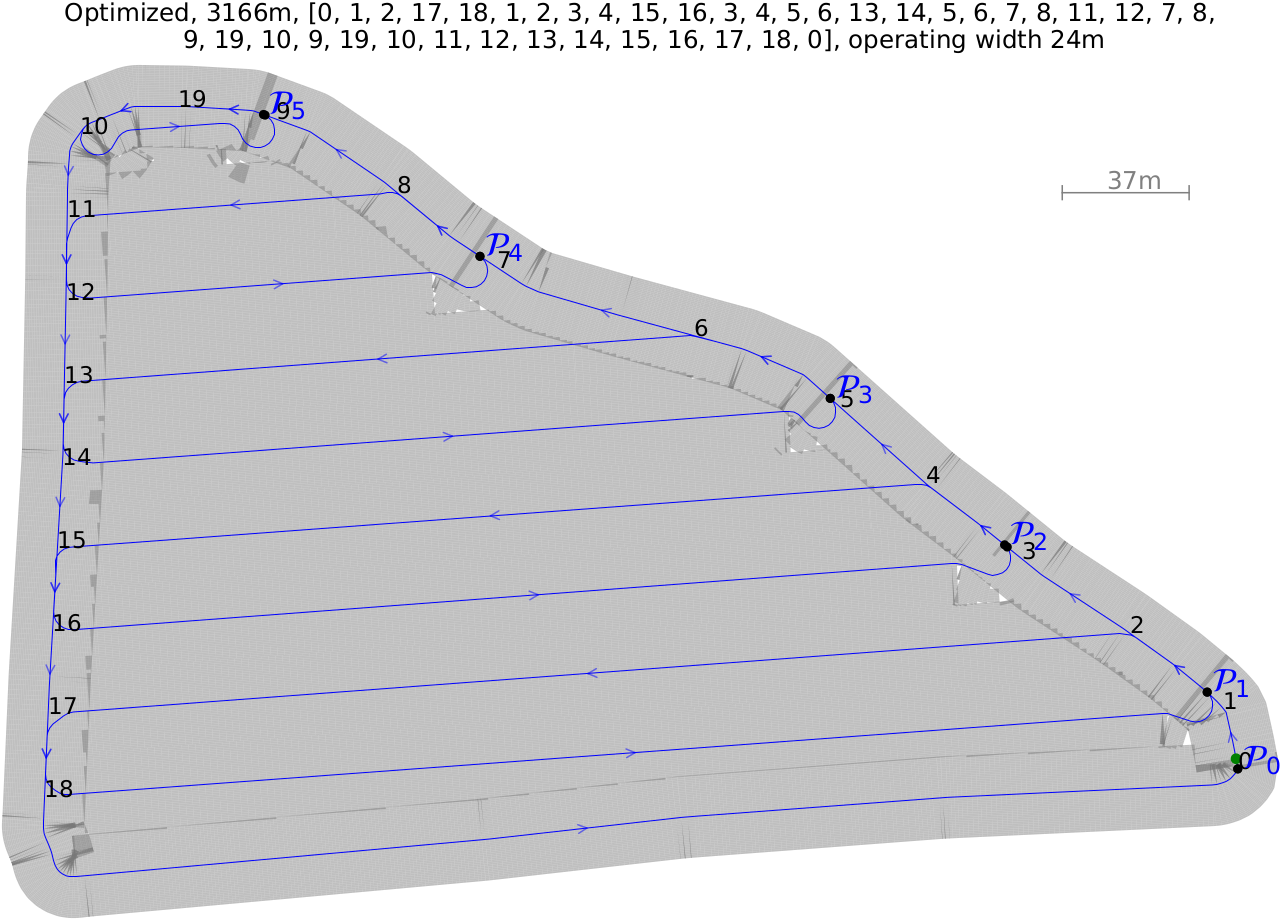}\\[-1pt]
\caption{$S_{\textsf{M}_2}^{48}$}
  \label{fig_f14_SO2}
\end{subfigure}
\caption{Field example 1: visual comparison of 6 different setups. See Table \ref{tab_s} for quantitative evaluation. }
\label{fig_f14}
\end{figure*}

\begin{figure*}
\centering
\begin{subfigure}[t]{.33\linewidth}
  \centering
  \includegraphics[width=.99\linewidth]{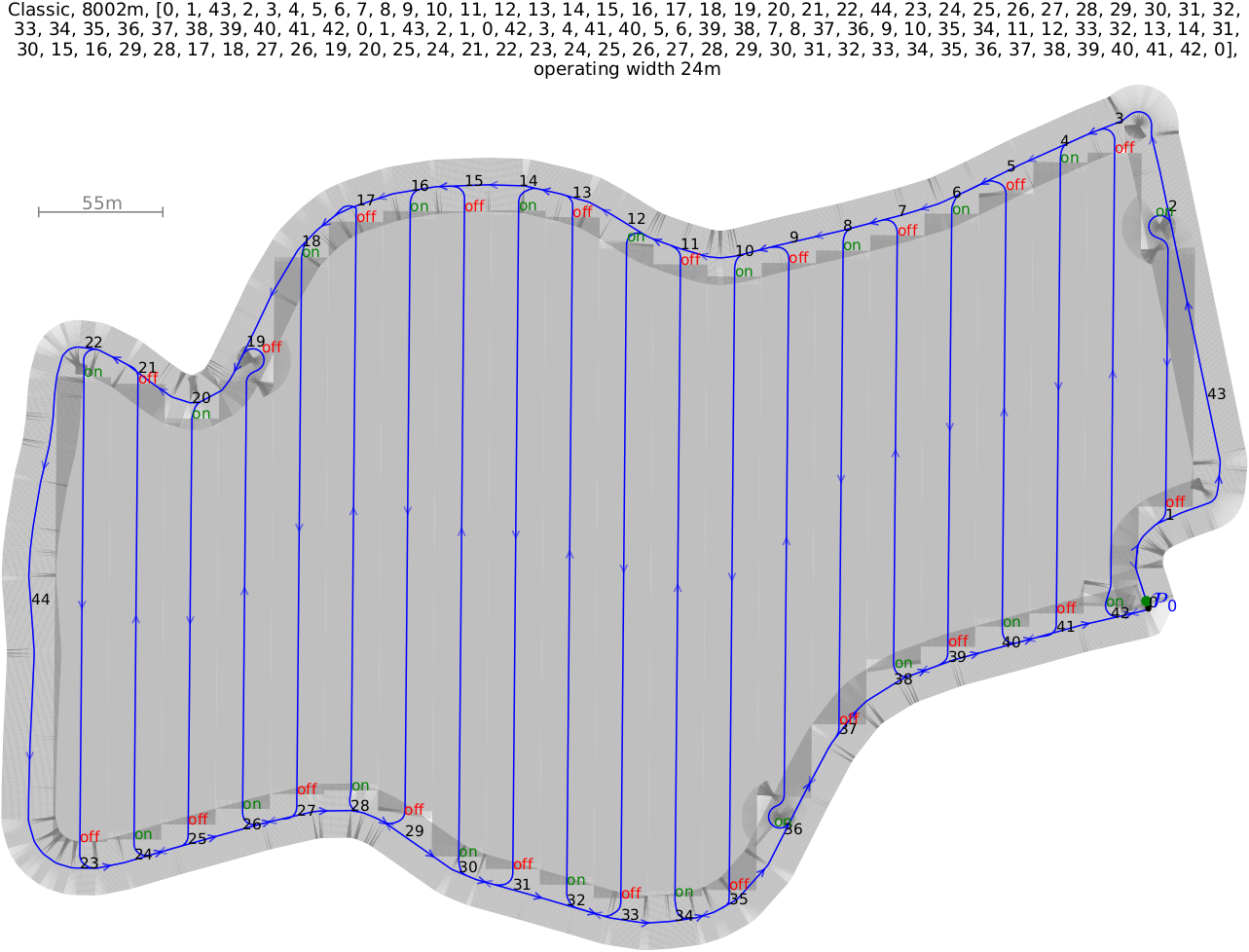}\\[-1pt]
\caption{$S_{\textsf{M}_1}^{1}$ \\[10pt]}
  \label{fig_f17_SC0}
\end{subfigure}
\begin{subfigure}[t]{.33\linewidth}
  \centering
  \includegraphics[width=.99\linewidth]{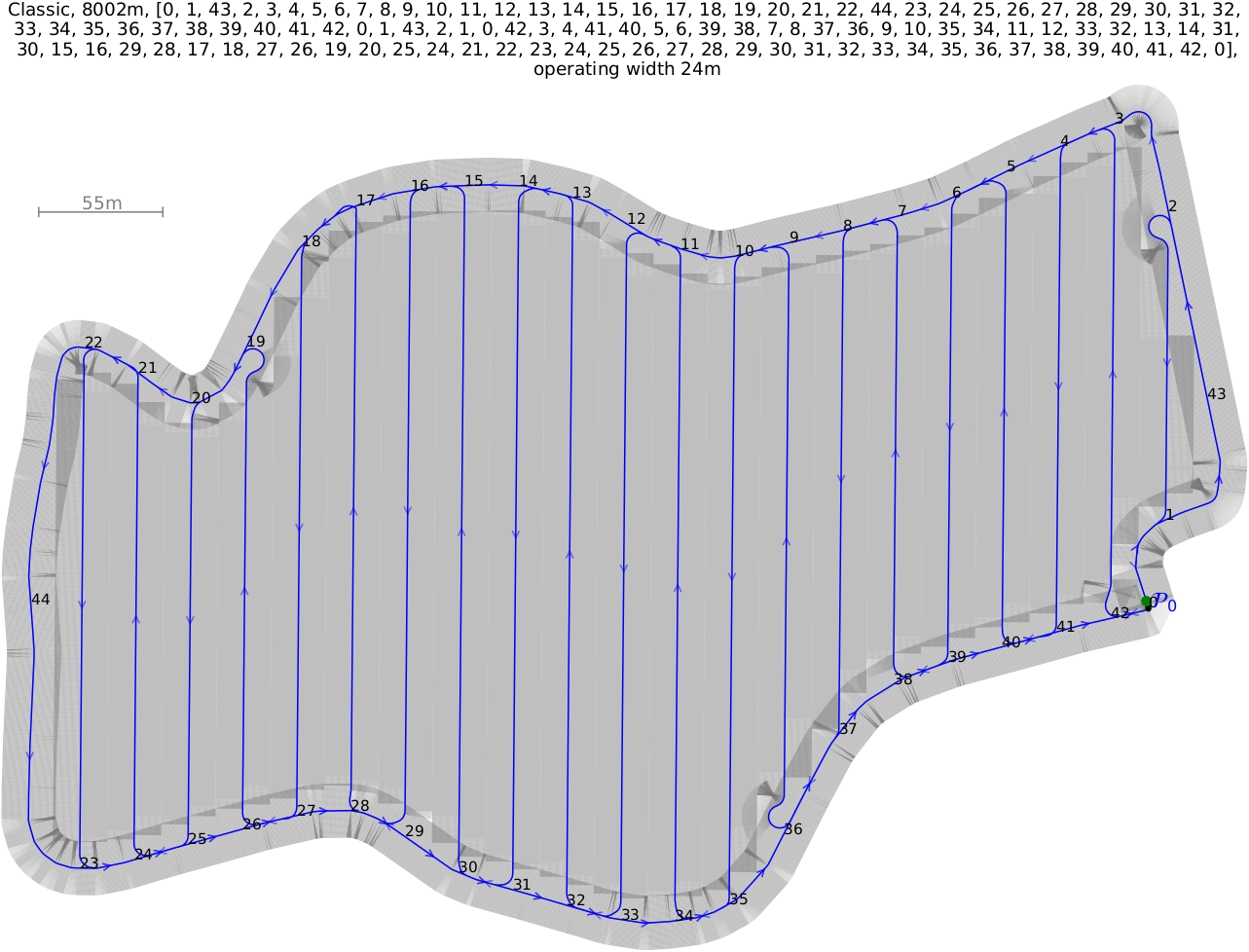}\\[-1pt]
\caption{$S_{\textsf{M}_1}^{2}$\\[10pt]}
  \label{fig_f17_SC1}
\end{subfigure}
\begin{subfigure}[t]{.33\linewidth}
  \centering
  \includegraphics[width=.99\linewidth]{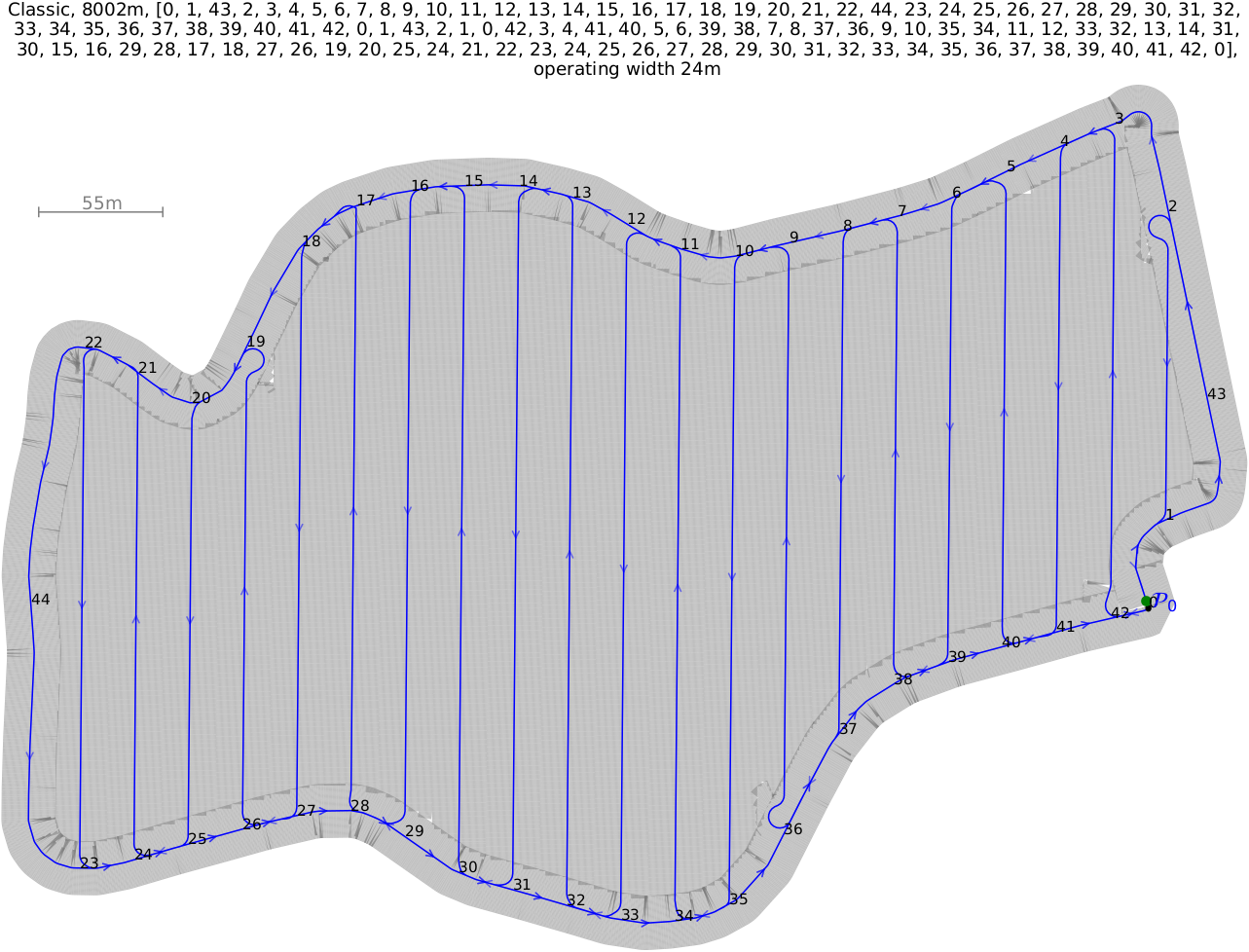}\\[-1pt]
\caption{$S_{\textsf{M}_1}^{48}$\\[10pt]}
  \label{fig_f17_SC2}
\end{subfigure}
\begin{subfigure}[t]{.33\linewidth}
  \centering
  \includegraphics[width=.99\linewidth]{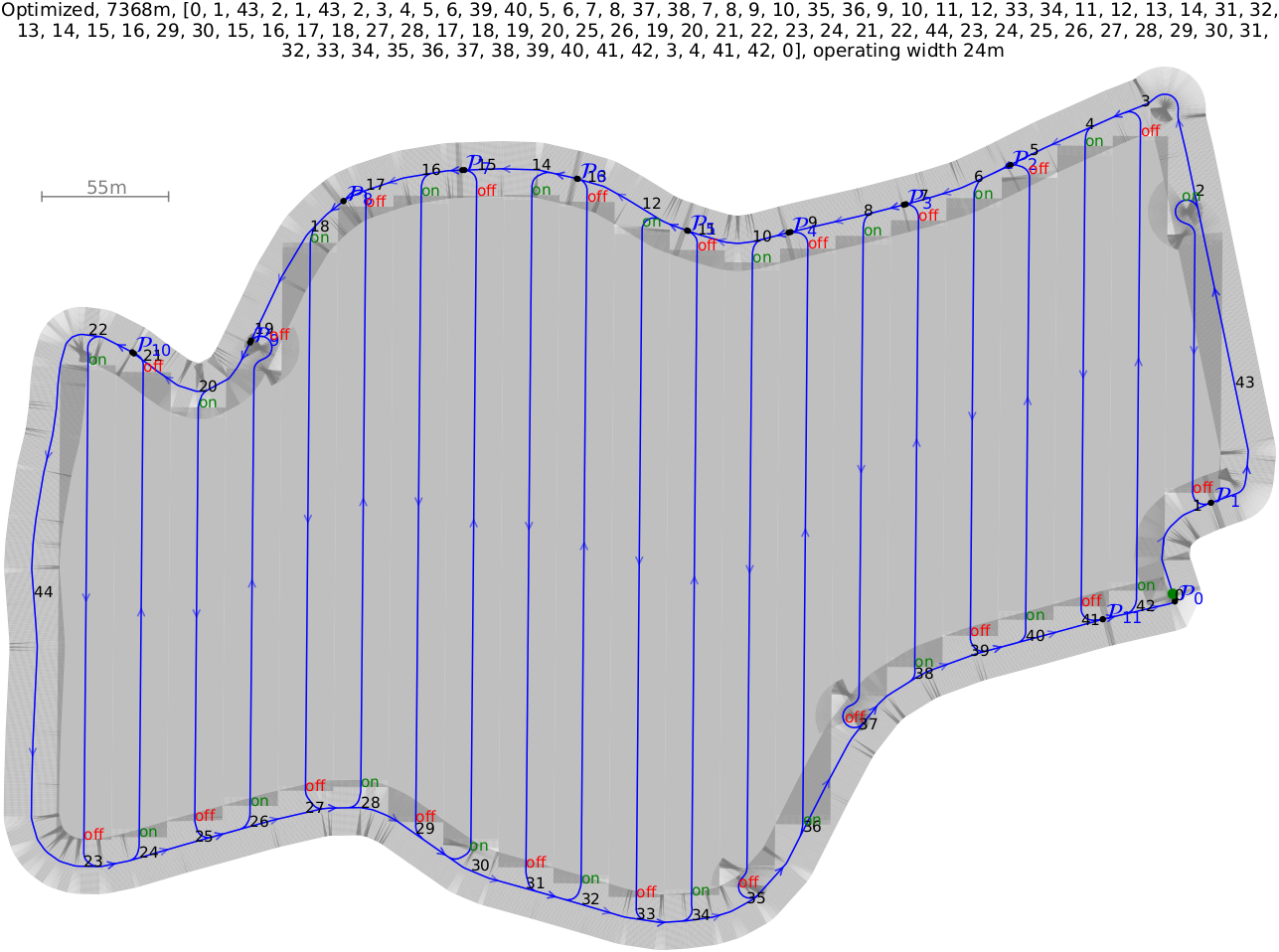}\\[-1pt]
\caption{$S_{\textsf{M}_2}^{1}$}
  \label{fig_f17_SO0}
\end{subfigure}
\begin{subfigure}[t]{.33\linewidth}
  \centering
  \includegraphics[width=.99\linewidth]{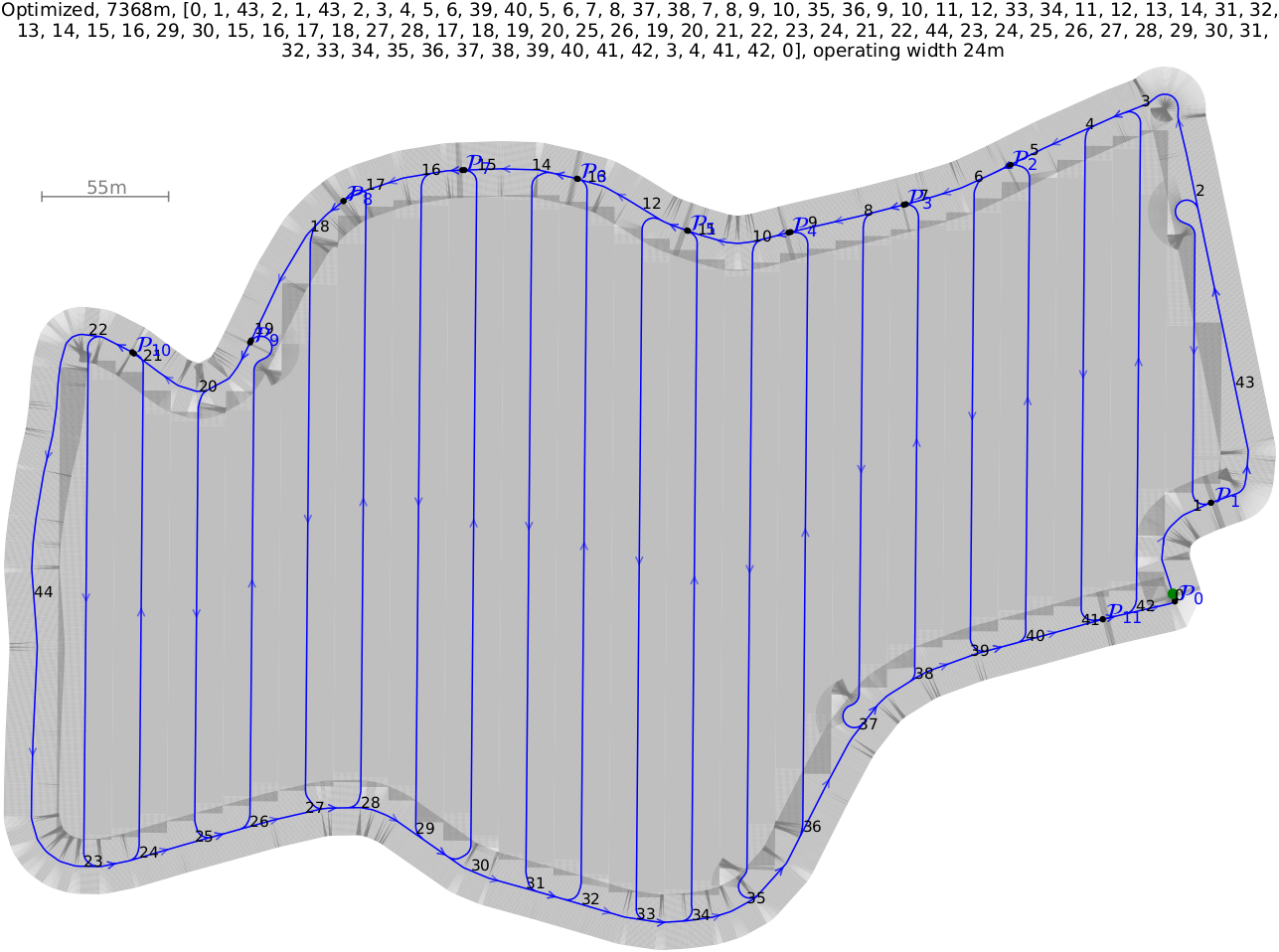}\\[-1pt]
\caption{$S_{\textsf{M}_2}^{2}$}
  \label{fig_f17_SO1}
\end{subfigure}
\begin{subfigure}[t]{.33\linewidth}
  \centering
  \includegraphics[width=.99\linewidth]{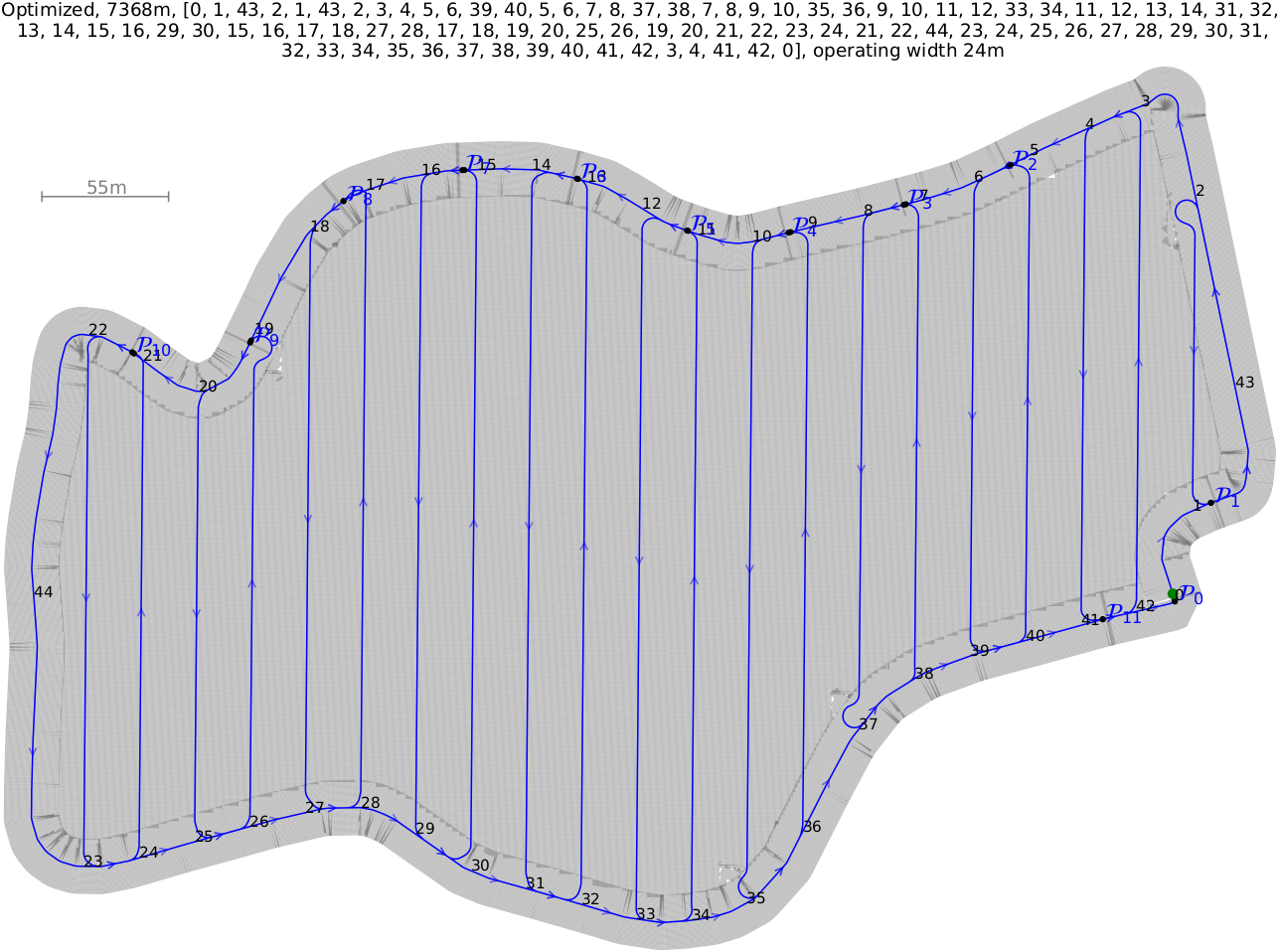}\\[-1pt]
\caption{$S_{\textsf{M}_2}^{48}$}
  \label{fig_f17_SO2}
\end{subfigure}
\caption{Field example 2: visual comparison of 6 different setups. See Table \ref{tab_s} for quantitative evaluation. }
\label{fig_f17}
\end{figure*}

The 10 real-world fields used for evaluation are depicted in Fig. \ref{fig_inputdata}. This data includes examples with multiple obstacle areas, freeform mainfield lanes, and non-convex field contours. Quantitative results are given in Table \ref{tab_L} and \ref{tab_s}. Qualitative results for 2 examples are visualised in Fig. \ref{fig_f14} and \ref{fig_f17}. Visualisations for all other examples are provided in \ref{sec_appendix}.

\begin{table}
\centering
\begin{tabular}{|c|r|rrrr|}
\hline 
\rowcolor[gray]{0.8} & & & & & \\[-8pt]
\rowcolor[gray]{0.8} Ex. & $A_\text{field}$ & $L_{\textsf{M}_1}$ & $L_{\textsf{M}_2}$ & $\Delta L$ (m) & $\Delta L$ ($\%$) \\[2pt]
\hline
& & & & & \\[-8pt]
1 & 6.0ha & 3394m & 3166m & -228m & \textbf{-6.7\%}\\\hline
2 & 14.6ha & 8002m & 7368m & -634m & \textbf{-7.9\%}\\\hline
3 & 10.1ha & 5995m & 5488m & -507m & \textbf{-8.5\%}\\\hline
4 & 11.8ha & 6428m & 5954m & -474m & \textbf{-7.4\%}\\\hline
5 & 13.5ha & 7231m & 6784m & -447m & \textbf{-6.2\%}\\\hline
6 & 38.5ha & 19413m & 18447m & -966m & \textbf{-5.0\%}\\\hline
7 & 6.5ha & 3512m & 3346m & -166m & \textbf{-4.7\%}\\\hline
8 & 13.6ha & 7713m & 7087m & -625m & \textbf{-8.1\%}\\\hline
9 & 4.5ha & 2834m & 2485m & -349m & \textbf{-12.3\%}\\\hline
10 & 7.2ha & 3890m & 3739m & -151m & \textbf{-3.9\%}\\
\hline
\end{tabular}
\caption{Pathlengths of numerical experiments for methods $\textsf{M}_1$ and $\textsf{M}_2$.}
\label{tab_L}
\end{table}


\begin{table*}
\centering
\begin{tabular}{|c|rr|rrr|rrr|}
\hline 
\rowcolor[gray]{0.8} & & & & & & & &  \\[-8pt]
\rowcolor[gray]{0.8} Ex. & $A_\text{field}$ & $S_\text{field}^\text{ref}$ & $S_{\textsf{M}_1}^{1}$ & $S_{\textsf{M}_1}^{2}$ & $S_{\textsf{M}_1}^{48}$ & $S_{\textsf{M}_2}^{1}$ & $S_{\textsf{M}_2}^{2}$ & $S_{\textsf{M}_2}^{48}$   \\[2pt]
\rowcolor[gray]{0.8}  &  &  & $\Delta S_{\textsf{M}_1,\text{m}}^{1,\text{ref}}$ & $\Delta S_{\textsf{M}_1,\text{m}}^{2,\text{ref}}$ & $\Delta S_{\textsf{M}_1,\text{m}}^{48,\text{ref}}$ & $\Delta S_{\textsf{M}_2,\text{m}}^{1,\text{ref}}$ & $\Delta S_{\textsf{M}_2,\text{m}}^{2,\text{ref}}$ & $\Delta S_{\textsf{M}_2,\text{m}}^{48,\text{ref}}$   \\[2pt]
\rowcolor[gray]{0.8}  &  &  & $\Delta S_{\textsf{M}_1,\%}^{1,\text{ref}}$ & $\Delta S_{\textsf{M}_1,\%}^{2,\text{ref}}$ & $\Delta S_{\textsf{M}_1,\%}^{48,\text{ref}}$ & $\Delta S_{\textsf{M}_2,\%}^{1,\text{ref}}$ & $\Delta S_{\textsf{M}_2,\%}^{2,\text{ref}}$ & $\Delta S_{\textsf{M}_2,\%}^{48,\text{ref}}$   \\[2pt]
\hline
& & & & & & & & \\[-8pt]
1 & 6.0ha & 282.8$\ell$  & 358.9$\ell$ & 342.3$\ell$ & 288.7$\ell$ & 372.7$\ell$ & 352.8$\ell$ & 289.7$\ell$ \\
  &  & & 76.1$\ell$ & 59.5$\ell$ & 5.9$\ell$ & 89.9$\ell$ & 70.0$\ell$ & 6.9$\ell$ \\
  &  & & 26.9\% & 21.0\% & 2.1\% & 31.8\% & 24.7\% & 2.4\% \\\hline 
2 & 14.6ha & 682.1$\ell$  & 881.2$\ell$ & 849.3$\ell$ & 695.7$\ell$ & 901.6$\ell$ & 856.0$\ell$ & 697.2$\ell$ \\
  &  & & 119.1$\ell$ & 167.2$\ell$ & 13.6$\ell$ & 219.5$\ell$ & 173.9$\ell$ & 15.1$\ell$ \\
  &  & & 29.2\% & 24.5\% & 2.0\% & 32.2\% & 25.5\% & 2.2\% \\\hline    
3 & 10.1ha & 473.8$\ell$  & 676.3$\ell$ & 646.8$\ell$ & 487.2$\ell$ & 681.2$\ell$ & 645.8$\ell$ & 488.9$\ell$ \\
  &  & & 202.5$\ell$ & 173.0$\ell$ & 13.4$\ell$ & 207.4$\ell$ & 172.0$\ell$ & 15.1$\ell$ \\
  &  & & 42.7\% & 36.5\% & 2.8\% & 43.8\% & 36.3\% & 3.2\% \\\hline 
4 & 11.8ha & 552.6$\ell$  & 702.9$\ell$ & 656.6$\ell$ & 560.6$\ell$ & 721.3$\ell$ & 665.5$\ell$ & 561.7$\ell$ \\
  &  & & 150.3$\ell$ & 104.0$\ell$ & 8.0$\ell$ & 168.7$\ell$ & 112.9$\ell$ & 9.1$\ell$ \\
  &  & & 27.2\% & 18.8\% & 1.4\% & 30.5\% & 20.4\% & 1.6\% \\\hline    
5 & 13.5ha & 629.7$\ell$  & 780.2$\ell$ & 758.9$\ell$ & 638.7$\ell$ & 756.3$\ell$ & 733.1$\ell$ & 640.5$\ell$ \\
  &  & & 150.5$\ell$ & 129.2$\ell$ & 9$\ell$ & 126.6$\ell$ & 103.4$\ell$ & 10.8$\ell$ \\
  &  & & 23.9\% & 20.5\% & 1.4\% & 20.1\% & 16.4\% & 1.7\% \\\hline  
6 & 38.5ha & 1799.7$\ell$  & 2621.3$\ell$ & 2592.1$\ell$ & 1859.1$\ell$ & 2627.2$\ell$ & 2593.2$\ell$ & 1860.9$\ell$ \\
  &  & & 821.6$\ell$ & 792.4$\ell$ & 59.4$\ell$ & 827.5$\ell$ & 793.5$\ell$ & 61.2$\ell$ \\
  &  & & 45.6\% & 44.0\% & 3.3\% & 45.9\% & 44.1\% & 3.4\% \\\hline 

7 & 6.5ha & 304.6$\ell$  & 489.6$\ell$ & 486.8$\ell$ & 314.6$\ell$ & 495.9$\ell$ & 490.1$\ell$ & 315.1$\ell$ \\
  &  & & 185.0$\ell$ & 182.2$\ell$ & 10.0$\ell$ & 191.3$\ell$ & 185.5$\ell$ & 10.5$\ell$ \\
  &  & & 60.7\% & 59.8\% & 3.2\% & 62.8\% & 60.9\% & 3.4\% \\\hline 
8 & 13.6ha & 636.4$\ell$  & 980.9$\ell$ & 938.5$\ell$ & 668.8$\ell$ & 1086.4$\ell$ & 972.4$\ell$ & 665.5$\ell$ \\
  &  & & 344.5$\ell$ & 302.1$\ell$ & 32.4$\ell$ & 450.0$\ell$ & 336.0$\ell$ & 29.1$\ell$ \\
  &  & & 54.1\% & 47.5\% & 5.1\% & 70.7\% & 52.8\% & 4.6\% \\\hline 
9 & 4.5ha & 211.1$\ell$  & 397.9$\ell$ & 371.8$\ell$ & 225.3$\ell$ & 479.1$\ell$ & 450.6$\ell$ & 225.8$\ell$ \\
  &  & & 186.8$\ell$ & 160.7$\ell$ & 14.2$\ell$ & 268.0$\ell$ & 239.5$\ell$ & 14.7$\ell$ \\
  &  & & 88.5\% & 76.1\% & 6.7\% & 126.9\% & 113.4\% & 6.9\% \\\hline 
10 & 7.2ha & 336.3$\ell$  & 434.2$\ell$ & 410.8$\ell$ & 342.7$\ell$ & 436.5$\ell$ & 413.0$\ell$ & 343.3$\ell$ \\
  &  & & 97.9$\ell$ & 74.5$\ell$ & 6.4$\ell$ & 100.2$\ell$ & 76.7$\ell$ & 7.0$\ell$ \\
  &  & & 29.1\% & 22.1\% & 1.9\% & 29.8\% & 22.8\% & 2.1\% \\        
\hline
\end{tabular}
\caption{Spray volume results for 6 different setups, $S_{\textsf{M}_k}^{j},~\forall k\in\{1,2\},~\forall j\in\{1,2,48\}$. For a given field area $A_\text{field}$, the theoretical ideal spray volume for uniform spraying is $S_\text{field}^\text{ref}=A_\text{field}s_\text{volume}^\text{ref}$. The differences measured in metres and percent are 
$\Delta S_{\textsf{M}_k,\text{m}}^{j,\text{ref}}=S_{\textsf{M}_k,\text{m}}^{j,\text{ref}}-S_\text{field}^\text{ref}$ and $\Delta S_{\textsf{M}_k,\%}^{j,\text{ref}}=(S_{\textsf{M}_k,\text{m}}^{j,\text{ref}}-S_\text{field}^\text{ref})/S_\text{field}^\text{ref}$, respectively.}
\label{tab_s}
\end{table*}

\section{Discussion\label{sec_discussion}}

\subsection{Key Findings}

The key findings of numerical experiments are discussed:
\begin{enumerate}

\item The pathlength savings potential for the optimisation-based $\textsf{M}_2$-method in comparison to the Boustrophedon-based $\textsf{M}_1$-method is between -3.9\% and -12.3\% (see Table \ref{tab_L}). $\textsf{M}_2$ yielded better results throughout all experiments. Since path planning precedes spraying according to the hierarchical structure of Fig. \ref{fig_4levels}, it can be designed independently from the spraying technique.

\item Regarding spray volume results in Table \ref{tab_s} two main comments can be made. First, the multi-sections method $S_{\textsf{M}_1}^{48}$ with 48 sections over a 24m wide boom bar approaches the reference spray volume $S_{\textsf{field}}^{\text{ref}}$. For field 1 the difference is 2.1\%. For a higher number of sections this percentage is expected to further decrease. In contrast, for the 1- and 2-sections case the percentage difference is higher. For example, for $S_{\textsf{M}_1}^{2}$ and $S_{\textsf{M}_1}^{1}$ the differences with respect to $S_{\textsf{field}}^{\text{ref}}$ are 21\% and 26.9\%, respectively. Second, while these and similar increases in spray volumes for other fields in Table \ref{tab_s} seem high, an important note with respect to spray composition is made. For cereal crop production the water-ratio in a spray mixture is typically very high. For example, for a 100$\ell$ spray volume typically around 99$\ell$ are water and only 1$\ell$ are chemical (fungicides, herbicides, etc). This is from an economic point of view very important, since economic input material costs for water and chemical greatly differ. Hence, aforementioned 21\%-increase for $S_{\textsf{M}_1}^{1}$ does not imply a 21\%-increase in input material costs. Instead, denoting chemical and water cost measured in monetary cost per litre (EUR/$\ell$) by $C_\text{chemical}$ and $C_\text{water}$ and assuming a water ratio of 99\%, the new cost is calculated as $C_\text{cost}=0.01\cdot S_{\textsf{M}_1}^{1}\cdot C_\text{chemical} + 0.99\cdot S_{\textsf{M}_1}^{1}\cdot C_\text{water}$. If $C_\text{water} \ll C_\text{chemical}$ the actual increase in input material costs is much lower than the aforementioned 21\%. Implications are discussed in detail in the next Sec. \ref{subsec_cost}.

\item Both the 1- and 2-sections setup can be implemented robustly, and crucially and in contrast to the ASC-setup, without the need for any localisation sensors. This can be achieved by planting crops in a different direction along the headland and mainfield areas, before using the intersection line between those 2 areas as visual reference line at which spray switching commands can be triggered. This approach is also appropriate for manual operation. See Fig. \ref{fig_2plantingdirections} for illustration.

\end{enumerate}

In view of these findings plus the listing of the plethora of complexities associated with the spraying process in Table \ref{tab_influences}, the 1- and 2-sections setup present a reasonable alternative to ASC and offer multiple benefits. A recommended summary would be to (i) operate localisation-sensor free (no expensive RTK-GPS needed), (ii) use the $\textsf{M}_2$-method for area coverage path planning, (iii) prepare the intersection-line during the seeding process by planting crops in a different direction along the headland and mainfield areas, before (iv) switching manually with one or two sections according to the methods $S_{\textsf{M}_2}^{1}$ or $S_{\textsf{M}_2}^{2}$. An economic evaluation is provided next.

\subsection{Economic cost discusstion\label{subsec_cost}}

Let the spray volume difference between a 1- or 2-sections solution and a multi-sections solution (ASC) be denoted by
\begin{equation}
\Delta S_{\textsf{M}_k}^{j,\text{ASC}}=S_{\textsf{M}_k}^{j}-S_{\textsf{M}_k}^{48},~\forall k\in\{1,2\},~\forall j\in\{1,2\}.\label{eq_dS}
\end{equation}

Then, assuming a water ratio of 99\% for the spray mixture, the corresponding cost difference is
\begin{equation}
\Delta C_{\textsf{M}_k}^{j,\text{ASC}}=(0.01 \cdot C_\text{chemical}+0.99 \cdot C_\text{water})\Delta S_{\textsf{M}_k}^{j,\text{ASC}},\label{eq_dC}
\end{equation}
$\forall k\in\{1,2\},~\forall j\in\{1,2\}$.

Let the hectares-normalised values for \eqref{eq_dS} and \eqref{eq_dC} be denoted by $\Delta \bar{S}_{\textsf{M}_k}^{j,\text{ASC}}$ and $\Delta \bar{C}_{\textsf{M}_k}^{j,\text{ASC}}$ and measured in ($\ell$/ha) and (EUR/ha), respectively. Let the purchase cost difference in EUR between a ASC-machinery and a simpler 1- or 2-sections alternative be denoted as $\Delta K_\text{ASC}$. Then, two economic cost discussions are given.

\begin{figure}
\captionsetup[subfigure]{labelformat=empty}
\centering
  \includegraphics[width=0.99\linewidth]{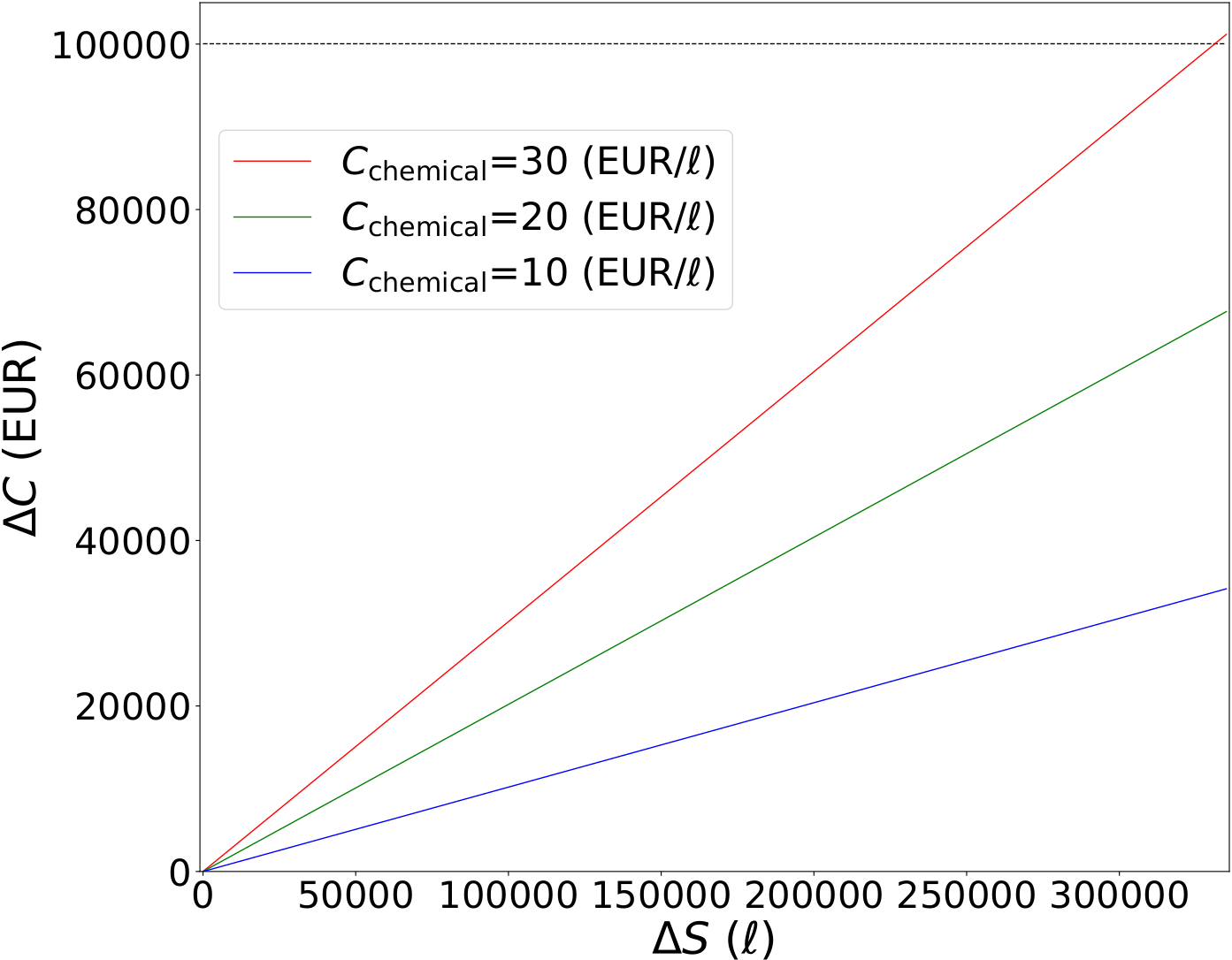}\\[2pt]
  \includegraphics[width=0.99\linewidth]{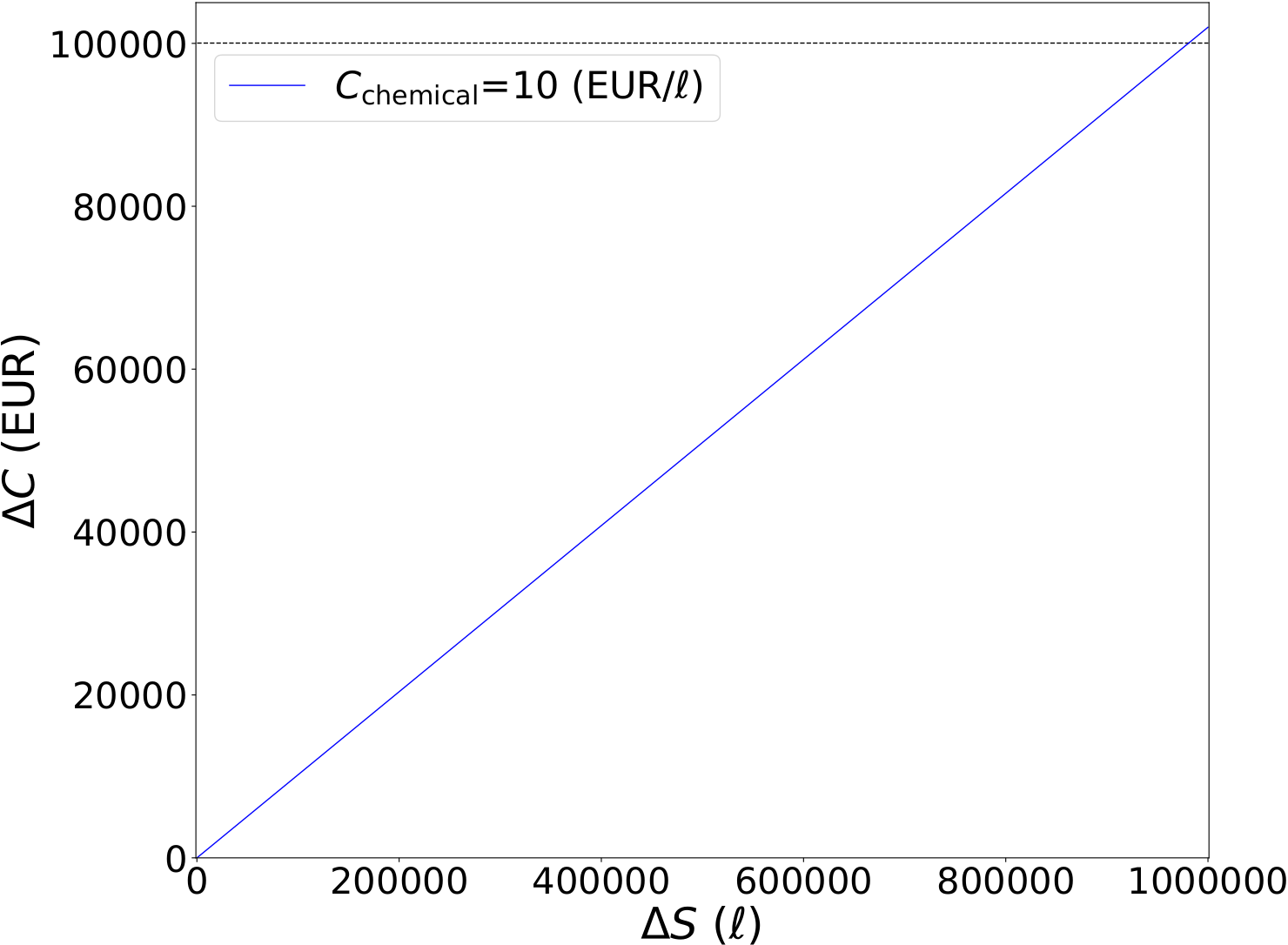}
\caption{Visualisation of \eqref{eq_dC} as a function of spray volume difference $\Delta S$. $C_\text{water}=0.002$EUR/$\ell$ is assumed. The influence of three cost-levels for $C_\text{chemical}$ is compared in the top subplot. Results are highlighted for $C_\text{chemical}=10$EUR/$\ell$ in the bottom subplot.}
\label{fig_cost}
\end{figure}

First, ASC promises to save spray volume. ASC is profitable in comparison to a simpler 1- or 2-sections solution if 
\begin{equation}
\Delta K_\text{ASC} \leq \Delta \bar{C}_{\textsf{M}_k}^{j,\text{ASC}}.\label{eq_Kineq}
\end{equation} 
Evaluating \eqref{eq_dC} and assuming $C_\text{water}=0.002$EUR/$\ell$, the spray volume difference at which ASC first becomes profitable is \textbf{331k $\ell$} for $C_\text{chemical}=30$EUR/$\ell$ and \textbf{981k $\ell$} for $C_\text{chemical}=10$EUR/$\ell$, respectively.
See Fig. \ref{fig_cost} for illustration. For reference, for the first field example of size 6ha the spray volume difference is $\Delta S_{\textsf{M}_2}^{1,\text{ASC}}=(89.9-6.9)~\ell=83~\ell$. If a farm consisted of just this field and assuming $C_\text{chemical}=30$EUR/$\ell$, this field would require 3988 field runs before ASC becomes profitable. For a lower $C_\text{chemical}=10$EUR/$\ell$, a remarkable 11819 field runs would be required. For a larger $\Delta K_\text{ASC}$ greater than 100k EUR (which is not uncommon for modern sprayers) even more field runs would be required before (an optimal) ASC becomes profitable.

Second, an alternative to evaluate the economic potential of ASC can be provided by the following second inequality,
\begin{equation}
\Delta K_\text{ASC} \leq \Delta \bar{C}_{\textsf{M}_k}^{j,\text{ASC}} A_\text{total} N_\text{runs}^\text{field} N_\text{years} ,\label{eq_Nyears}
\end{equation}
where $A_\text{total}$ denotes total farm-wide field area in hectares, $N_\text{runs}^\text{field}$ the number of field runs per year required for spraying, and $N_\text{years}$ a number of years, respectively. The reformulation of \eqref{eq_Nyears} permits to calculate after how many years the spray volume savings of an ASC-solution economically outweigh a simpler 1- or 2-sections alternative. The number of years at which equality is attained in \eqref{eq_Nyears} shall be denoted by $N_\text{years}^\star$.

To evaluate \eqref{eq_Nyears} multiple parameters have to be chosen. $\Delta \bar{C}_{\textsf{M}_k}^{j,\text{ASC}}$ is computed based on data, by (i) computing the average $\Delta \bar{S}_{\textsf{M}_k}^{j,\text{ASC}}$ from the 10 field examples according to Table \ref{tab_dSnormalized} before (ii) computing its corresponding cost difference $\Delta \bar{C}_{\textsf{M}_k}^{j,\text{ASC}}$ according to \eqref{eq_dC}. Values for $C_\text{water}$, $C_\text{chemical}$, 
$A_\text{total}$ and $N_\text{runs}^\text{field}$ are selected according to Table \ref{tab_param}. Results for the evaluation of \eqref{eq_Nyears} are presented in Table \ref{tab_results}.

Several observations can be made. First, for small farms (e.g., $A_\text{total}=30$ha) and for the given parameters, ASC does economically never outweigh a simpler 1- or 2-sections alternative. This is since only after \emph{decades} the spray volume savings of an ASC-solution economically outweigh a simpler 1- or 2-sections alternative. For example, for $A_\text{total}=30$ha, 
$\Delta \bar{S}_{\textsf{M}_1,\ell/\text{ha}}^{1,\text{48}}$=18.6$\ell$/ha, $C_\text{chemical}=30$EUR/$\ell$ and $\Delta K_\text{ASC}=100000$EUR, this occurs only after 74.2 years. For $C_\text{chemical}=10$EUR/$\ell$ even after only 219.7 years.

Second, for larger farm sizes of $A_\text{total}=1000$ha the situation changes. For example, here it is $N_\text{years}^{\star,30}=2.2$ years.

Overall, Table \ref{tab_results} demonstrates that different values for $A_\text{total}$, $C_\text{chemical}$ and $\Delta K_\text{ASC}$ largely shape after how many years the spray volume savings of an ASC-solution economically outweigh a simpler 1- or 2-sections alternative. The smaller the farm, the smaller the cost for chemicals and the larger $\Delta K_\text{ASC}$, the less ASC is economically worth it.

\begin{table}
\centering
\begin{tabular}{|c|r|rr|rr|}
\hline 
\rowcolor[gray]{0.8} & & & & & \\[-8pt]
\rowcolor[gray]{0.8} Ex. & $A_\text{field}$  & $\Delta S_{\textsf{M}_1,\ell/\text{ha}}^{1,\text{ASC}}$ & $\Delta S_{\textsf{M}_1,\ell/\text{ha}}^{2,\text{ASC}}$  & $\Delta S_{\textsf{M}_2,\ell/\text{ha}}^{1,\text{ASC}}$ & $\Delta S_{\textsf{M}_2,\ell/\text{ha}}^{2,\text{ASC}}$ \\[2pt]
\rowcolor[gray]{0.8}  & (ha)  & ($\ell$/ha) & ($\ell$/ha) & ($\ell$/ha) & ($\ell$/ha) \\[2pt]
\hline
& & & & &  \\[-8pt]
1 & 6.0  & 11.7 & 8.9 & 13.8 & 10.5 \\\hline 
2 & 14.6  & 12.7 & 10.5 & 14.0 & 10.9  \\\hline 
3 & 10.1   & 18.7 & 15.8 & 19.0 & 13.5  \\\hline  
4 & 11.8  & 12.0 & 8.1  & 13.5 & 8.8 \\\hline   
5 & 13.5  & 10.5 & 8.9  & 8.6 & 6.9  \\\hline  
6 & 38.5  & 19.8 & 19.0  & 19.9 & 19.0 \\\hline 
7 & 6.5   & 26.9 & 26.5  & 27.8 & 26.9 \\\hline 
8 & 13.6   & 22.9 & 19.8 & 30.9 & 22.6 \\\hline 
9 & 4.5   & 38.4 & 32.6 & 56.3 & 50.0 \\\hline 
10 & 7.2   & 12.7 & 9.4 & 12.9 & 9.7 \\\hline
\textbf{Avg.} & \textbf{12.6}  & \textbf{18.6} & \textbf{16.7} & \textbf{22.5} & \textbf{18.7}  \\
%
\hline
\end{tabular}
\caption{The hectares-normalised spray volumes that a 1- or 2-sections solution requires more than ASC. Data is deduced from Table \ref{tab_s}.}
\label{tab_dSnormalized}
\end{table}

\begin{table}
\centering
\begin{tabular}{|l|l|l|}
\hline 
\rowcolor[gray]{0.8} & &  \\[-8pt]
\rowcolor[gray]{0.8} Parameter & Value & Unit \\[2pt]
\hline
& &  \\[-8pt]
$C_\text{water}$ & 0.002 & EUR/$\ell$ \\
$C_\text{chemical}$ & $\in\{10,30\}$ & EUR/$\ell$ \\
$A_\text{total}$ & $\in\{30,100,300,600,1000\}$ & ha \\
$N_\text{runs}^\text{field}$ & 8 & - \\
\hline
\end{tabular}
\caption{Parameters used for economic cost calculations.}
\label{tab_param}
\end{table}

%
\begin{table}
\centering
\begin{tabular}{|l|l|l|}
\hline 
\rowcolor[gray]{0.9} \multicolumn{3}{|c|}{} \\[-8pt]
\rowcolor[gray]{0.9} \multicolumn{3}{|l|}{For $\Delta \bar{S}_{\textsf{M}_1,\ell/\text{ha}}^{1,\text{48}}$=18.6$\ell$/ha } \\[2pt]
\rowcolor[gray]{0.8} \multicolumn{3}{|c|}{} \\[-8pt]
\rowcolor[gray]{0.8} \multicolumn{3}{|c|}{$\Delta K_\text{ASC}=100000$EUR} \\[2pt]
\hline
& &  \\[-8pt]
$A_\text{total}$ & $N_\text{years}^{\star,30}$ & $N_\text{years}^{\star,10}$ \\[2pt]\hline 
30 & \textbf{74.2} & \textbf{219.7} \\
100 & 22.3 & 65.9 \\
300 & 7.4 & 22.0 \\
600 & 3.7 & 11.0 \\
1000 & \textbf{2.2} & 6.6 \\
%
\rowcolor[gray]{0.8} \multicolumn{3}{|c|}{} \\[-8pt]
\rowcolor[gray]{0.8} \multicolumn{3}{|c|}{$\Delta K_\text{ASC}=200000$EUR} \\[2pt]
\hline
& &  \\[-8pt]
$A_\text{total}$ & $N_\text{years}^{\star,30}$  & $N_\text{years}^{\star,10}$ \\[2pt]\hline 
30 & 148.4 & 439.3 \\
100 & 44.5 & 131.8 \\
300 & 14.8 & 43.9 \\
600 & 7.4 & 22.0 \\
1000 & 4.5 & \textbf{13.2} \\
%
\hline
\rowcolor[gray]{1.0} \multicolumn{3}{c}{} \\[-4pt]
\hline
\rowcolor[gray]{0.9} \multicolumn{3}{|c|}{} \\[-8pt]
\rowcolor[gray]{0.9} \multicolumn{3}{|l|}{For $\Delta \bar{S}_{\textsf{M}_1,\ell/\text{ha}}^{2,\text{48}}$=16.7$\ell$/ha } \\[2pt]
\rowcolor[gray]{0.8} \multicolumn{3}{|c|}{} \\[-8pt]
\rowcolor[gray]{0.8} \multicolumn{3}{|c|}{$\Delta K_\text{ASC}=100000$EUR} \\[2pt]
\hline
& &  \\[-8pt]
$A_\text{total}$ & $N_\text{years}^{\star,30}$  & $N_\text{years}^{\star,10}$ \\[2pt]\hline 
30 & \textbf{82.6} & \textbf{244.7} \\
100 & 24.8 & 73.4\\
300 & 8.3 & 24.5 \\
600 & 4.1 & 12.2 \\
1000 & \textbf{2.5} & 7.3 \\
%
\rowcolor[gray]{0.8} \multicolumn{3}{|c|}{} \\[-8pt]
\rowcolor[gray]{0.8} \multicolumn{3}{|c|}{$\Delta K_\text{ASC}=200000$EUR} \\[2pt]
\hline
& &  \\[-8pt]
$A_\text{total}$ & $N_\text{years}^{\star,30}$  & $N_\text{years}^{\star,10}$  \\[2pt]\hline 
30 & 165.2 & 489.3 \\
100 & 49.6 & 146.8 \\
300 & 16.5 & 48.9 \\
600 & 8.3 & 24.5 \\
1000 & 5.0 & \textbf{14.7} \\
%
\hline
\end{tabular}~~~
\begin{tabular}{|l|l|l|}
\hline 
\rowcolor[gray]{0.9} \multicolumn{3}{|c|}{} \\[-8pt]
\rowcolor[gray]{0.9} \multicolumn{3}{|l|}{For $\Delta \bar{S}_{\textsf{M}_2,\ell/\text{ha}}^{1,\text{48}}$=22.5$\ell$/ha } \\[2pt]
\rowcolor[gray]{0.8} \multicolumn{3}{|c|}{} \\[-8pt]
\rowcolor[gray]{0.8} \multicolumn{3}{|c|}{$\Delta K_\text{ASC}=100000$EUR} \\[2pt]
\hline
& &  \\[-8pt]
$A_\text{total}$ & $N_\text{years}^{\star,30}$ & $N_\text{years}^{\star,10}$ \\[2pt]\hline 
30 & \textbf{61.3} & \textbf{181.6} \\
100 & 18.4 & 54.5 \\
300 & 6.1 & 18.2 \\
600 & 3.1 & 9.1 \\
1000 & \textbf{1.8} & 5.4 \\
%
\rowcolor[gray]{0.8} \multicolumn{3}{|c|}{} \\[-8pt]
\rowcolor[gray]{0.8} \multicolumn{3}{|c|}{$\Delta K_\text{ASC}=200000$EUR} \\[2pt]
\hline
& &  \\[-8pt]
$A_\text{total}$ & $N_\text{years}^{\star,30}$ & $N_\text{years}^{\star,10}$ \\[2pt]\hline 
30 & 122.6 & 363.2 \\
100 & 36.8 & 109.0 \\
300 & 12.3 & 36.3 \\
600 & 6.1 & 18.2 \\
1000 & 3.7 & \textbf{10.9} \\
%
\hline
\rowcolor[gray]{1.0} \multicolumn{3}{c}{} \\[-4pt]
\hline
\rowcolor[gray]{0.9} \multicolumn{3}{|c|}{} \\[-8pt]
\rowcolor[gray]{0.9} \multicolumn{3}{|l|}{For $\Delta \bar{S}_{\textsf{M}_2,\ell/\text{ha}}^{2,\text{ASC}}$=18.7$\ell$/ha } \\[2pt]
\rowcolor[gray]{0.8} \multicolumn{3}{|c|}{} \\[-8pt]
\rowcolor[gray]{0.8} \multicolumn{3}{|c|}{$\Delta K_\text{ASC}=100000$EUR} \\[2pt]
\hline
& &  \\[-8pt]
$A_\text{total}$ & $N_\text{years}^{\star,30}$ & $N_\text{years}^{\star,10}$ \\[2pt]\hline 
30 & \textbf{73.8} & \textbf{218.5} \\
100 & 22.1 & 65.5\\
300 & 7.4 & 21.8 \\
600 & 3.7 & 10.9 \\
1000 & \textbf{2.2} & 6.5 \\
%
\rowcolor[gray]{0.8} \multicolumn{3}{|c|}{} \\[-8pt]
\rowcolor[gray]{0.8} \multicolumn{3}{|c|}{$\Delta K_\text{ASC}=200000$EUR} \\[2pt]
\hline
& &  \\[-8pt]
$A_\text{total}$ & $N_\text{years}^{\star,30}$ & $N_\text{years}^{\star,10}$ \\[2pt]\hline 
30 & 147.6 & 437.0 \\
100 & 44.3 & 131.1 \\
300 & 14.8 & 43.7 \\
600 & 7.4 & 21.8 \\
1000 & 4.4 & \textbf{13.1} \\
%
\hline
\end{tabular}
\caption{Numerical results for the evaluation of \eqref{eq_Nyears}. $N_\text{years}^{\star,30}$ implies that cost $C_\text{chemical}=30$EUR/$\ell$ was assumed. For $N_\text{years}^{\star,10}$ it was employed $C_\text{chemical}=10$EUR/$\ell$. Numbers in bold are for emphasis (see Sec. \ref{subsec_cost}).}
\label{tab_results}
\end{table}

Above results are not flattering with respect to the economic profitability of ASC. Therefore, the natural question arises: \textbf{How inaccurate is above parameter-based economic cost discussion?}

The cost of pesticides (including herbicides, fungicides, and insecticides) for cereal crop production can vary widely based on several factors, including the type of persticide used, regional market conditions, the size of the farm, and the specific application needs (e.g., pre-emergence, post-emergence products). It’s also important to note that pesticide costs can vary greatly depending on the region, with some countries offering subsidies or differing regulations that might affect price points. Additionally, larger-scale operations might benefit from bulk discounts. Thus, overall parameter $C_\text{chemical}$ is difficult to estimate since it can regionally greatly vary. Above provided two inequalities, \eqref{eq_Kineq} and \eqref{eq_Nyears}, permit a reader and practitioner to quickly evaluate their own cost models for $C_\text{chemical}$, and similarly for $C_\text{water}$. In contrast to $C_\text{chemical}$, $A_\text{total}$ and $N_\text{runs}^\text{field}$ are well measurable.

The water ratio is very influential on results. A water ratio of 99\% was assumed for the spray mixture (see \eqref{eq_dC}). If this changed to 98\% then $N_\text{years}^\star$-results in Table \ref{tab_results} are approximately halved. Nevertheless,  
for $A_\text{total}=30$ha $N_\text{years}^{\star,30}$ would still in all scenarios in the order of multiple decades. Inversely, for a ratio larger than 99\%, $N_\text{years}^\star$ would be increased.

$\Delta \bar{S}_{\textsf{M}_k,\ell/\text{ha}}^{j,\text{ASC}}$ is data-based. It depends on the methods from Sec. \ref{subsec_4hierarchicallevels}-\ref{subsec_strategy2} and field contour data. Parameters include $v_\text{ref}$, $W$, $w$ and $s_\text{volume}^\text{ref}$. Several comments can be made. First, traveling velocities and path are the same for ASC and the proposed 1- and 2-sections alternatives. Therefore nominally, $v_\text{ref}$ has no effect on above relative comparison of the methods. In practice, ASC becomes more difficult for higher velocities. To avoid coverage gaps  (which are more important to avoid then spray overlaps) ASC has to be set more conservatively, which would result in overlaps and reduction in spray volume savings.

Second, the larger $w$ the more ASC converges to the 1- or 2-sections solution. The larger $W$ the less the number of mainfield lanes and number of turns overall. On the other hand, during turns higher velocities at the outer sections along the boom bar.

Third, a change in $s_\text{volume}^\text{ref}$ does not change switching signals but only the magnitude of spray volume. Thus, the coverage maps, e.g. in Fig. \ref{fig_f14} and Fig. \ref{fig_f17}, do not change and thus also the relative difference between ASC and proposed 1- or 2-sections alternative does not change.

Fourth, above cost discussion is linear in parameters. 

Finally, while the purchase price of an ASC-machinery is well measurable, the actual performance is in generally non-linear degraded by the effects listed in Table \ref{tab_influences}. Thus, above economic cost discussion is optimistic with respect to ASC since \emph{nominal} optimal ASC was assumed for simulation experiments. This was done intentionally as outlined in Problem \ref{problem1} to derive a best-case \emph{upper performance bound} for ASC. 

ASC-maintenance and software upgrade costs are not included in the cost analysis and would further degrade the economic potential of ASC, since these costs are expected to be higher than for a 1- or 2-sections sensor-free alternative.

To summarise, two inequalities \eqref{eq_Kineq} and \eqref{eq_Nyears} were used to evaluate the economic potential of ASC in comparison to a simpler 1- or 2-sections alternative. Overall, and assuming selected parameters according to Table \ref{tab_param} and \ref{tab_results}, results strongly suggest that ASC is purely economically speaking (and this is decisive for many small farms) not profitable for at least small farms, and a simpler 1- or 2-sections alternative is recommended. Margins in Fig. \ref{fig_cost} and Table \ref{tab_results} are simply too large to recommend otherwise. As long as costs $C_\text{chemical}$ are low and price difference $\Delta K_\text{ASC}$ high this statement will hold. Above provided two inequalities \eqref{eq_Kineq} and \eqref{eq_Nyears} permit a practitioner to quickly evaluate their own cost models for $C_\text{chemical}$ and $\Delta K_\text{ASC}$.

\subsection{Limitations}

Limitations of the method are discussed. As stressed throughout the paper, \emph{nominal} conditions are assumed to evaluate the \emph{best-case} performance that ASC can deliver. Thus, performance in any real-world conditions are degraded according to the effects listed in Table \ref{tab_influences}. However, the methods $S_{\textsf{M}_2}^{1}$ or $S_{\textsf{M}_2}^{2}$ are naturally also affected by uncertainties and inaccuracies in practice. For example, human driving experience is required to compensate for delays when switching at the intersection line. Nevertheless, the decisive advantages besides the economic benefits of the recommended method are the potential for (i) a sensor-free mechanical implementation, and (ii) robustness of this method.

\section{Conclusion\label{sec_conclusion}}

This paper presented a nominal evaluation of Automatic Multi-Sections Control (ASC) in comparison to a proposed simpler one- or two-sections alternative with predictive spray switching. Four different hierarchical planning levels for area coverage were discussed. These include a path planning, a block switching, a sections switching and an individual nozzle control-level. The focus of this paper was on the first three levels. Two area coverage path planning patterns, the Boustrophedon-pattern and an alternative pattern, were discussed.  Optimised block switching-logics for each of these two patterns were presented. Three different variations on the sections-switching level were compared.

The complexities of ASC were highlighted, including necessity for accurate localisation sensors and automation to individually control a plethora of different nozzles. 

In contrast to ASC, a preferred method is suggested that (i) minimises area coverage path length by employing a technique different from Boustrophedon-based path planning, (ii) offers intermediate overlap, (iii) crucially and in contrast to ASC can be implemented sensor-free (low-cost) by visual inspection of the intersection line resulting when employing different planting directions in the mainfield and headland area, and (iv) is suitable for manual driving by following a pre-planned predictive spray switching logic and area coverage path plan.

The preferred method minimises pathlength, however requires overall more spray volume tha more complex and expensive ASC. A detailed economic cost analysis was provided. Fundamental to this discussion was that spray mixtures consist to a very large degree out of water and only to a small degree, often 0.5-2\%, out of chemical component (herbicides, fungicides and the like), but water costs are typically much cheaper than the cost for chemicals. Surprisingly strong economic arguments were found to not recommend ASC for small farms.

\bibliography{mybibfile.bib}
\nocite{*}








\appendix

\section{Visualisation of results for remaining fields\label{sec_appendix}}

\begin{figure*}
\centering
\begin{subfigure}[t]{.33\linewidth}
  \centering
  \includegraphics[width=.99\linewidth]{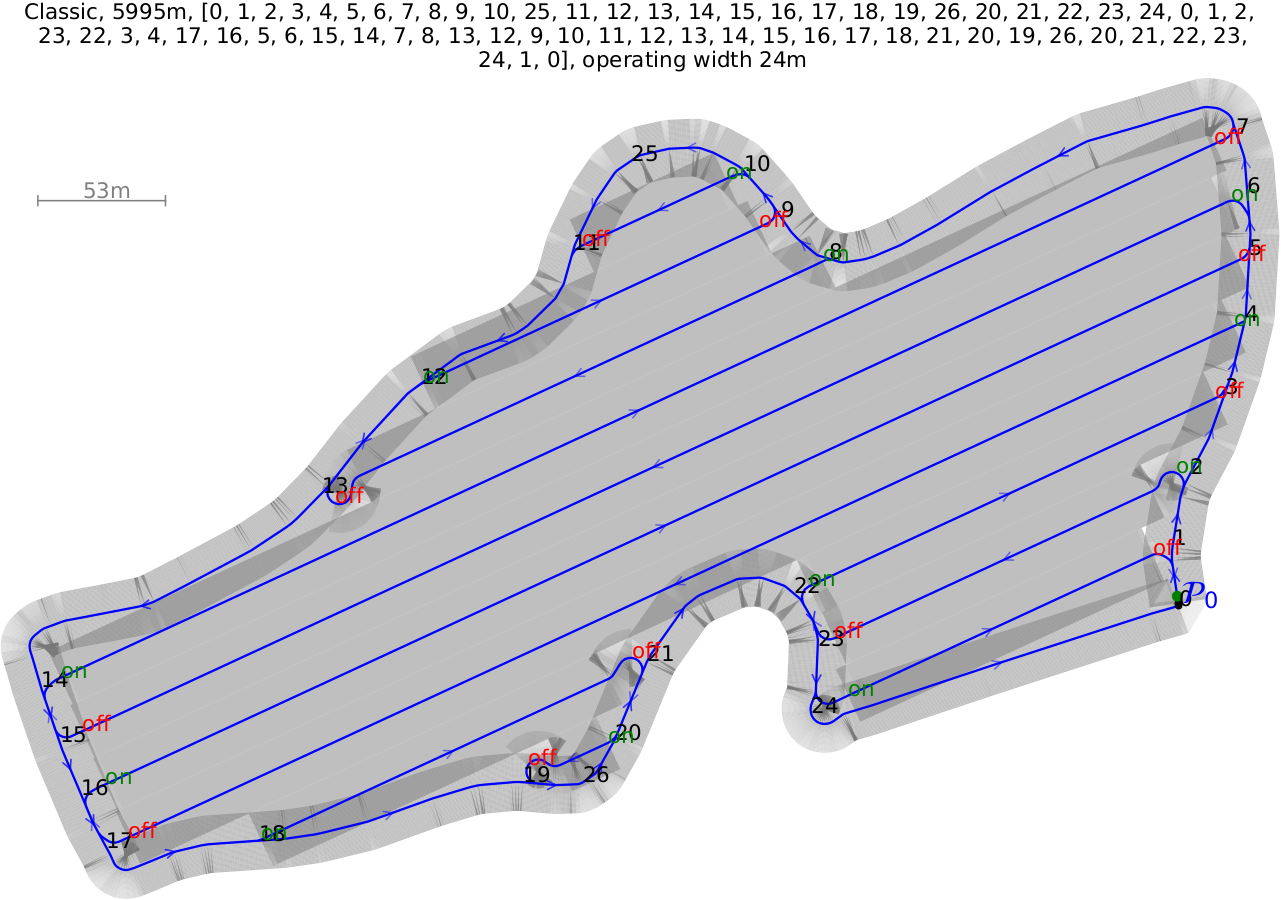}\\[-1pt]
\caption{$S_{\textsf{M}_1}^{1}$ \\[10pt]}
  \label{fig_f16_SC0}
\end{subfigure}
\begin{subfigure}[t]{.33\linewidth}
  \centering
  \includegraphics[width=.99\linewidth]{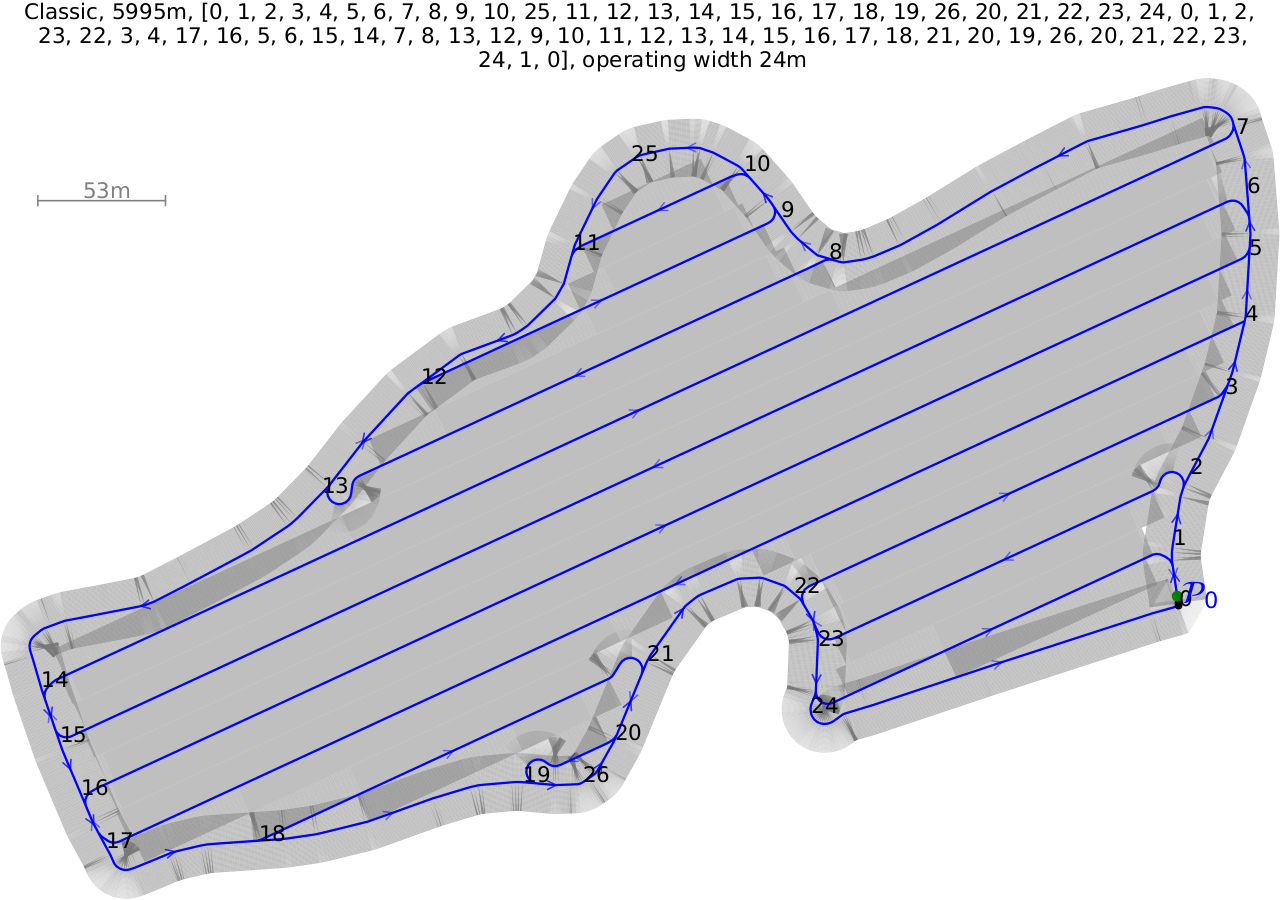}\\[-1pt]
\caption{$S_{\textsf{M}_1}^{2}$\\[10pt]}
  \label{fig_f16_SC1}
\end{subfigure}
\begin{subfigure}[t]{.33\linewidth}
  \centering
  \includegraphics[width=.99\linewidth]{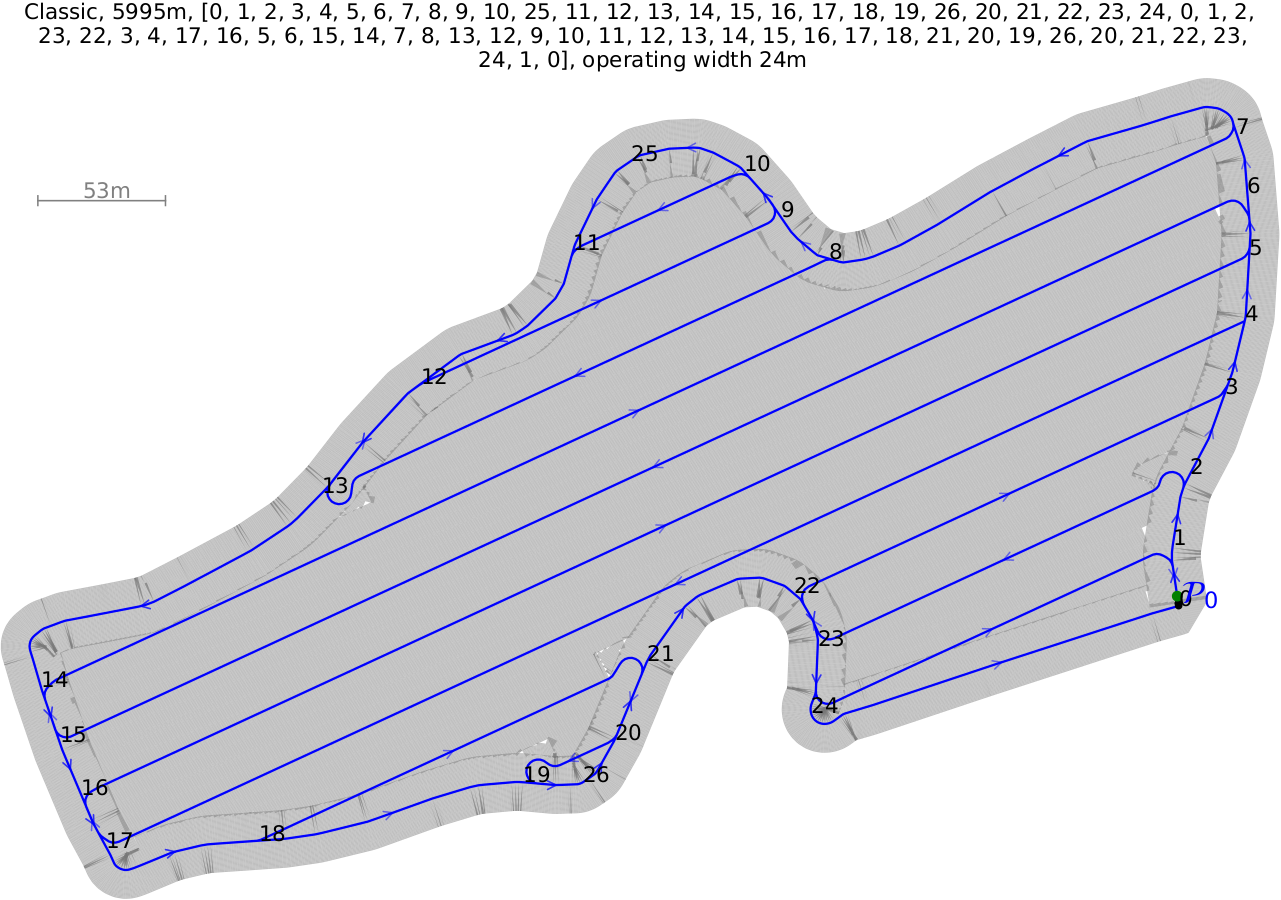}\\[-1pt]
\caption{$S_{\textsf{M}_1}^{48}$\\[10pt]}
  \label{fig_f16_SC2}
\end{subfigure}
\begin{subfigure}[t]{.33\linewidth}
  \centering
  \includegraphics[width=.99\linewidth]{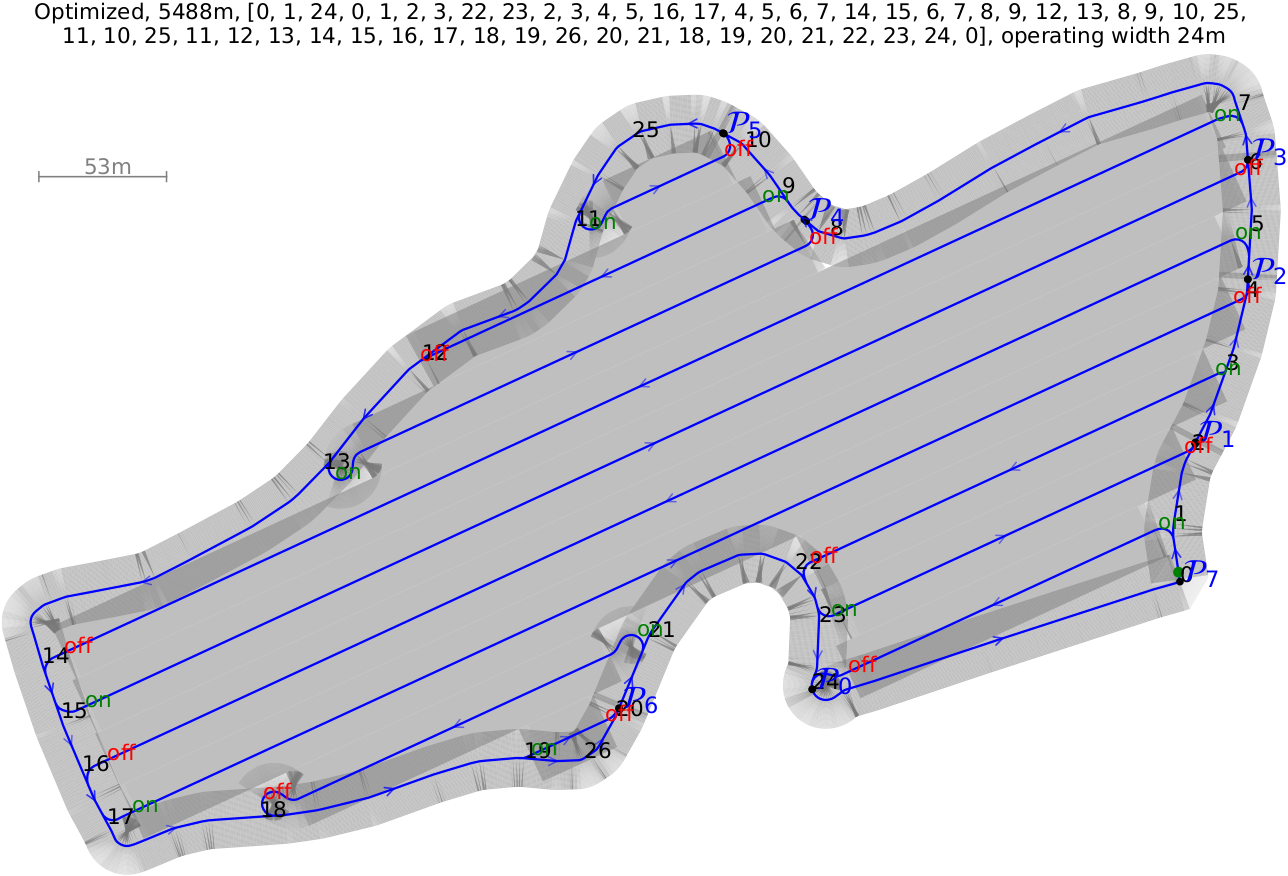}\\[-1pt]
\caption{$S_{\textsf{M}_2}^{1}$\\[10pt]}
  \label{fig_f16_SO0}
\end{subfigure}
\begin{subfigure}[t]{.33\linewidth}
  \centering
  \includegraphics[width=.99\linewidth]{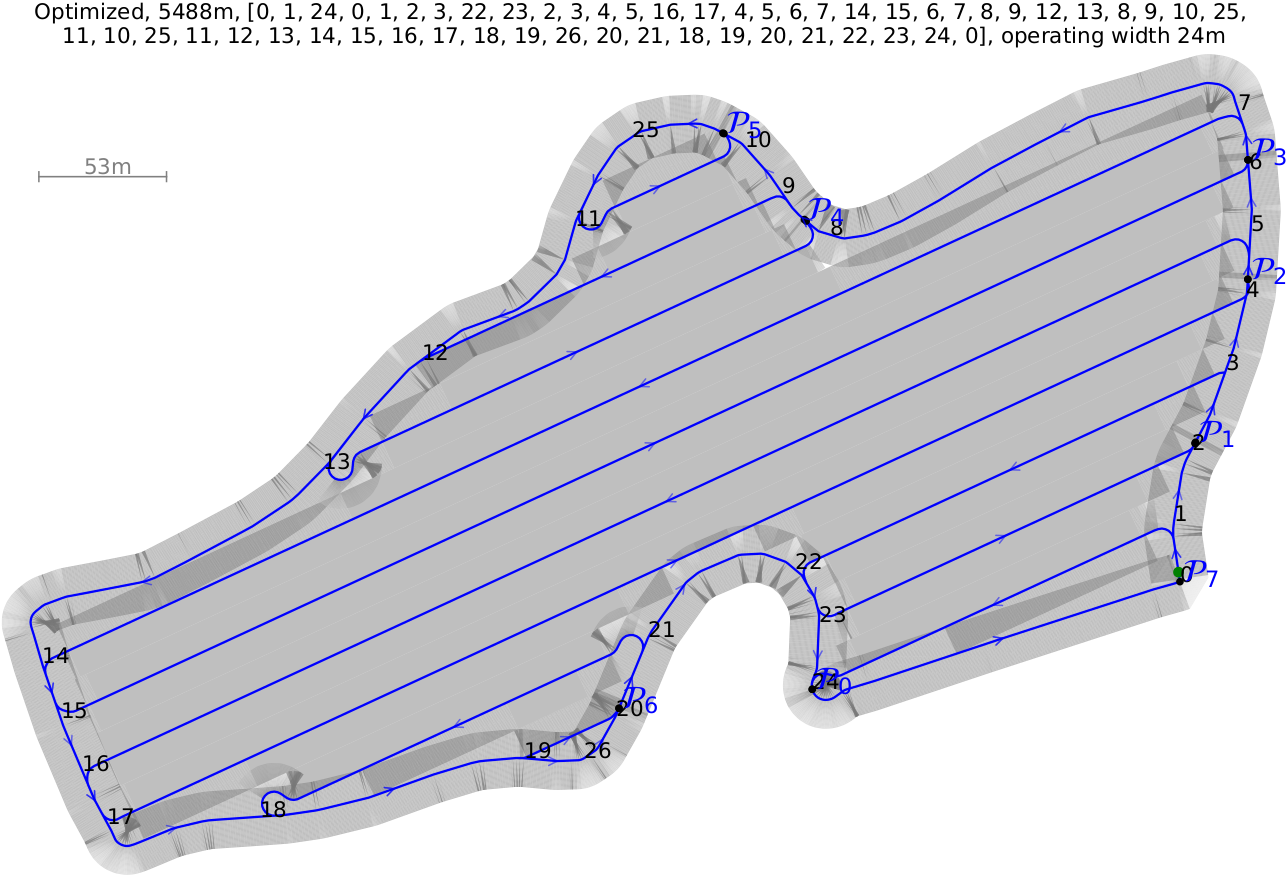}\\[-1pt]
\caption{$S_{\textsf{M}_2}^{2}$\\[10pt]}
  \label{fig_f16_SO1}
\end{subfigure}
\begin{subfigure}[t]{.33\linewidth}
  \centering
  \includegraphics[width=.99\linewidth]{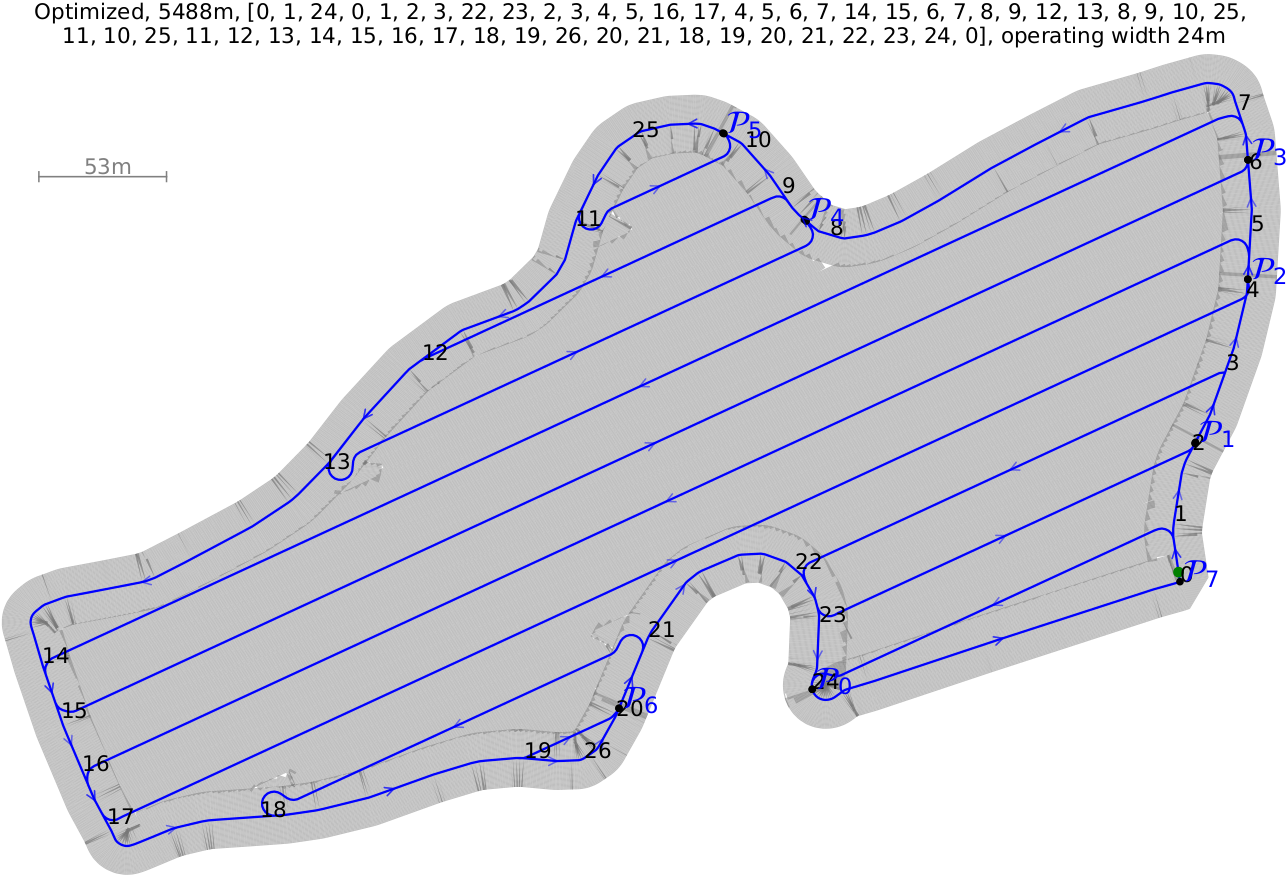}\\[-1pt]
\caption{$S_{\textsf{M}_2}^{48}$\\[10pt]}
  \label{fig_f16_SO2}
\end{subfigure}
\caption{Field example 3: visual comparison of 6 different setups. See Table \ref{tab_s} for quantitative evaluation. }
\label{fig_f16}
\end{figure*}

\begin{figure*}
\centering
\begin{subfigure}[t]{.33\linewidth}
  \centering
  \includegraphics[width=.99\linewidth]{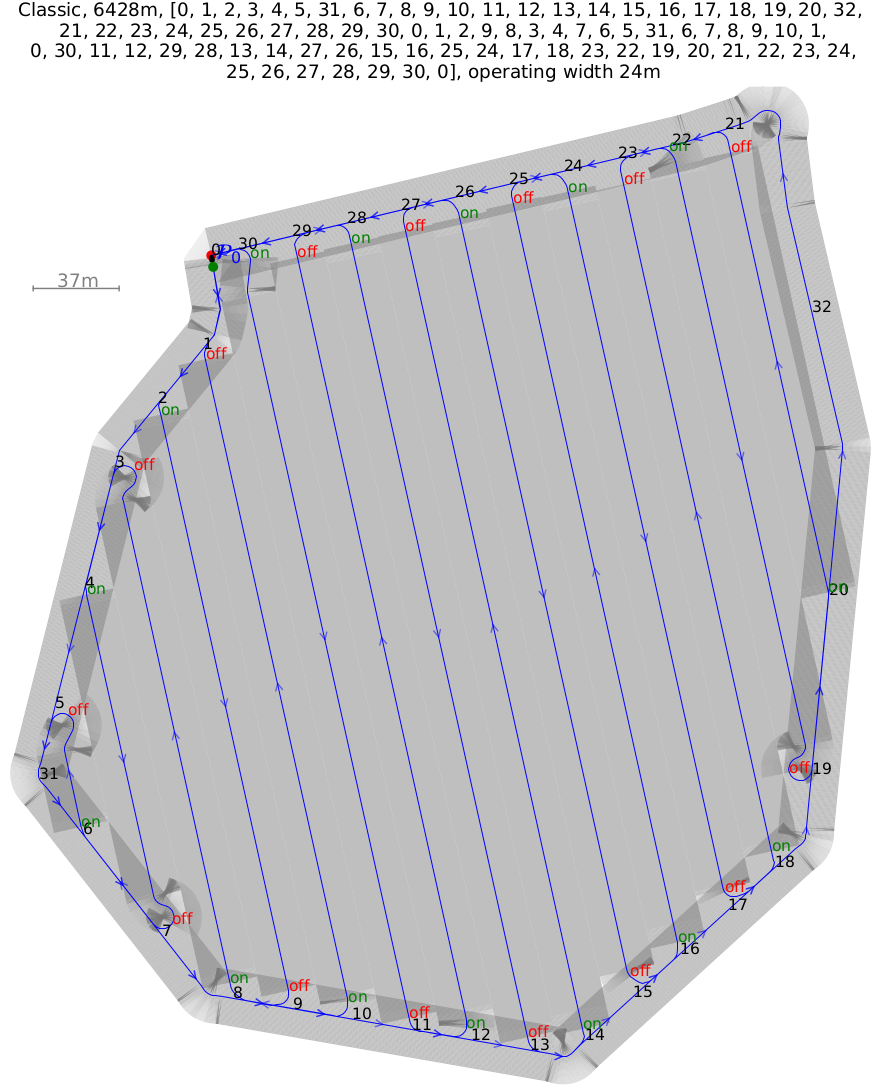}\\[-1pt]
\caption{$S_{\textsf{M}_1}^{1}$ \\[5pt]}
  \label{fig_f6_SC0}
\end{subfigure}
\begin{subfigure}[t]{.33\linewidth}
  \centering
  \includegraphics[width=.99\linewidth]{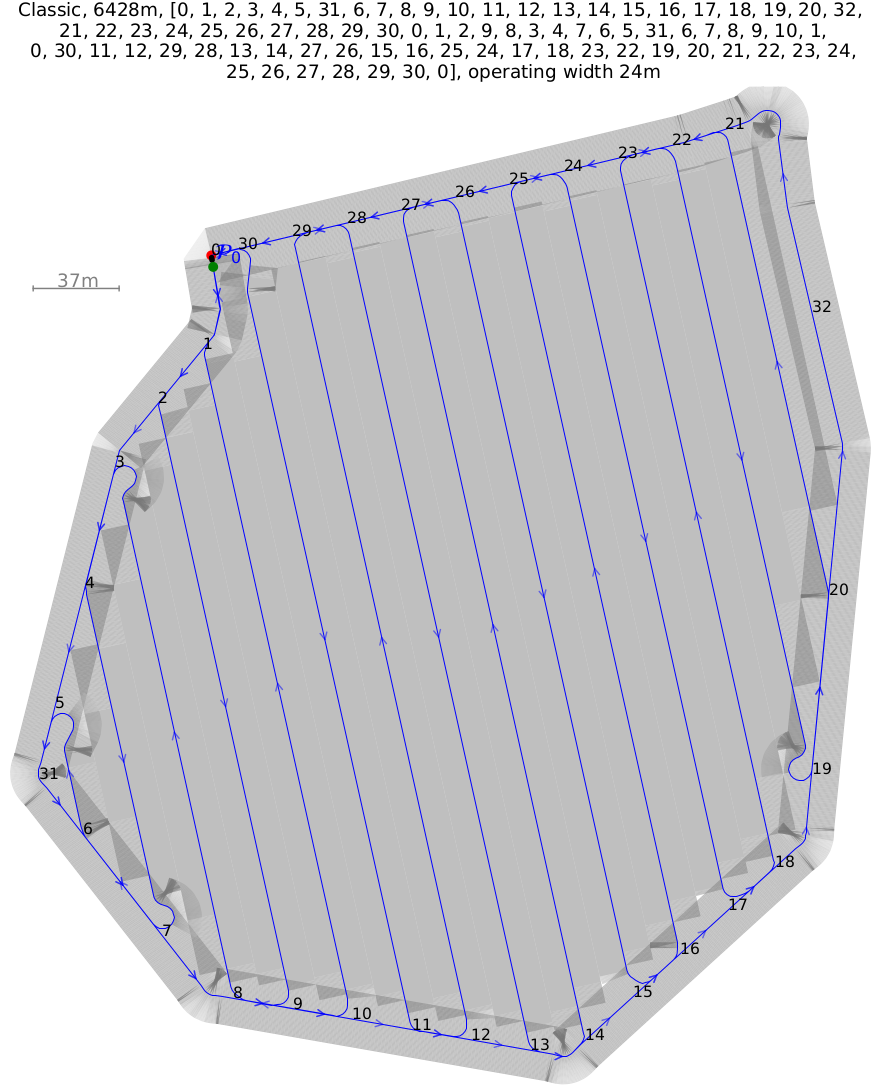}\\[-1pt]
\caption{$S_{\textsf{M}_1}^{2}$\\[5pt]}
  \label{fig_f6_SC1}
\end{subfigure}
\begin{subfigure}[t]{.33\linewidth}
  \centering
  \includegraphics[width=.99\linewidth]{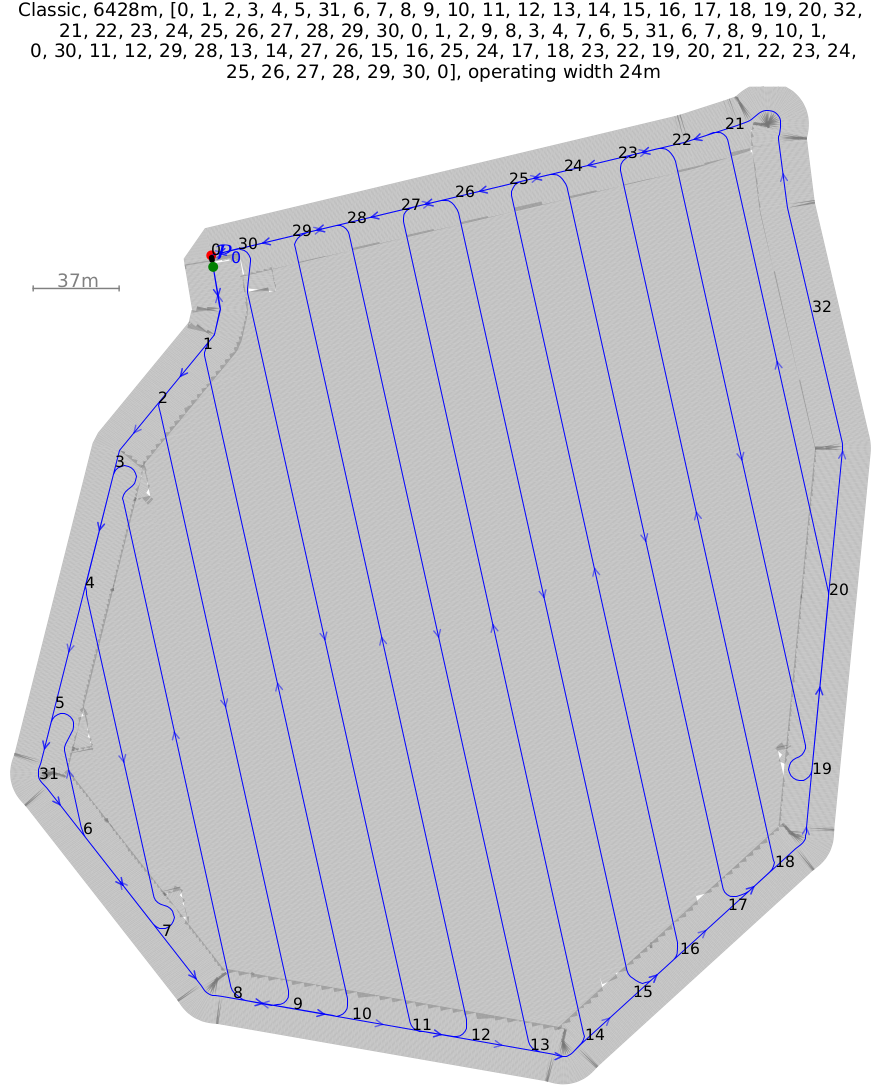}\\[-1pt]
\caption{$S_{\textsf{M}_1}^{48}$\\[5pt]}
  \label{fig_f6_SC2}
\end{subfigure}
\begin{subfigure}[t]{.33\linewidth}
  \centering
  \includegraphics[width=.99\linewidth]{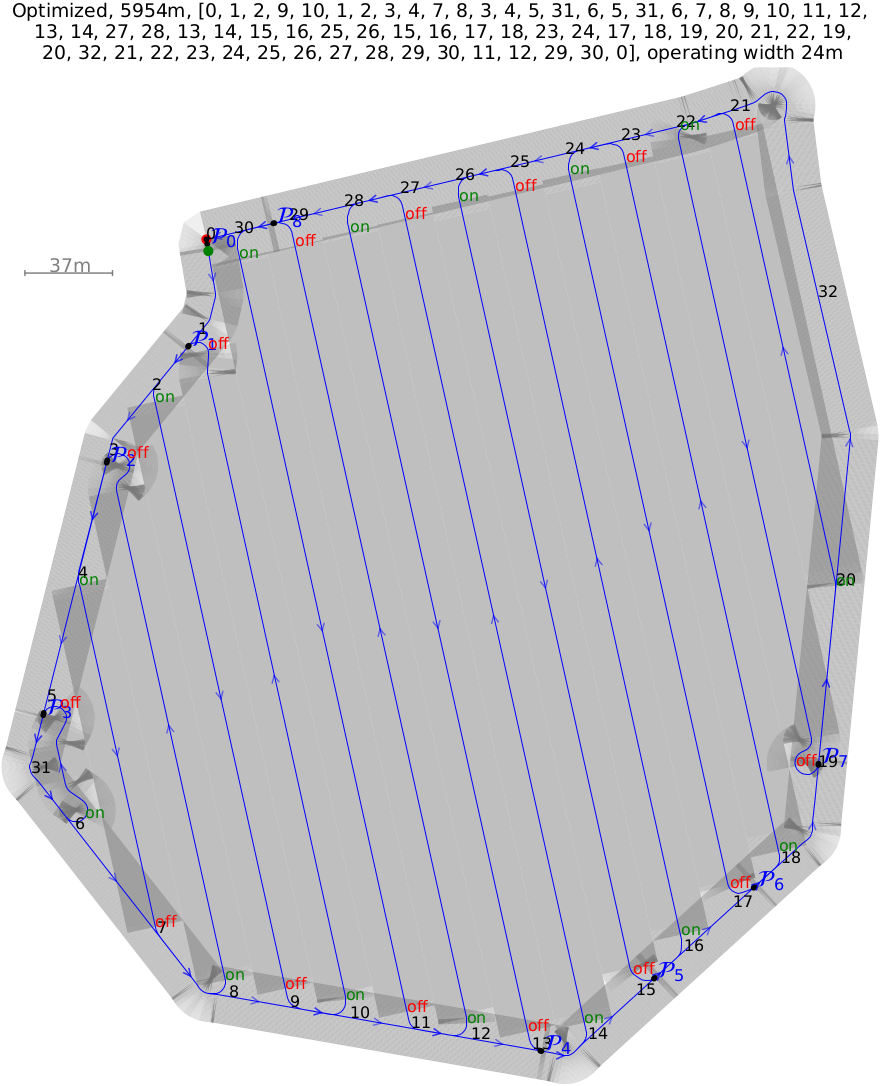}\\[-1pt]
\caption{$S_{\textsf{M}_2}^{1}$}
  \label{fig_f6_SO0}
\end{subfigure}
\begin{subfigure}[t]{.33\linewidth}
  \centering
  \includegraphics[width=.99\linewidth]{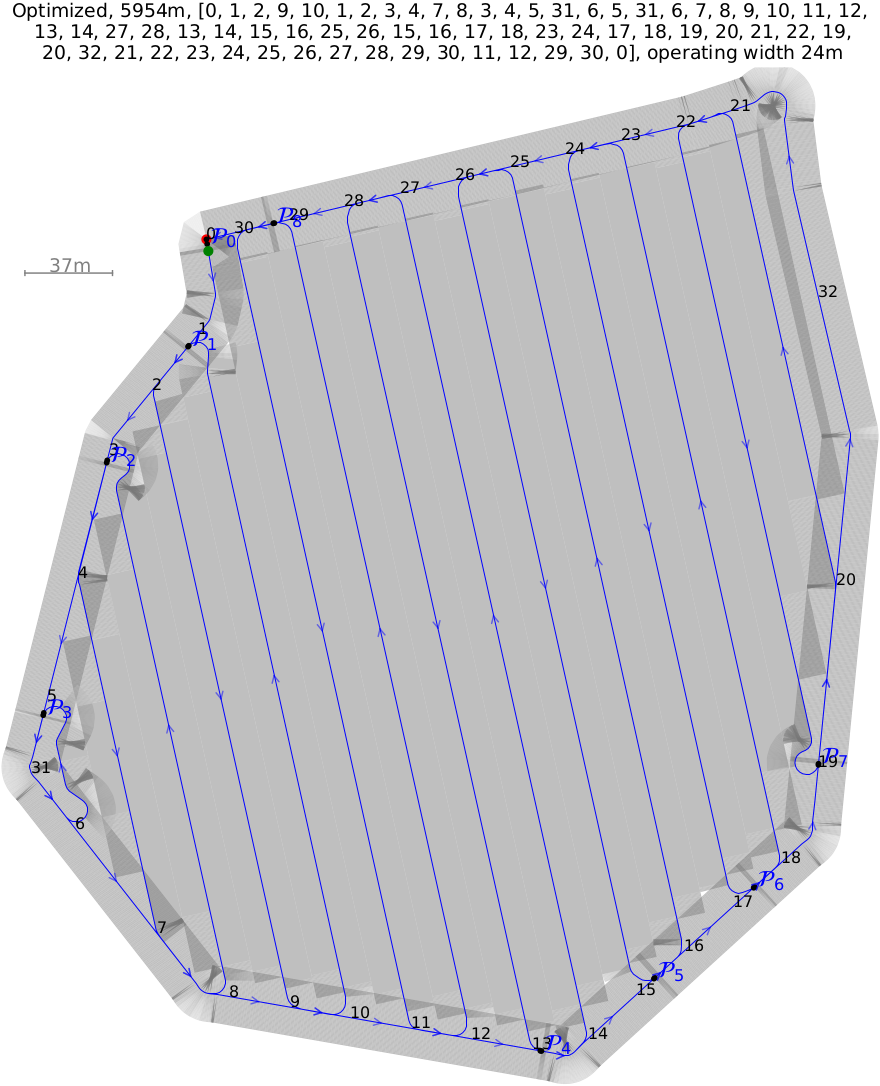}\\[-1pt]
\caption{$S_{\textsf{M}_2}^{2}$}
  \label{fig_f6_SO1}
\end{subfigure}
\begin{subfigure}[t]{.33\linewidth}
  \centering
  \includegraphics[width=.99\linewidth]{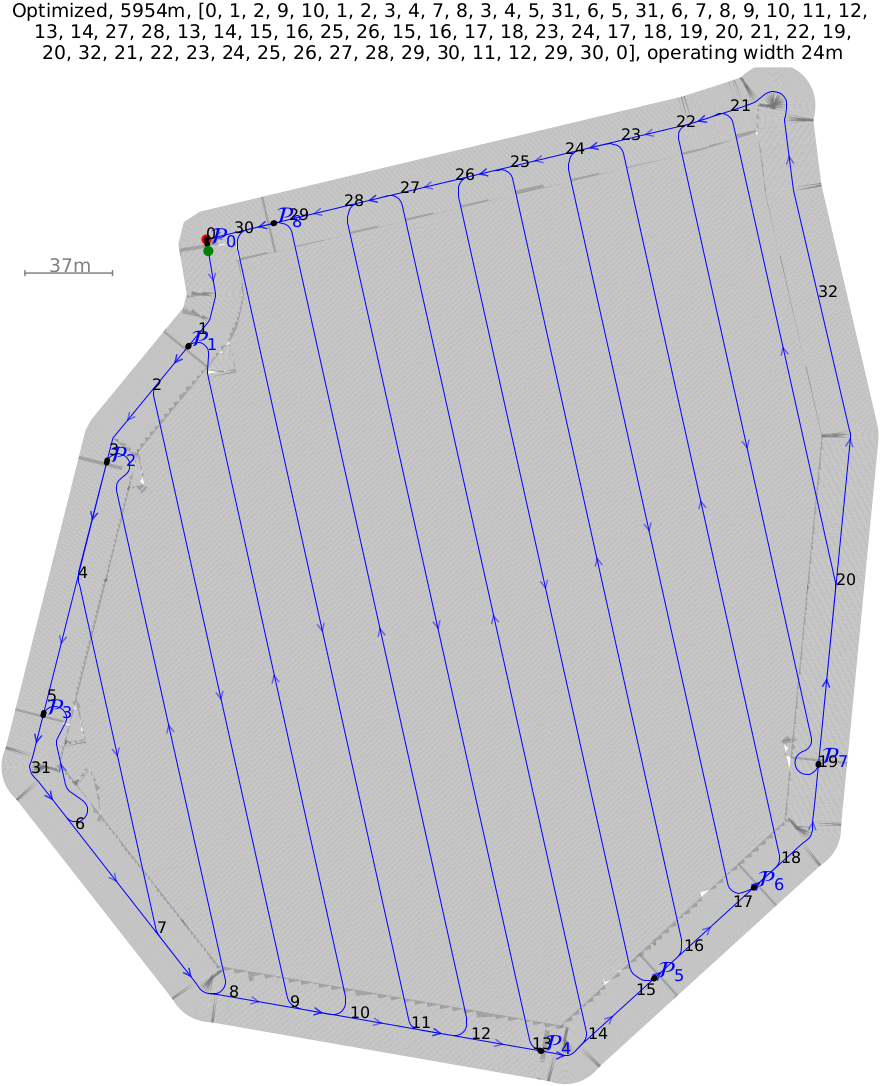}\\[-1pt]
\caption{$S_{\textsf{M}_2}^{48}$}
  \label{fig_f6_SO2}
\end{subfigure}
\caption{Field example 4: visual comparison of 6 different setups. See Table \ref{tab_s} for quantitative evaluation. }
\label{fig_f6}
\end{figure*}

\begin{figure*}
\centering
\begin{subfigure}[t]{.33\linewidth}
  \centering
  \includegraphics[width=.99\linewidth]{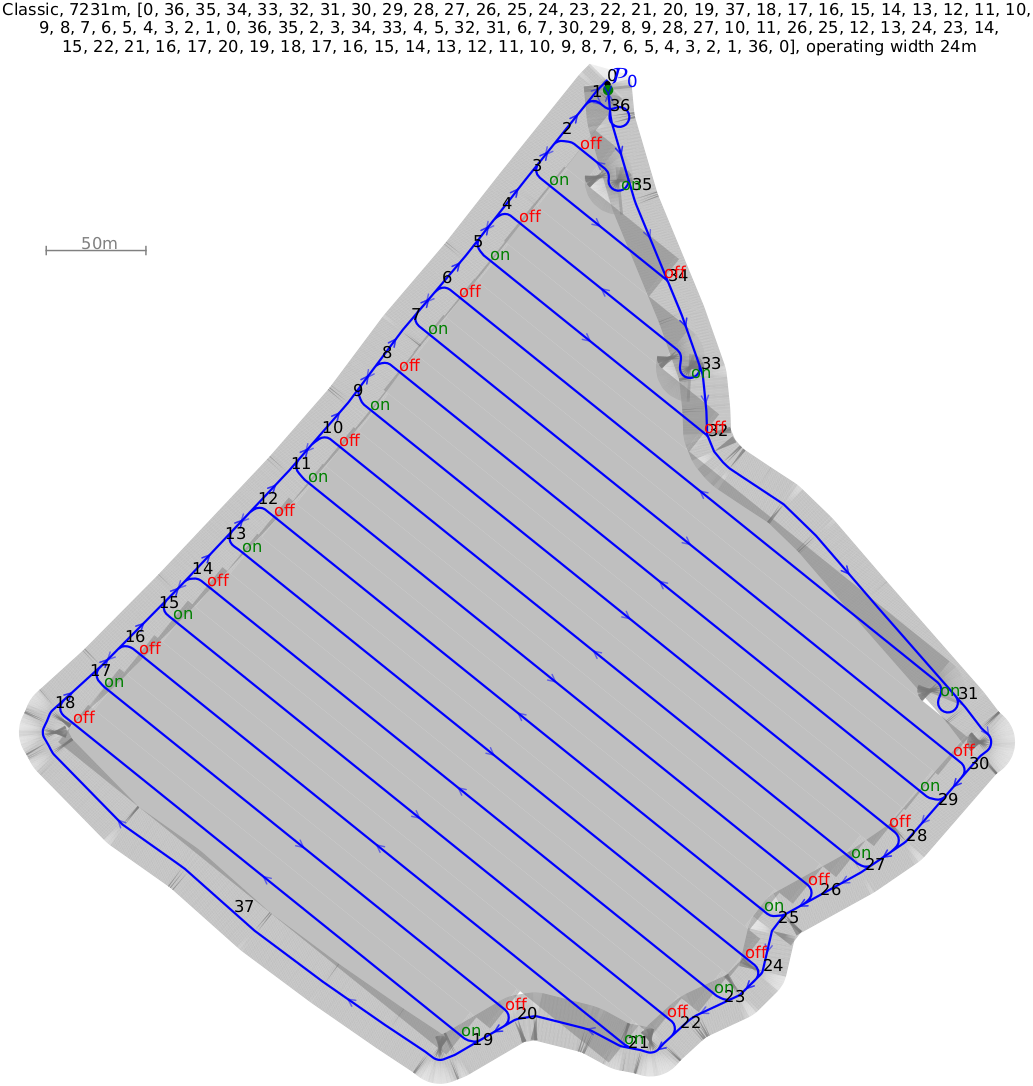}\\[-1pt]
\caption{$S_{\textsf{M}_1}^{1}$ \\[10pt]}
  \label{fig_f7_SC0}
\end{subfigure}
\begin{subfigure}[t]{.33\linewidth}
  \centering
  \includegraphics[width=.99\linewidth]{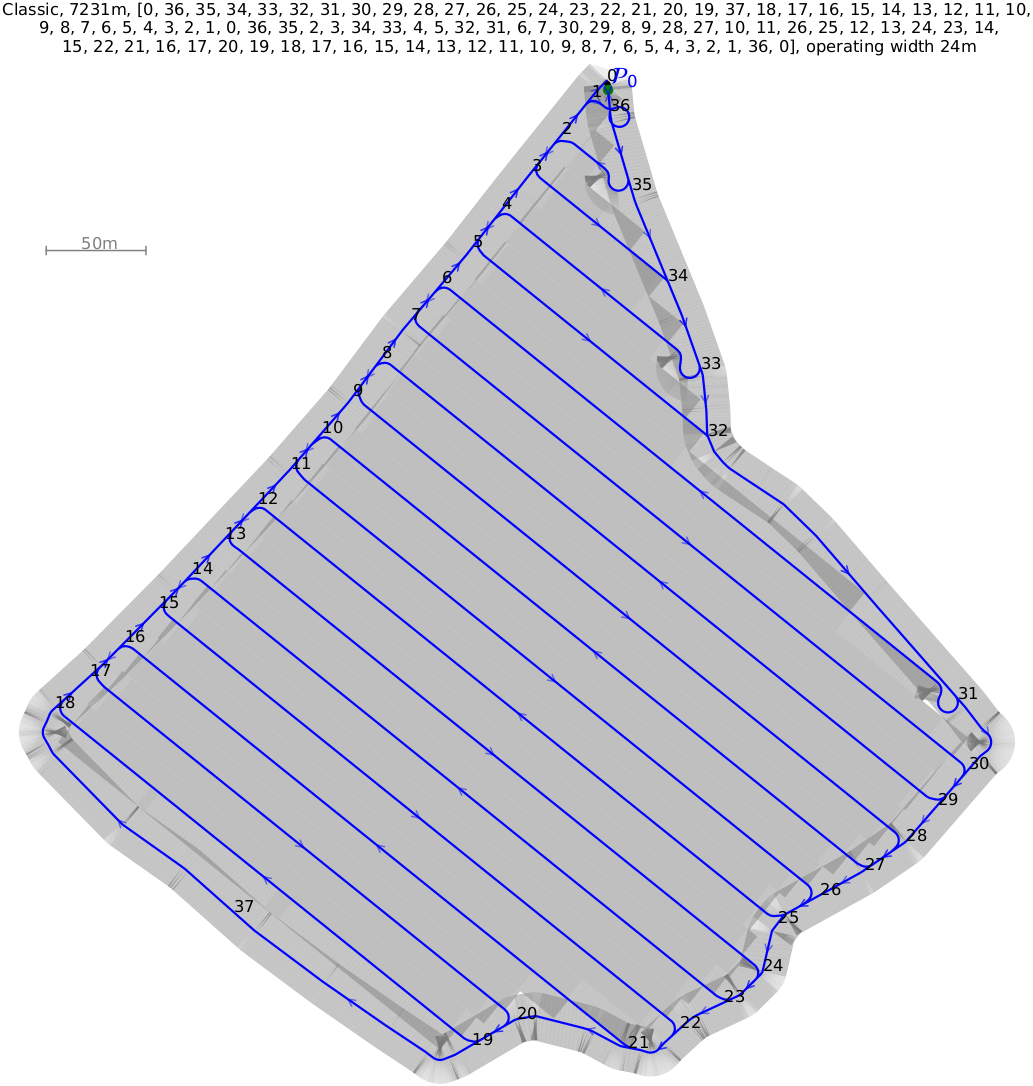}\\[-1pt]
\caption{$S_{\textsf{M}_1}^{2}$\\[10pt]}
  \label{fig_f7_SC1}
\end{subfigure}
\begin{subfigure}[t]{.33\linewidth}
  \centering
  \includegraphics[width=.99\linewidth]{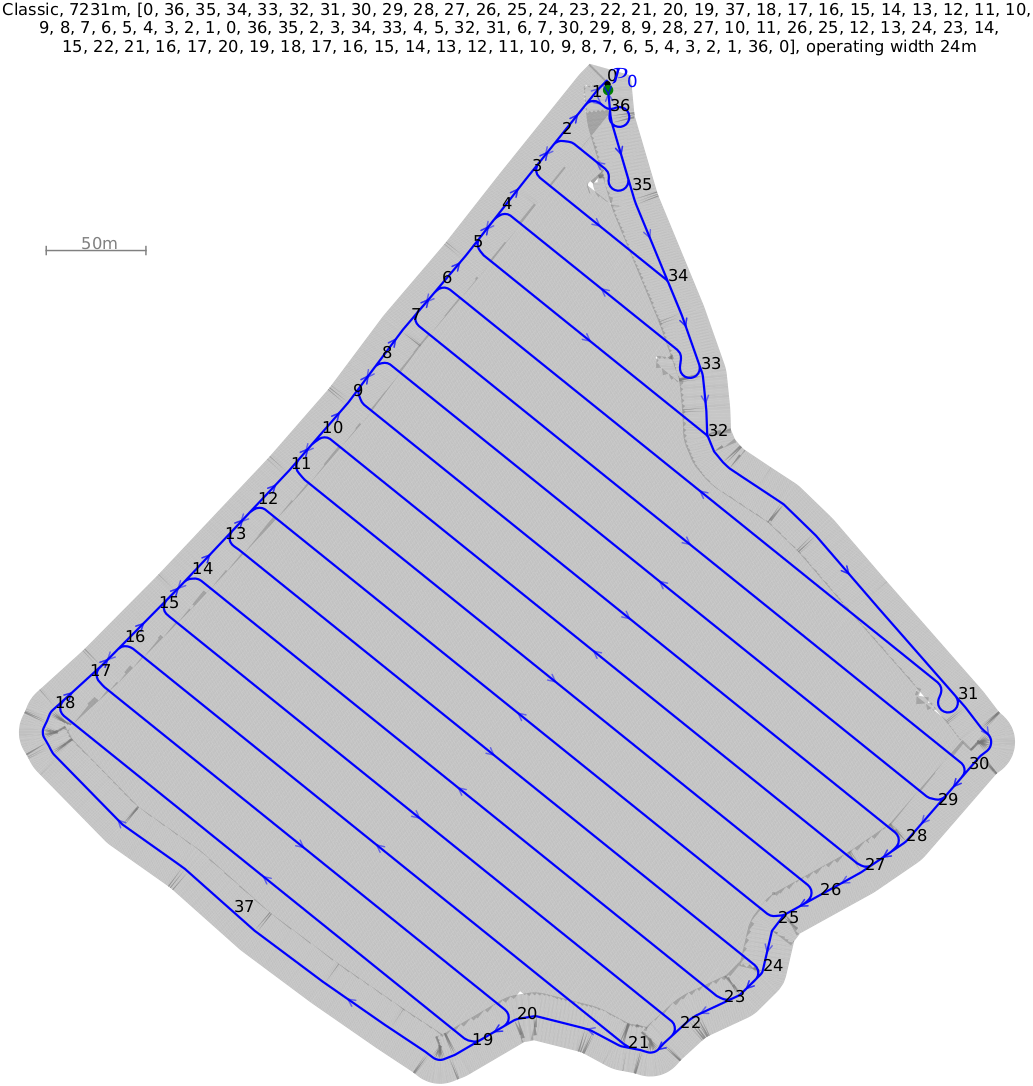}\\[-1pt]
\caption{$S_{\textsf{M}_1}^{48}$\\[10pt]}
  \label{fig_f7_SC2}
\end{subfigure}
\begin{subfigure}[t]{.33\linewidth}
  \centering
  \includegraphics[width=.99\linewidth]{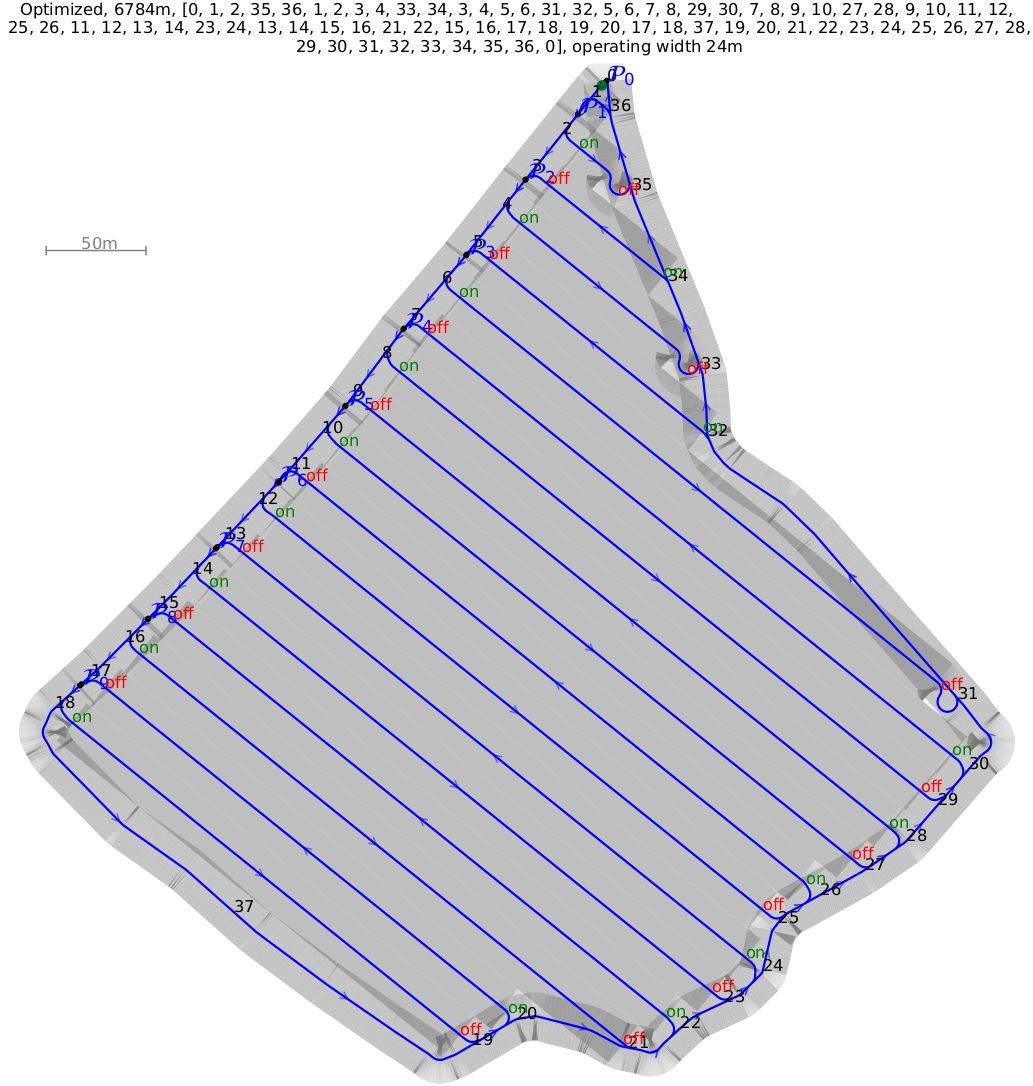}\\[-1pt]
\caption{$S_{\textsf{M}_2}^{1}$}
  \label{fig_f7_SO0}
\end{subfigure}
\begin{subfigure}[t]{.33\linewidth}
  \centering
  \includegraphics[width=.99\linewidth]{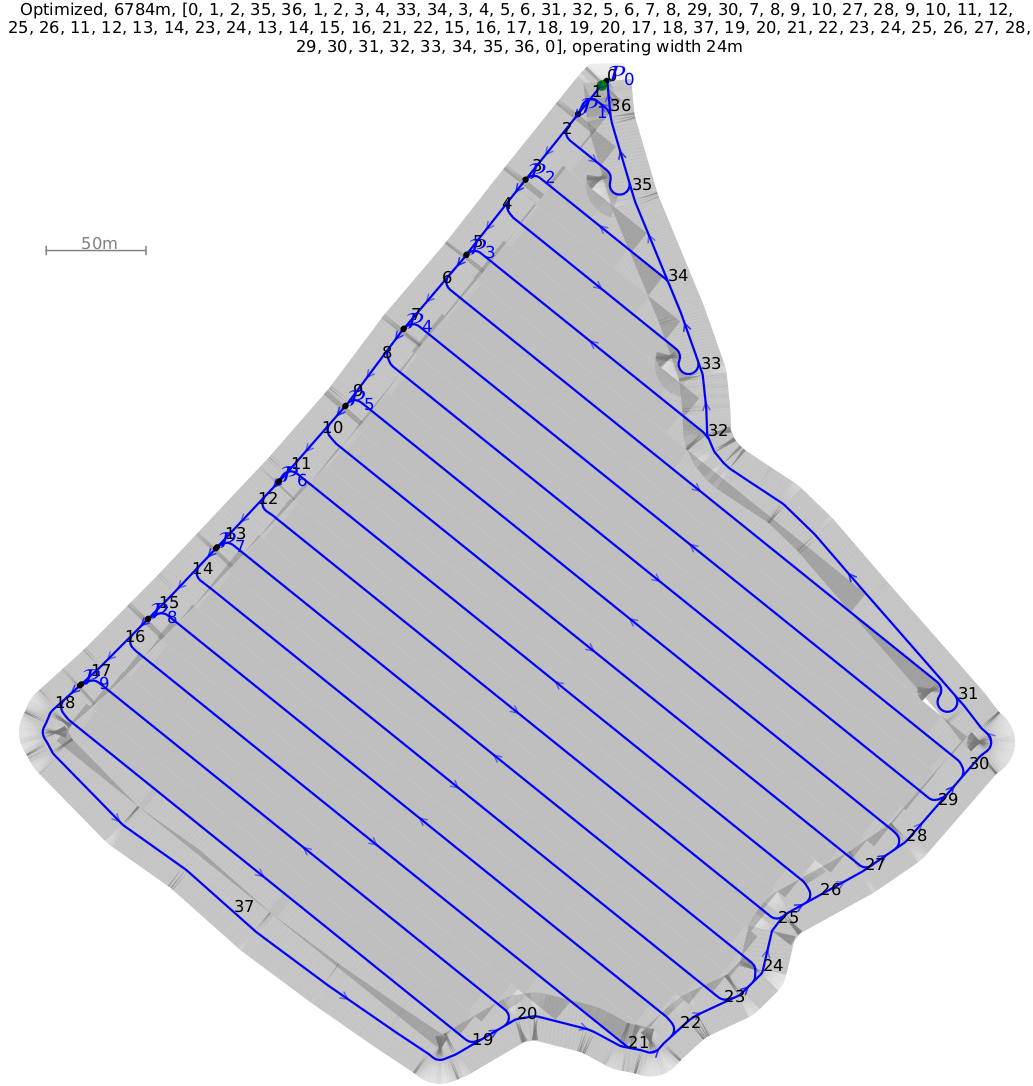}\\[-1pt]
\caption{$S_{\textsf{M}_2}^{2}$}
  \label{fig_f7_SO1}
\end{subfigure}
\begin{subfigure}[t]{.33\linewidth}
  \centering
  \includegraphics[width=.99\linewidth]{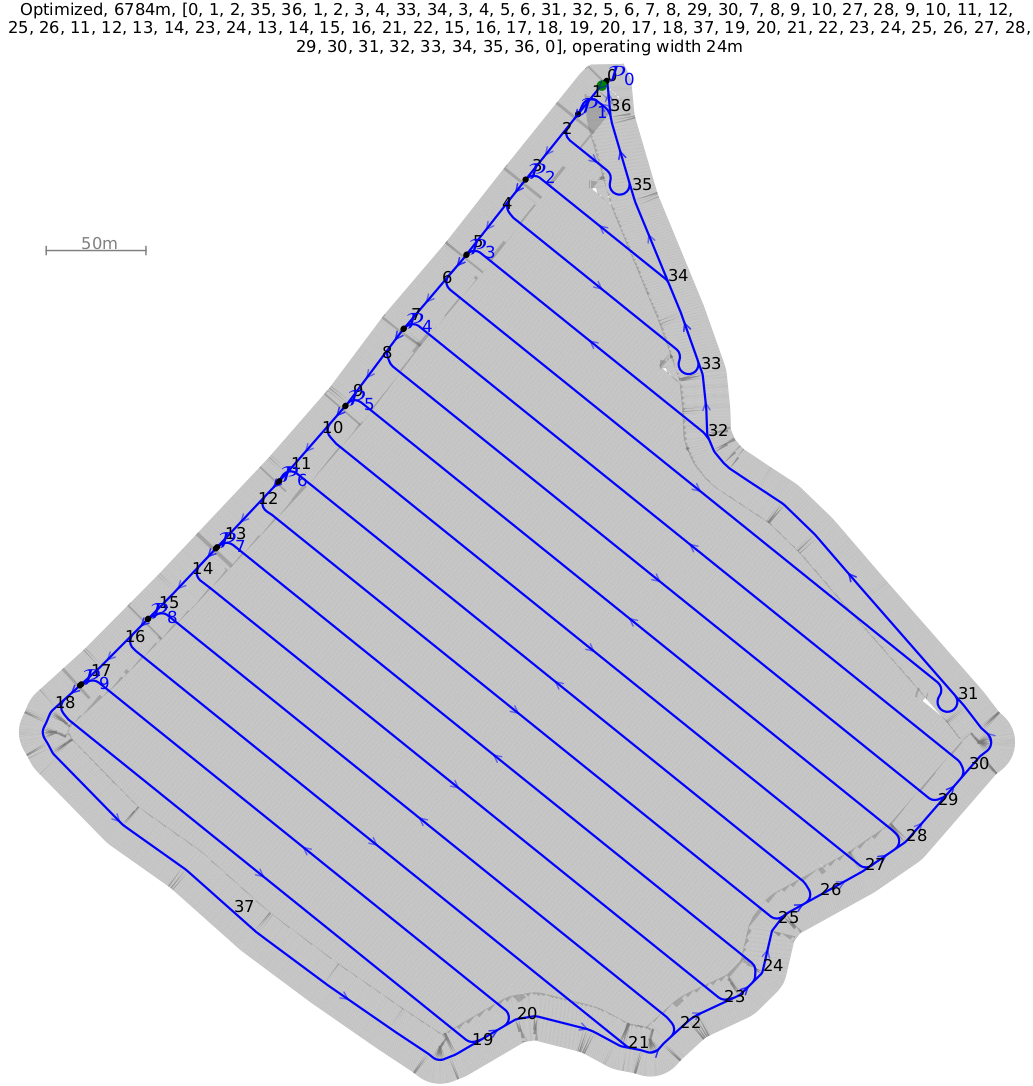}\\[-1pt]
\caption{$S_{\textsf{M}_2}^{48}$}
  \label{fig_f7_SO2}
\end{subfigure}
\caption{Field example 5: visual comparison of 6 different setups. See Table \ref{tab_s} for quantitative evaluation. }
\label{fig_f6}
\end{figure*}

\begin{figure*}
\centering
\begin{subfigure}[t]{.33\linewidth}
  \centering
  \includegraphics[width=.99\linewidth]{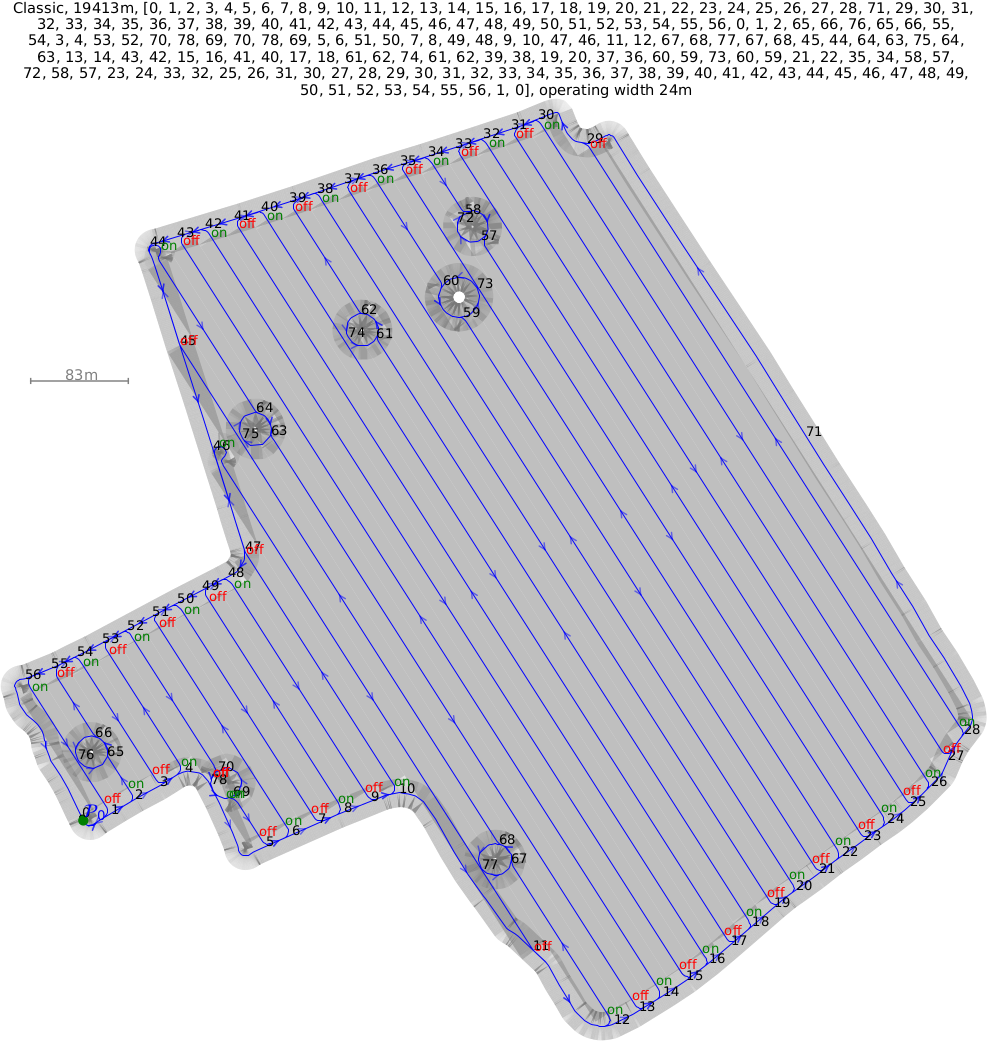}\\[-1pt]
\caption{$S_{\textsf{M}_1}^{1}$ \\[10pt]}
  \label{fig_f10_SC0}
\end{subfigure}
\begin{subfigure}[t]{.33\linewidth}
  \centering
  \includegraphics[width=.99\linewidth]{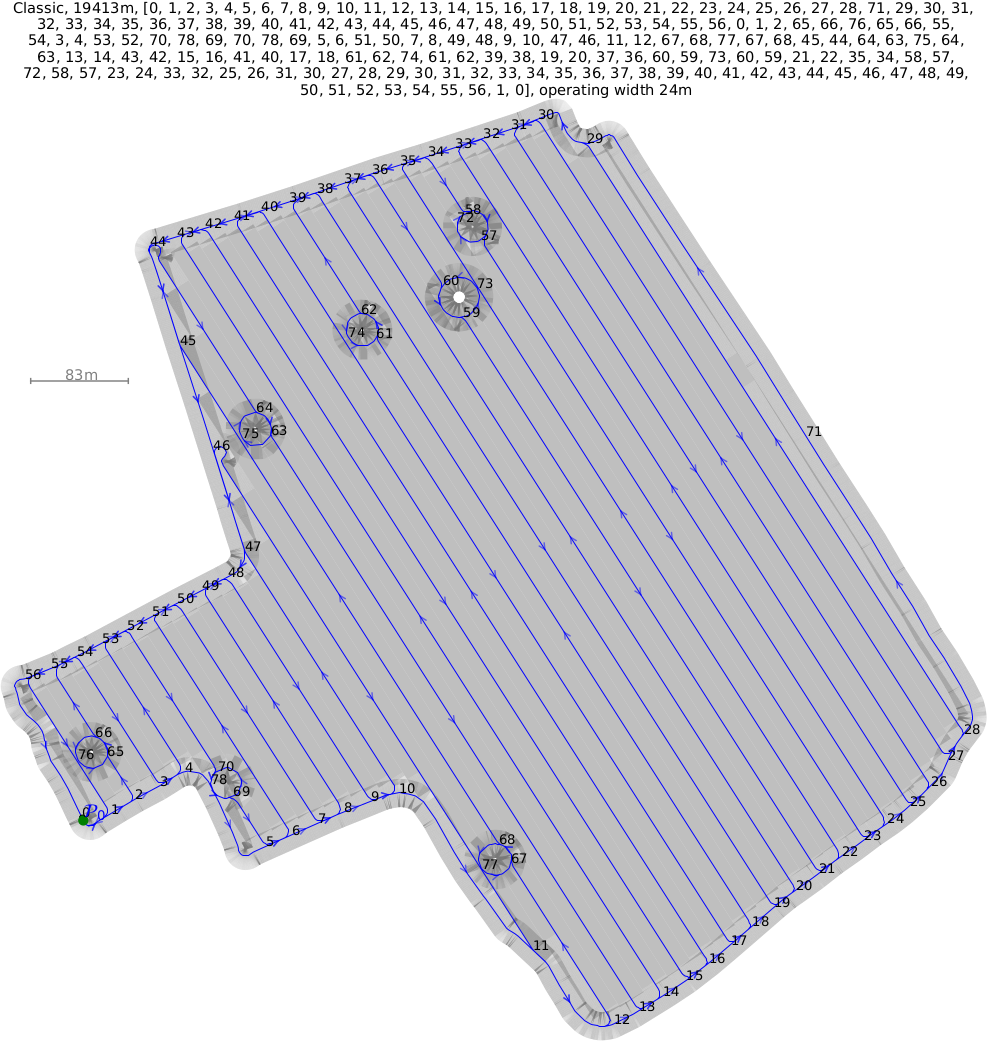}\\[-1pt]
\caption{$S_{\textsf{M}_1}^{2}$\\[10pt]}
  \label{fig_f10_SC1}
\end{subfigure}
\begin{subfigure}[t]{.33\linewidth}
  \centering
  \includegraphics[width=.99\linewidth]{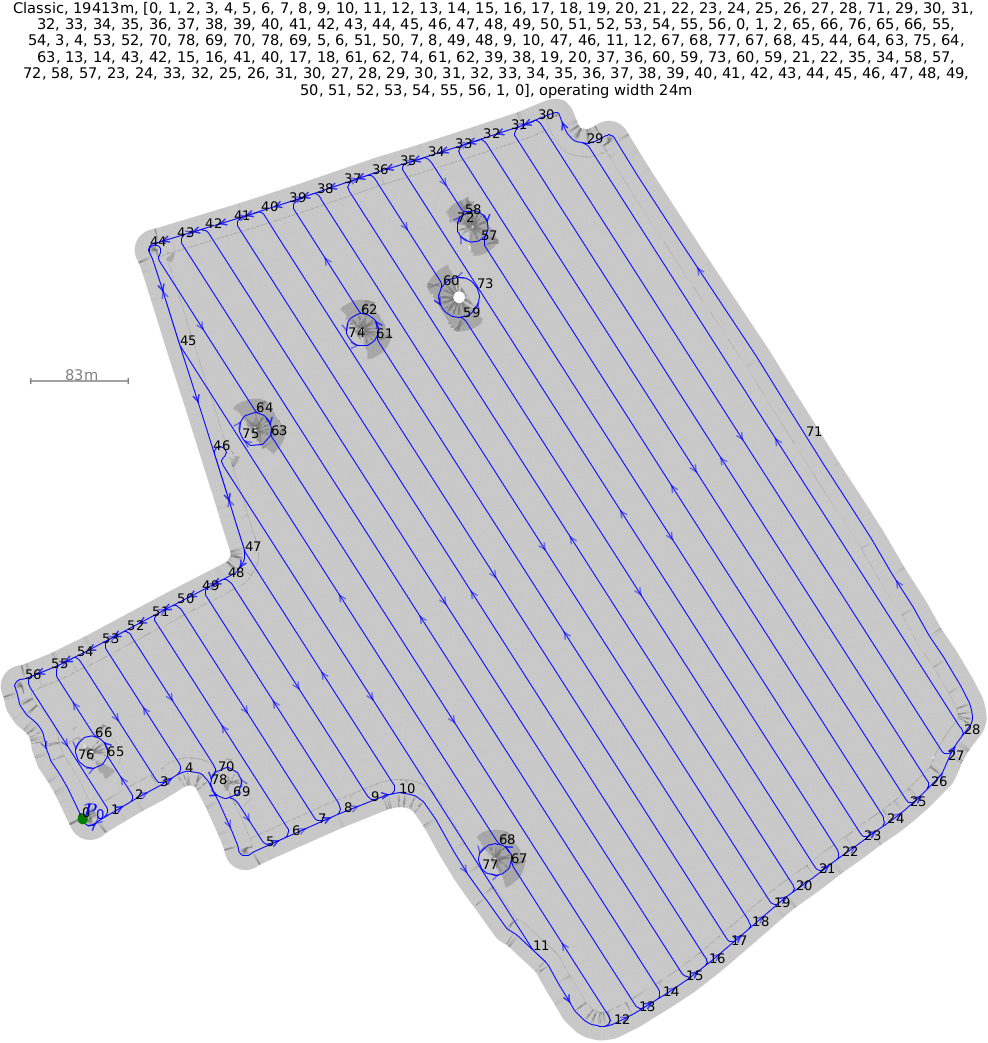}\\[-1pt]
\caption{$S_{\textsf{M}_1}^{48}$\\[10pt]}
  \label{fig_f10_SC2}
\end{subfigure}
\begin{subfigure}[t]{.33\linewidth}
  \centering
  \includegraphics[width=.99\linewidth]{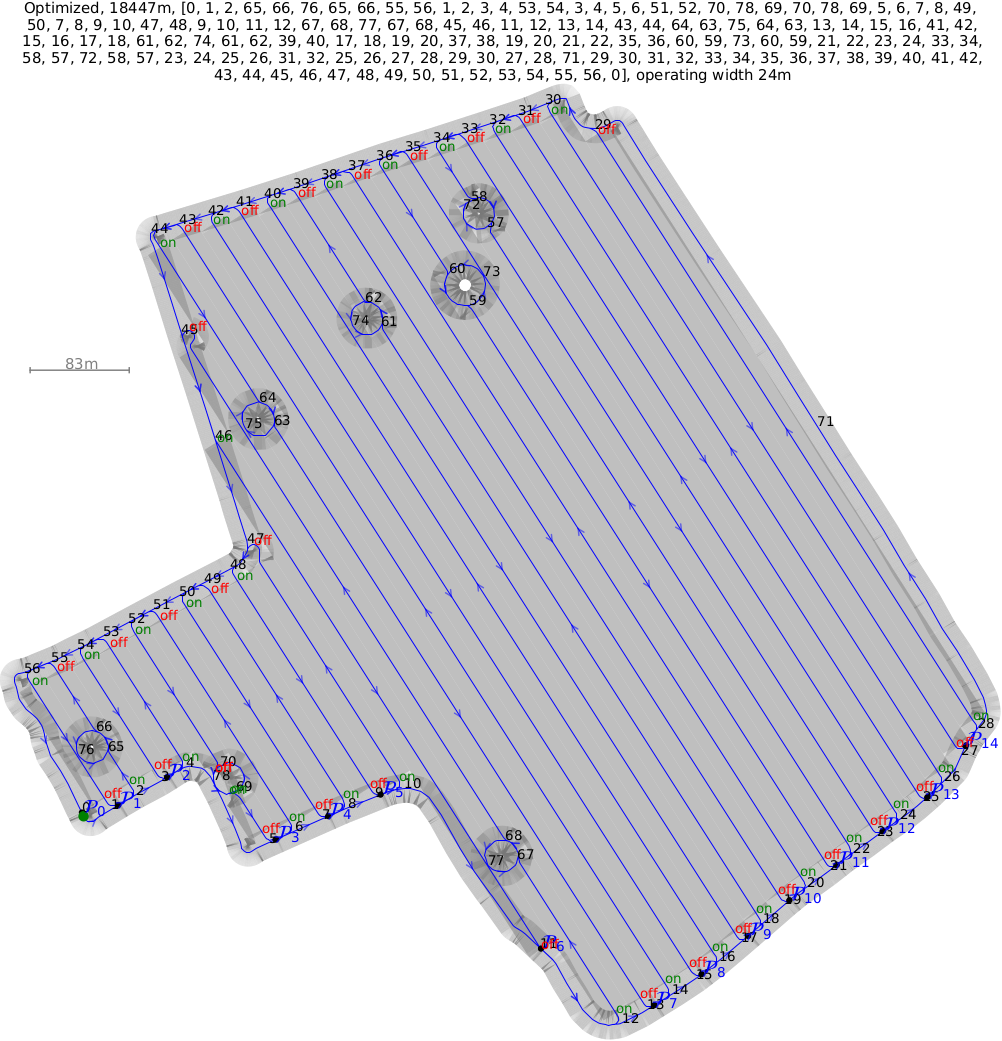}\\[-1pt]
\caption{$S_{\textsf{M}_2}^{1}$}
  \label{fig_f10_SO0}
\end{subfigure}
\begin{subfigure}[t]{.33\linewidth}
  \centering
  \includegraphics[width=.99\linewidth]{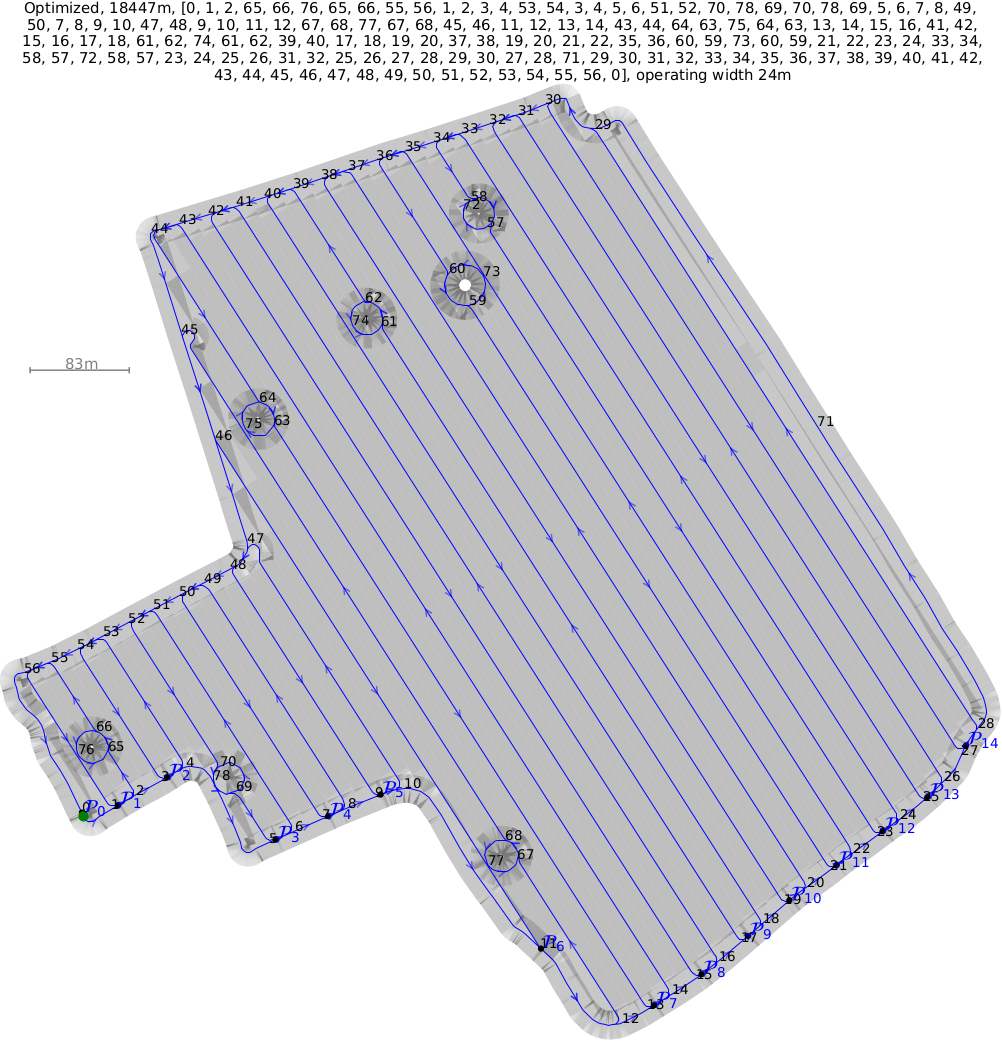}\\[-1pt]
\caption{$S_{\textsf{M}_2}^{2}$}
  \label{fig_f10_SO1}
\end{subfigure}
\begin{subfigure}[t]{.33\linewidth}
  \centering
  \includegraphics[width=.99\linewidth]{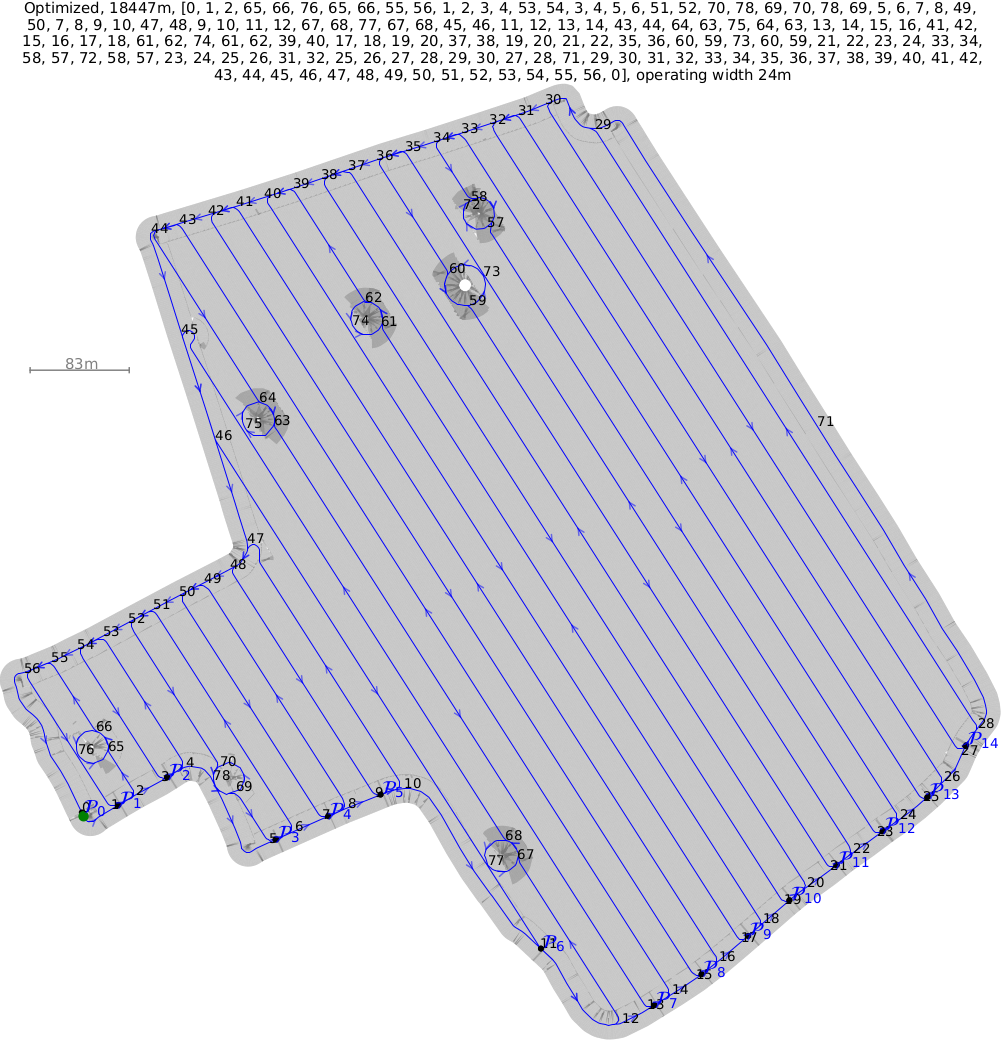}\\[-1pt]
\caption{$S_{\textsf{M}_2}^{48}$}
  \label{fig_f10_SO2}
\end{subfigure}
\caption{Field example 6: visual comparison of 6 different setups. See Table \ref{tab_s} for quantitative evaluation. }
\label{fig_f10}
\end{figure*}

\begin{figure*}
\centering
\begin{subfigure}[t]{.33\linewidth}
  \centering
  \includegraphics[width=.99\linewidth]{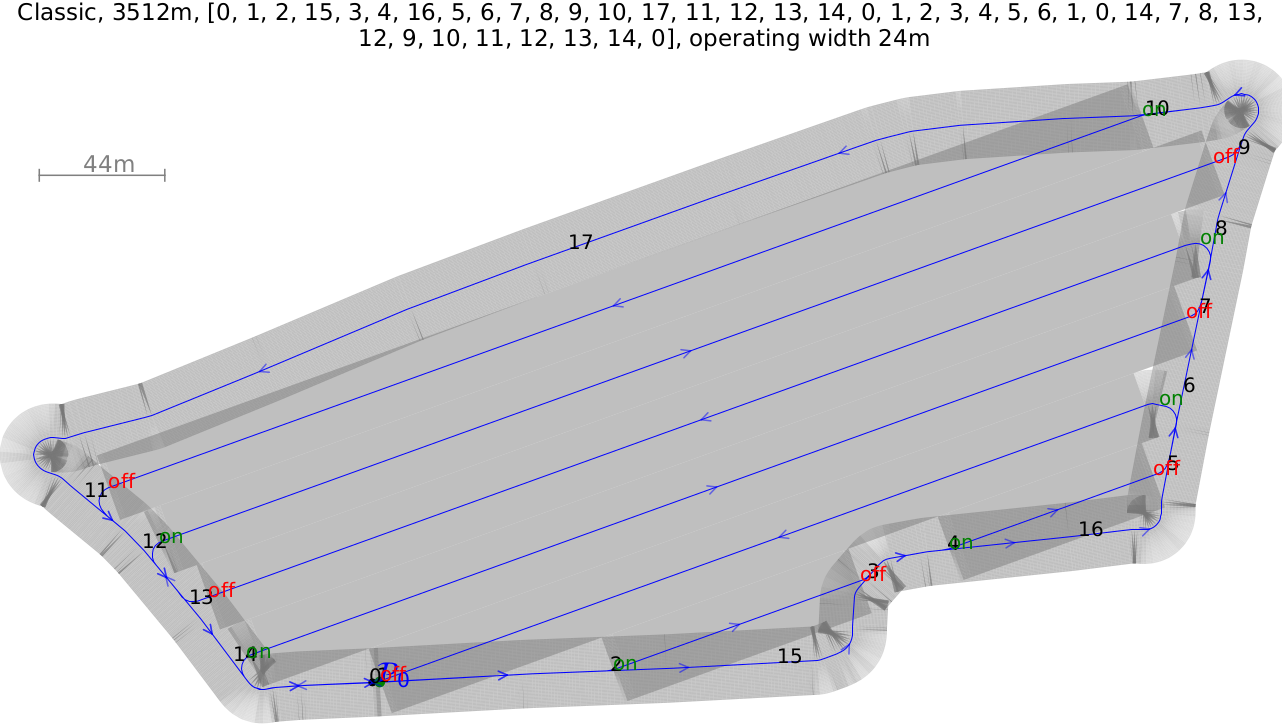}\\[-1pt]
\caption{$S_{\textsf{M}_1}^{1}$ \\[10pt]}
  \label{fig_f5_SC0}
\end{subfigure}
\begin{subfigure}[t]{.33\linewidth}
  \centering
  \includegraphics[width=.99\linewidth]{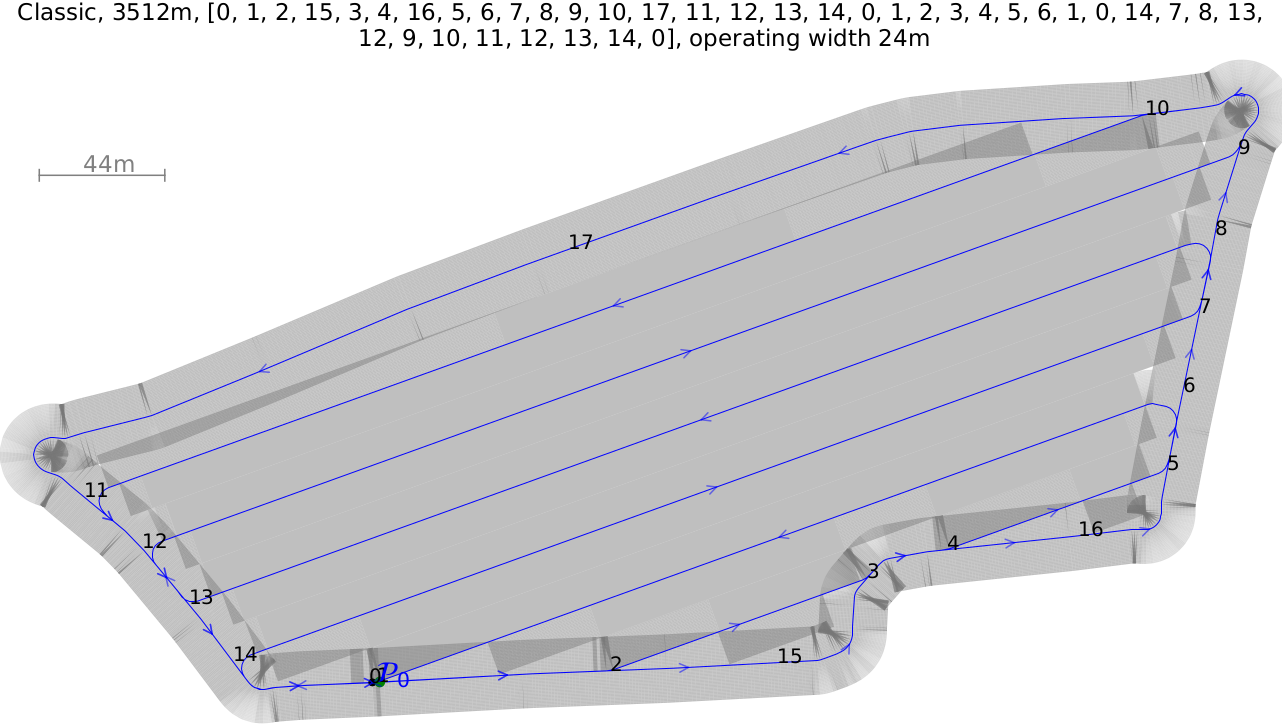}\\[-1pt]
\caption{$S_{\textsf{M}_1}^{2}$\\[10pt]}
  \label{fig_f5_SC1}
\end{subfigure}
\begin{subfigure}[t]{.33\linewidth}
  \centering
  \includegraphics[width=.99\linewidth]{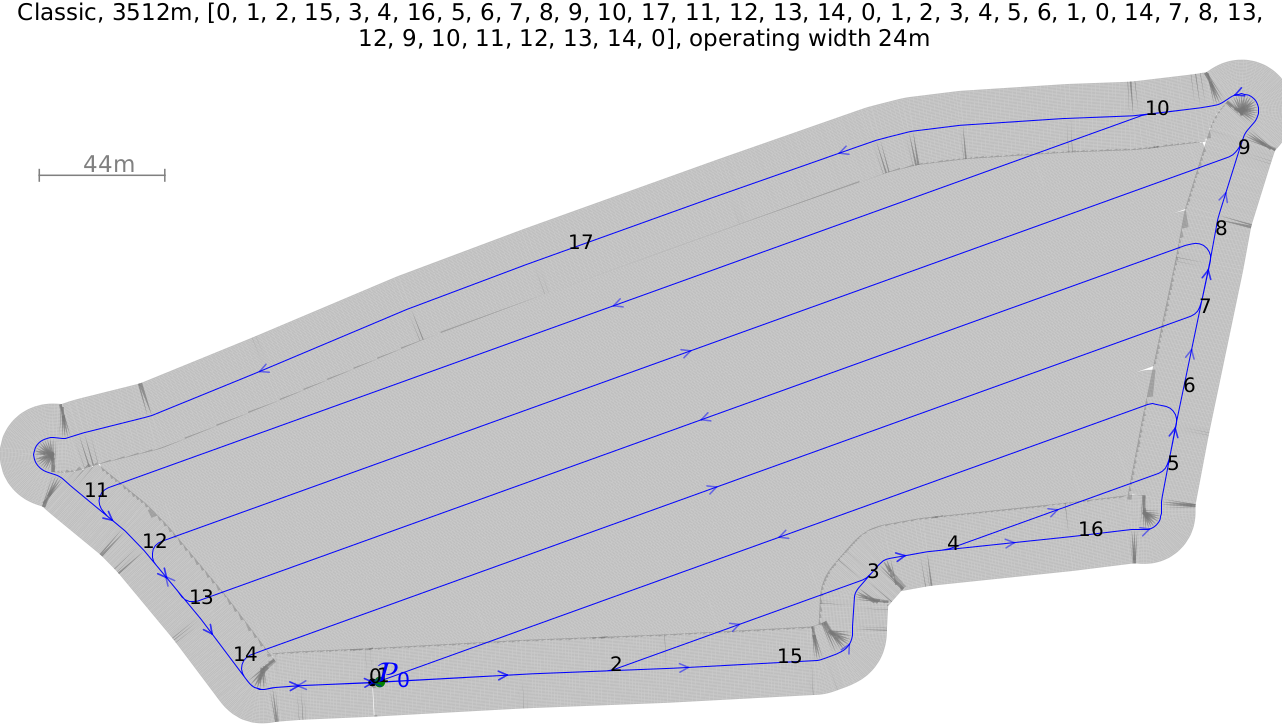}\\[-1pt]
\caption{$S_{\textsf{M}_1}^{48}$\\[10pt]}
  \label{fig_f5_SC2}
\end{subfigure}
\begin{subfigure}[t]{.33\linewidth}
  \centering
  \includegraphics[width=.99\linewidth]{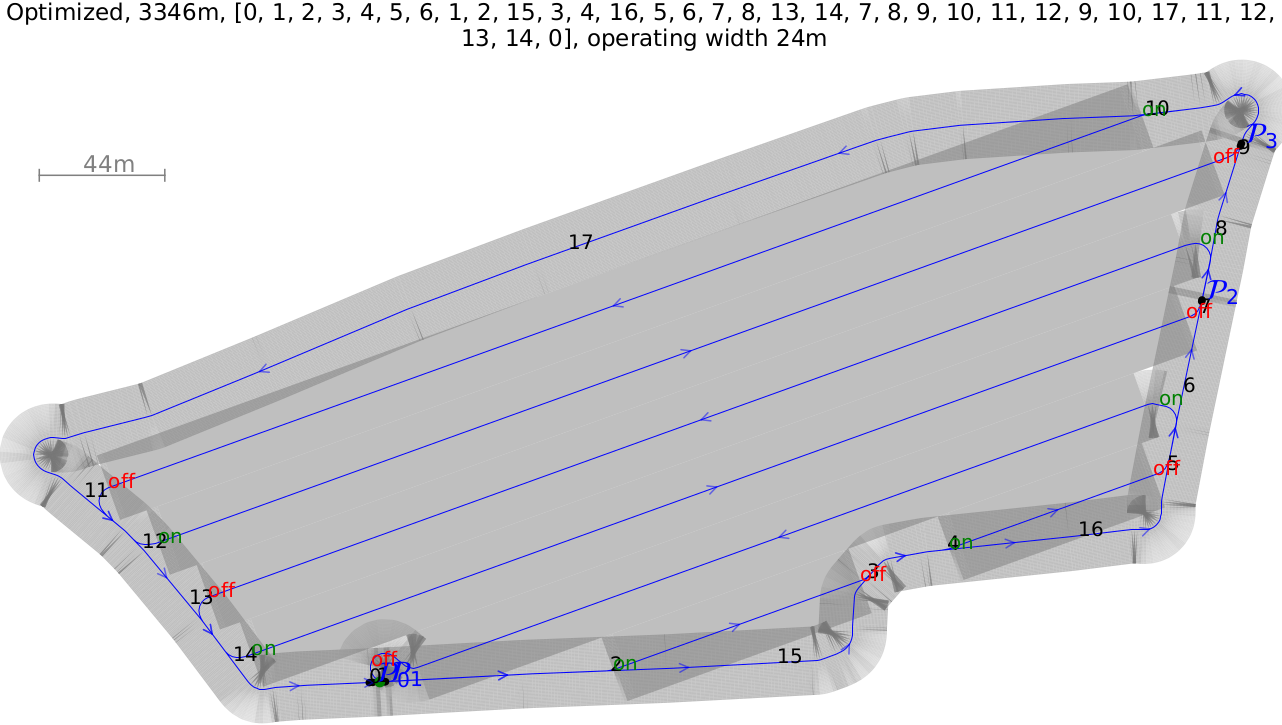}\\[-1pt]
\caption{$S_{\textsf{M}_2}^{1}$}
  \label{fig_f5_SO0}
\end{subfigure}
\begin{subfigure}[t]{.33\linewidth}
  \centering
  \includegraphics[width=.99\linewidth]{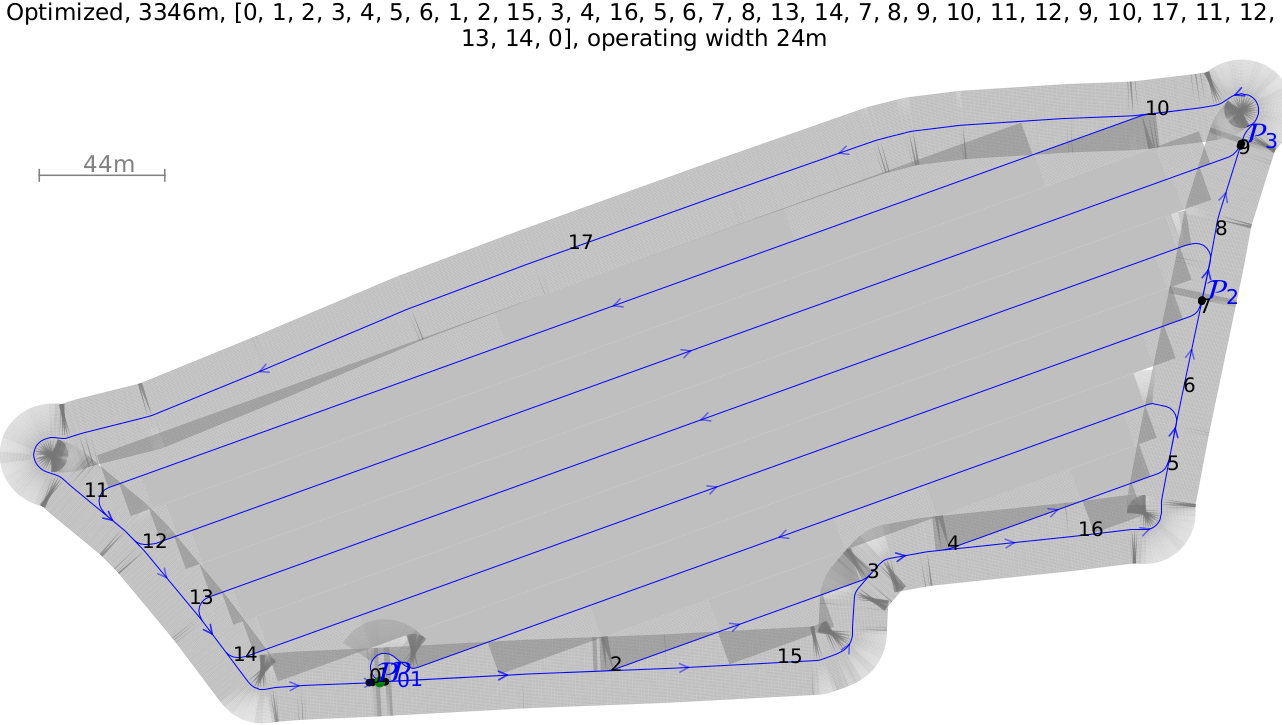}\\[-1pt]
\caption{$S_{\textsf{M}_2}^{2}$}
  \label{fig_f5_SO1}
\end{subfigure}
\begin{subfigure}[t]{.33\linewidth}
  \centering
  \includegraphics[width=.99\linewidth]{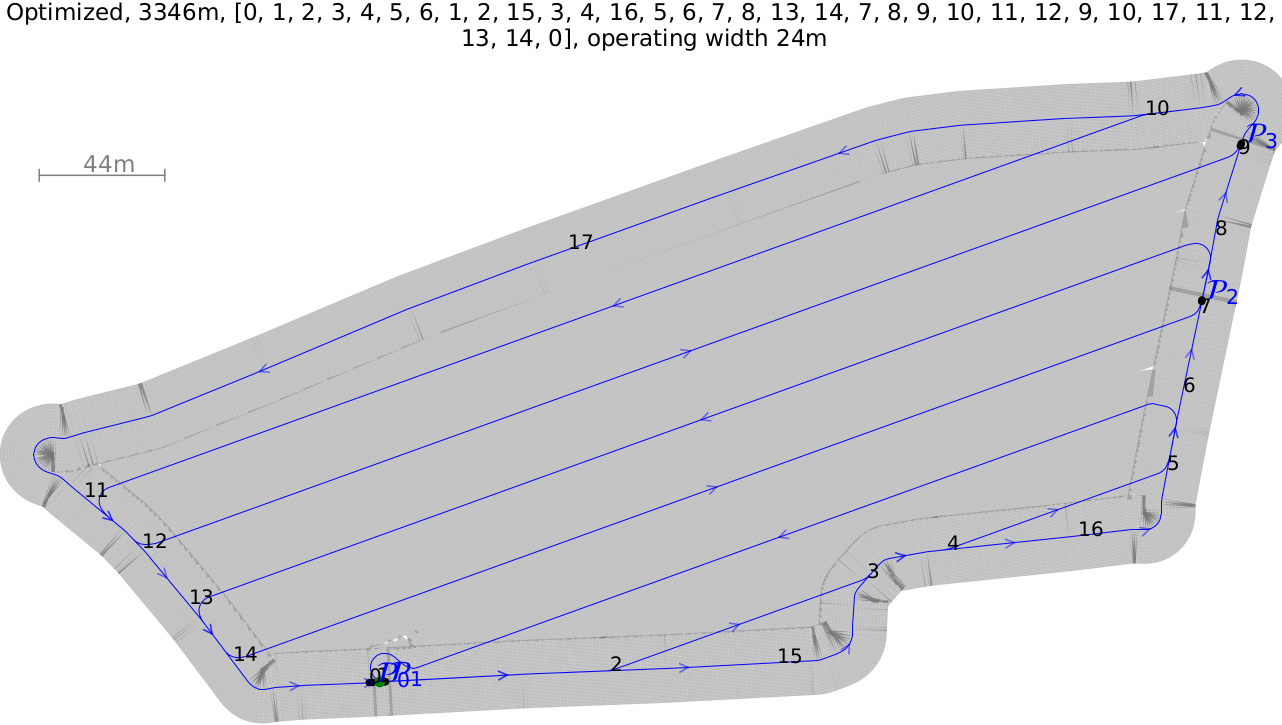}\\[-1pt]
\caption{$S_{\textsf{M}_2}^{48}$}
  \label{fig_f5_SO2}
\end{subfigure}
\caption{Field example 7: visual comparison of 6 different setups. See Table \ref{tab_s} for quantitative evaluation. }
\label{fig_f5}
\end{figure*}

\begin{figure*}
\centering
\begin{subfigure}[t]{.33\linewidth}
  \centering
  \includegraphics[width=.99\linewidth]{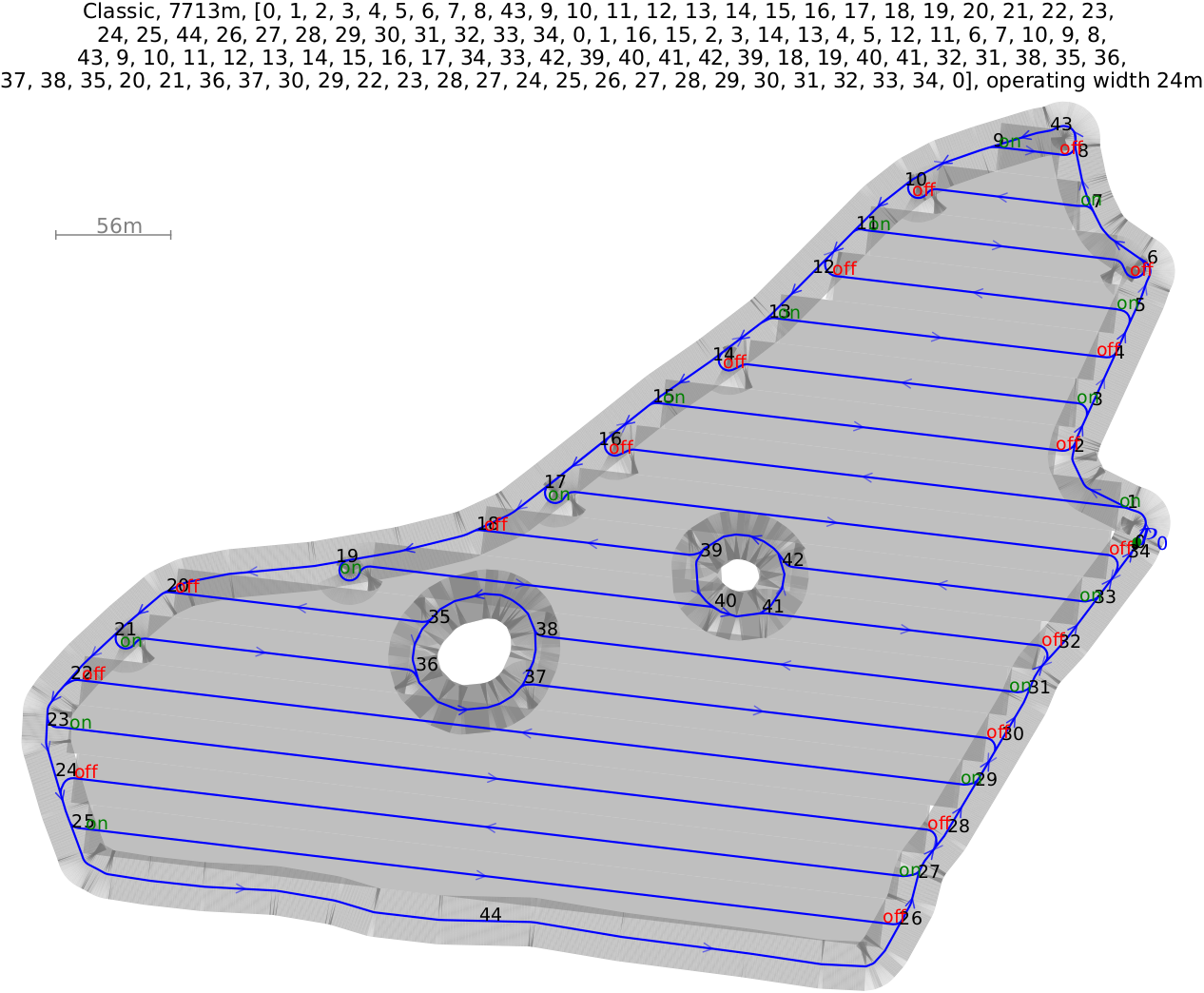}\\[-1pt]
\caption{$S_{\textsf{M}_1}^{1}$ \\[10pt]}
  \label{fig_f8_SC0}
\end{subfigure}
\begin{subfigure}[t]{.33\linewidth}
  \centering
  \includegraphics[width=.99\linewidth]{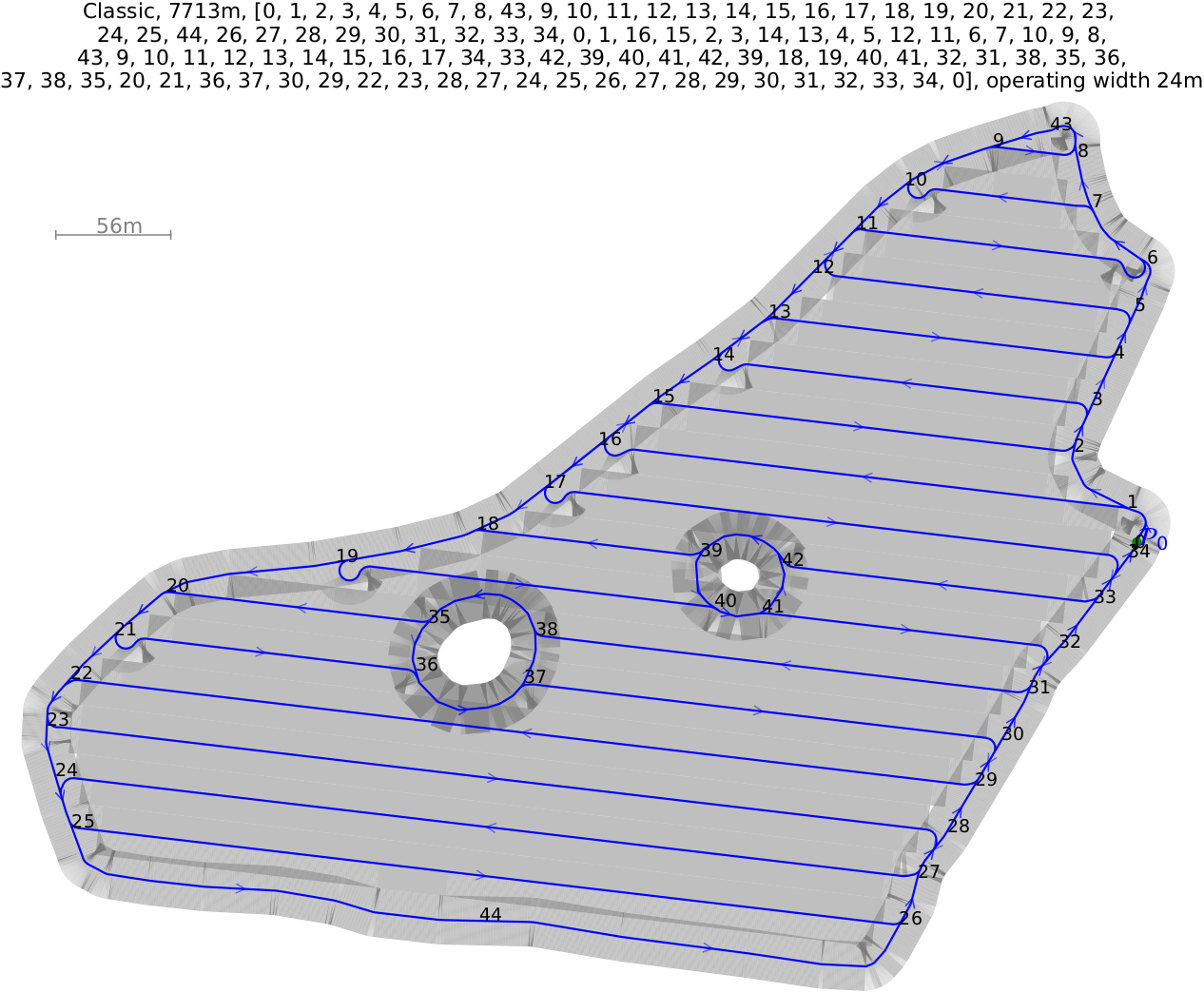}\\[-1pt]
\caption{$S_{\textsf{M}_1}^{2}$\\[10pt]}
  \label{fig_f8_SC1}
\end{subfigure}
\begin{subfigure}[t]{.33\linewidth}
  \centering
  \includegraphics[width=.99\linewidth]{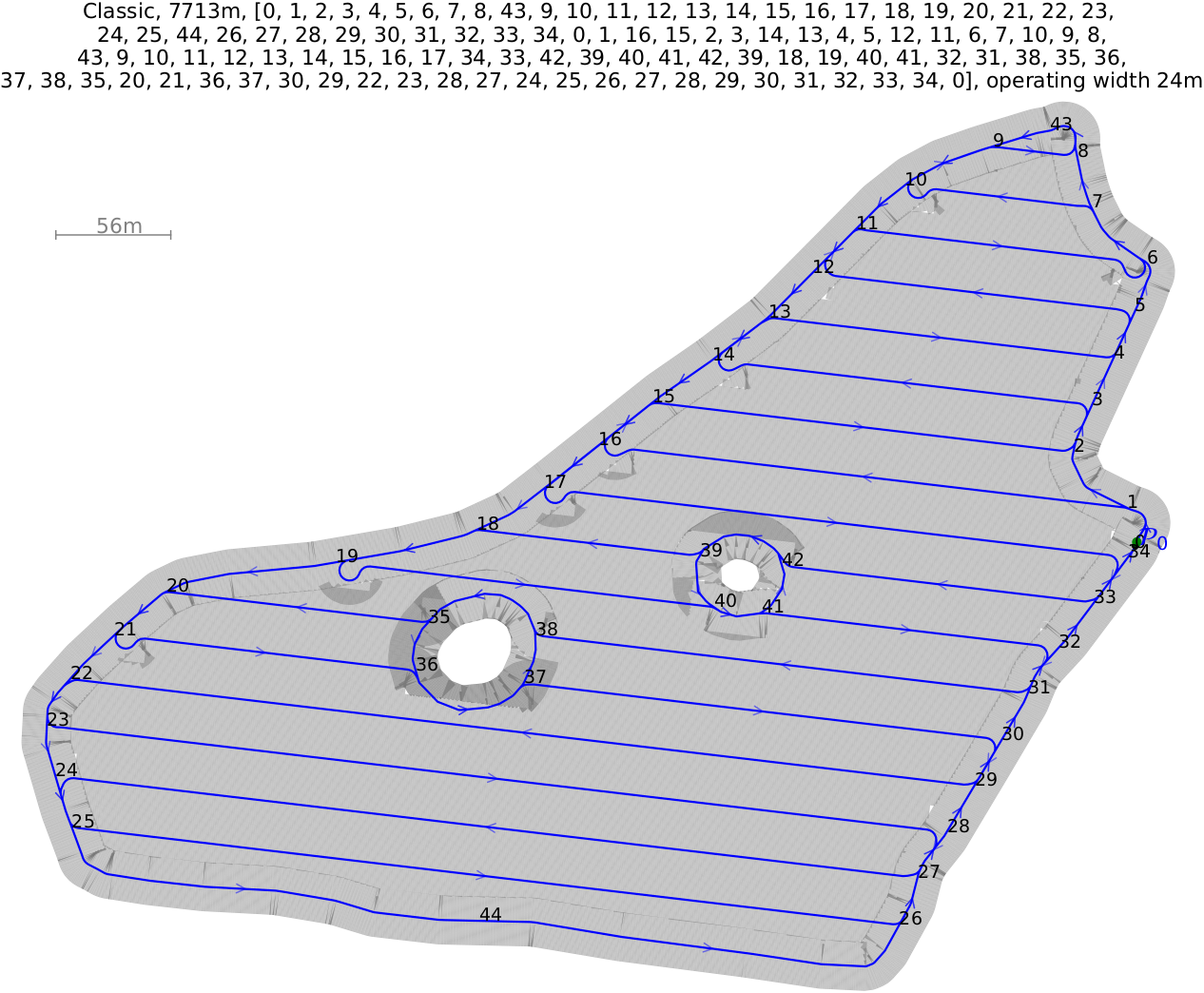}\\[-1pt]
\caption{$S_{\textsf{M}_1}^{48}$\\[10pt]}
  \label{fig_f8_SC2}
\end{subfigure}
\begin{subfigure}[t]{.33\linewidth}
  \centering
  \includegraphics[width=.99\linewidth]{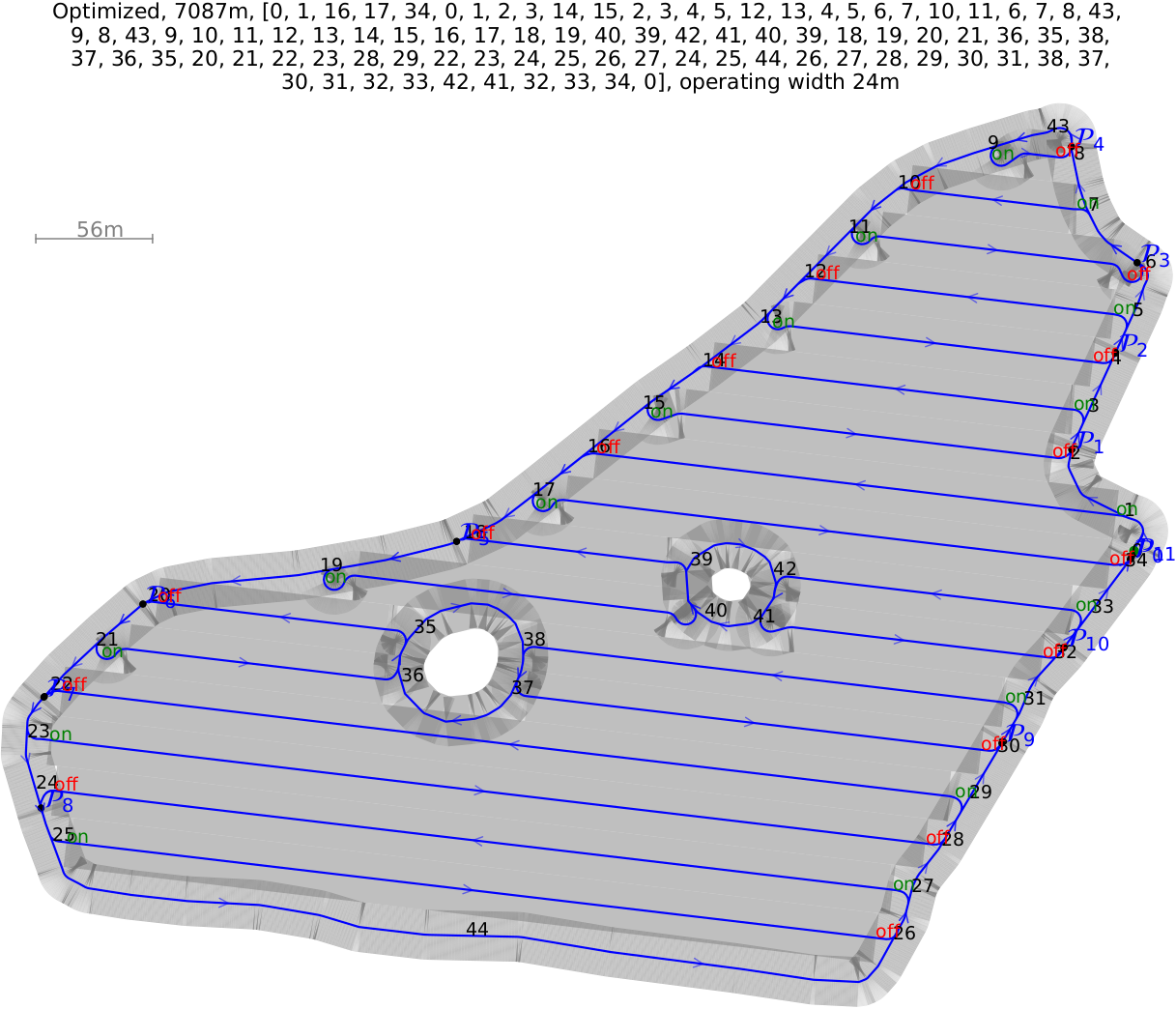}\\[-1pt]
\caption{$S_{\textsf{M}_2}^{1}$}
  \label{fig_f8_SO0}
\end{subfigure}
\begin{subfigure}[t]{.33\linewidth}
  \centering
  \includegraphics[width=.99\linewidth]{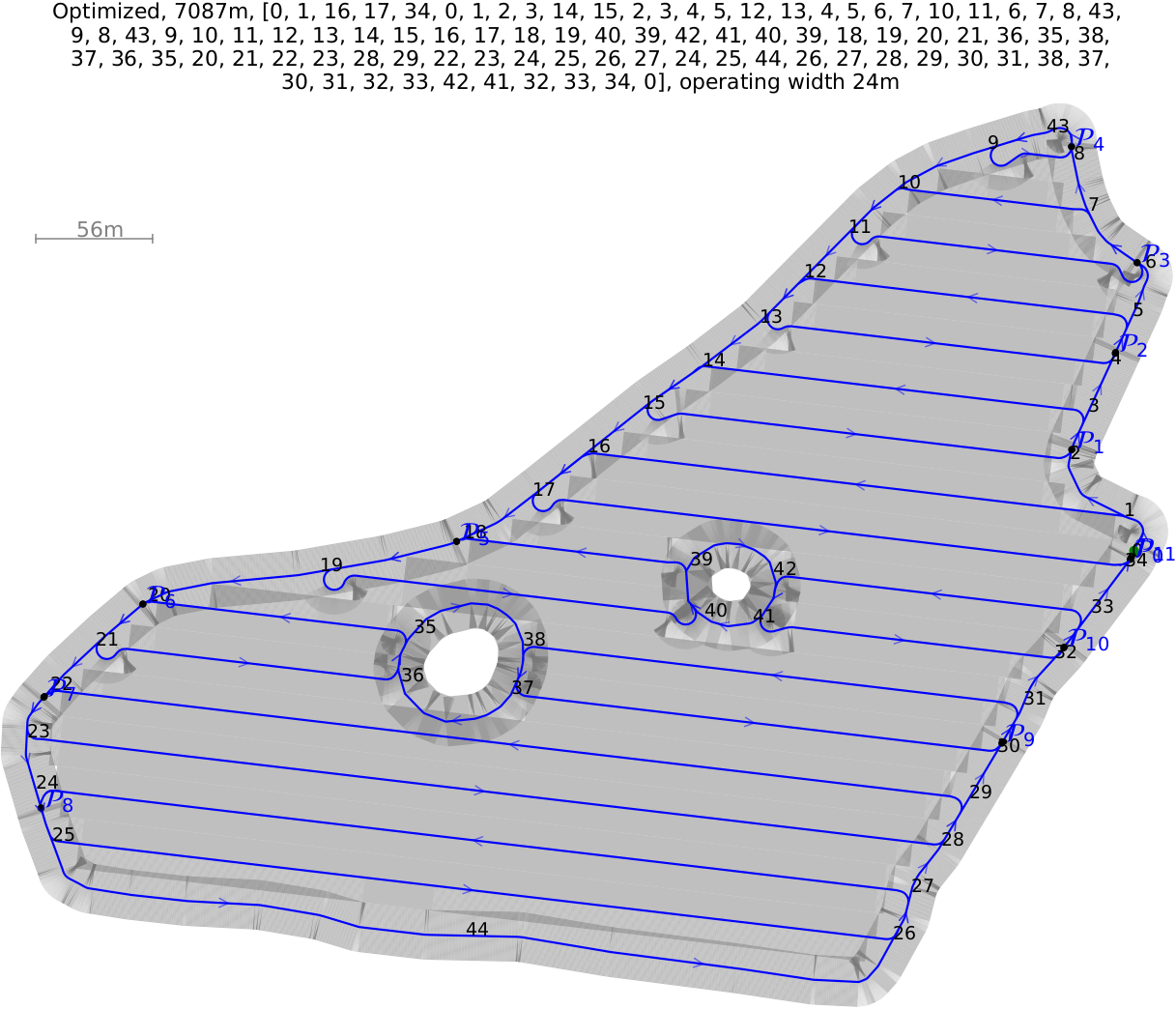}\\[-1pt]
\caption{$S_{\textsf{M}_2}^{2}$}
  \label{fig_f8_SO1}
\end{subfigure}
\begin{subfigure}[t]{.33\linewidth}
  \centering
  \includegraphics[width=.99\linewidth]{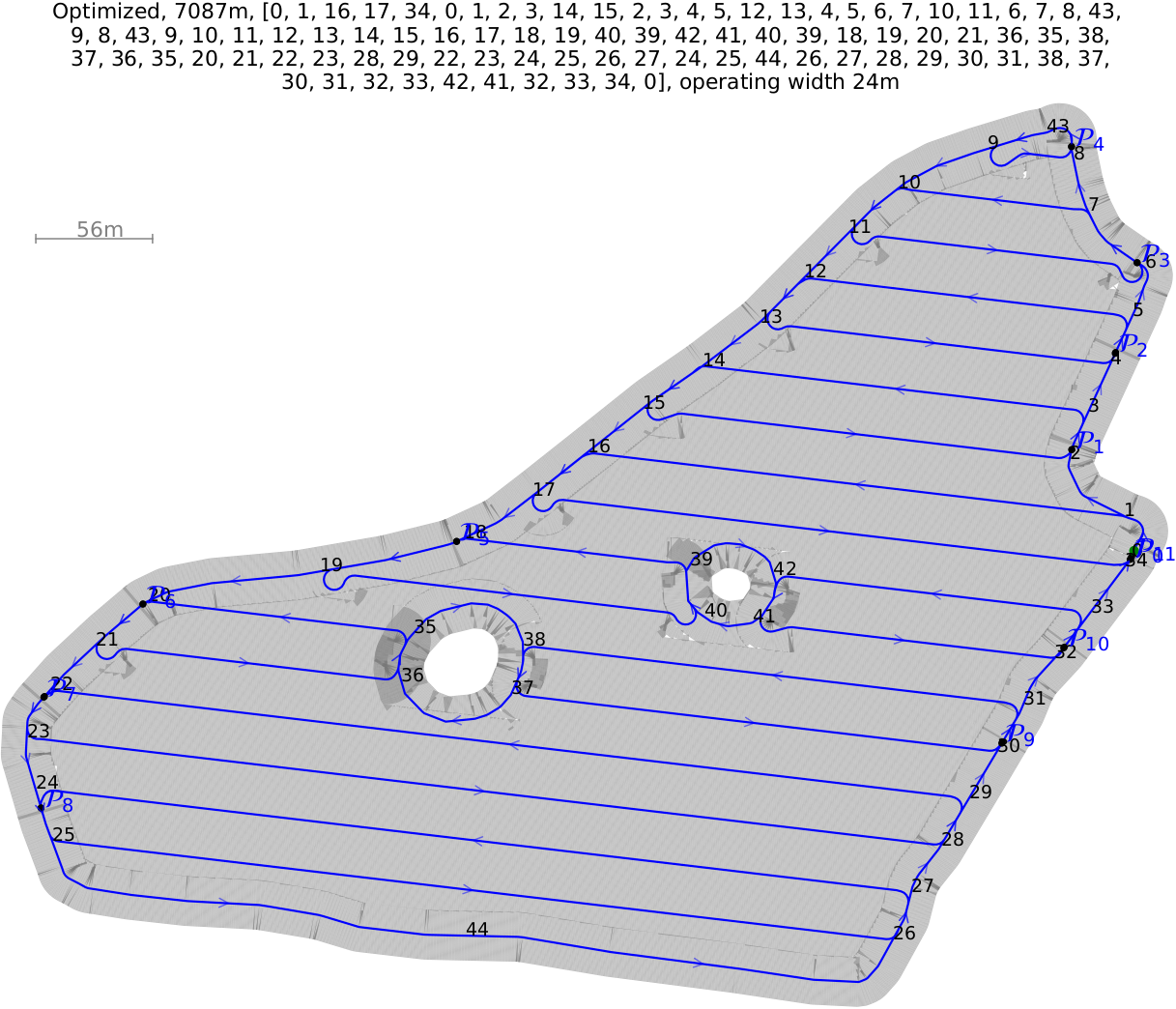}\\[-1pt]
\caption{$S_{\textsf{M}_2}^{48}$}
  \label{fig_f8_SO2}
\end{subfigure}
\caption{Field example 8: visual comparison of 6 different setups. See Table \ref{tab_s} for quantitative evaluation. }
\label{fig_f8}
\end{figure*}

\begin{figure*}
\centering
\begin{subfigure}[t]{.33\linewidth}
  \centering
  \includegraphics[width=.99\linewidth]{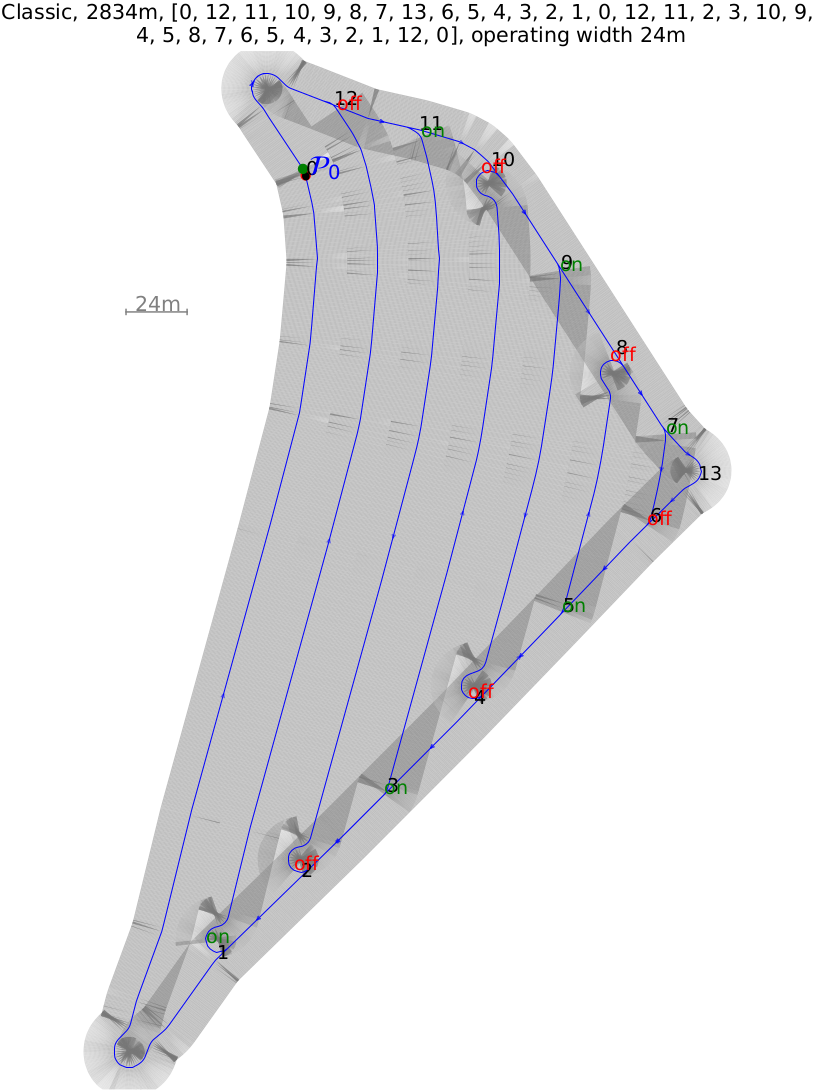}\\[-1pt]
\caption{$S_{\textsf{M}_1}^{1}$ \\[10pt]}
  \label{fig_f1_CC0}
\end{subfigure}
\begin{subfigure}[t]{.33\linewidth}
  \centering
  \includegraphics[width=.99\linewidth]{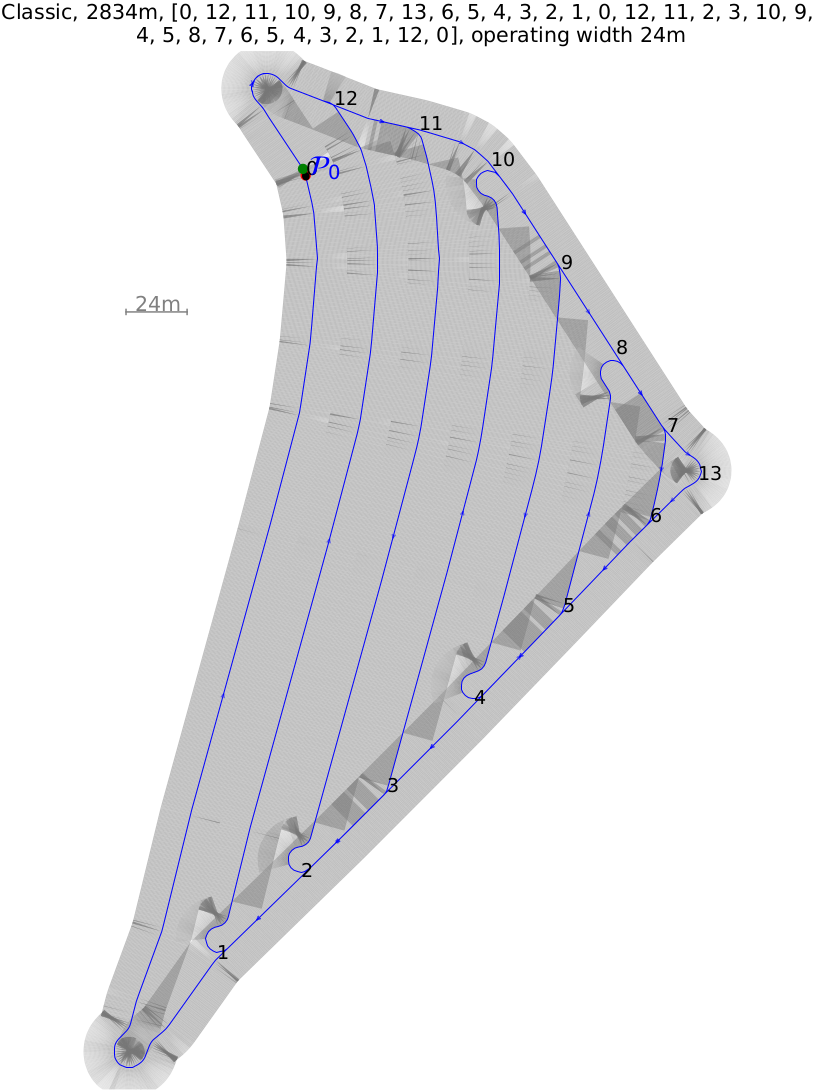}\\[-1pt]
\caption{$S_{\textsf{M}_1}^{2}$\\[10pt]}
  \label{fig_f1_CC1}
\end{subfigure}
\begin{subfigure}[t]{.33\linewidth}
  \centering
  \includegraphics[width=.99\linewidth]{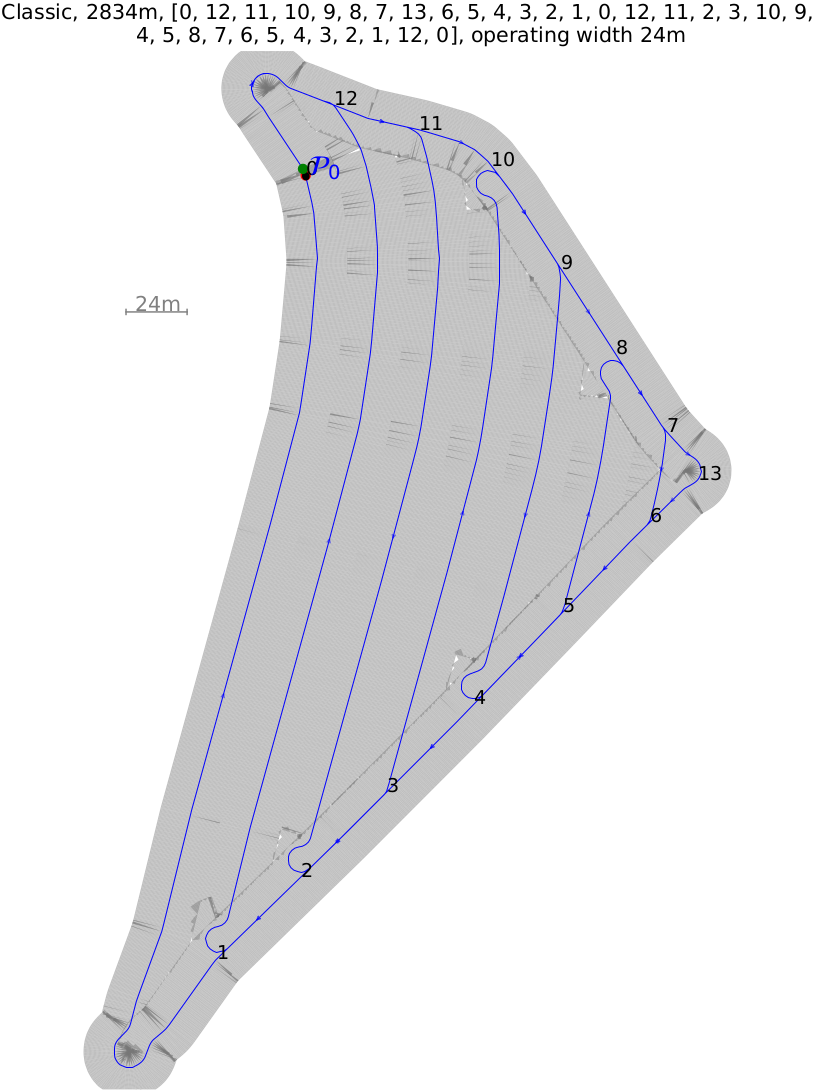}\\[-1pt]
\caption{$S_{\textsf{M}_1}^{48}$\\[10pt]}
  \label{fig_f1_CC2}
\end{subfigure}
\begin{subfigure}[t]{.33\linewidth}
  \centering
  \includegraphics[width=.99\linewidth]{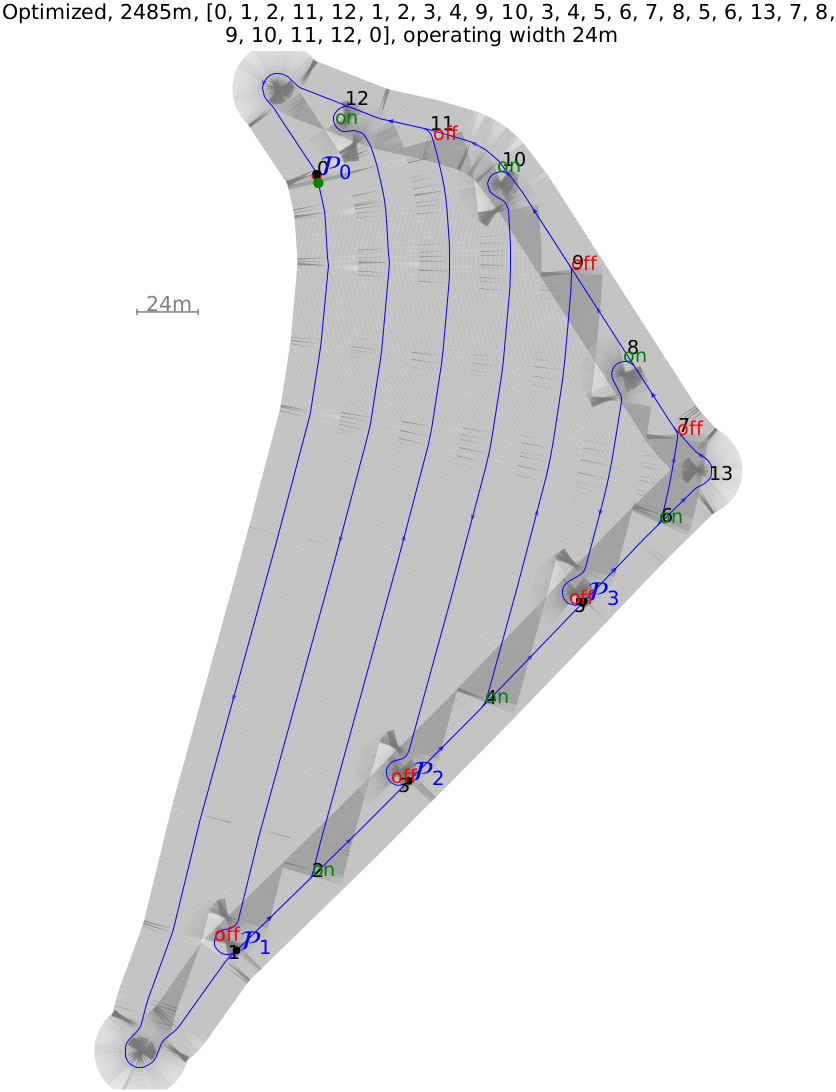}\\[-1pt]
\caption{$S_{\textsf{M}_2}^{1}$}
  \label{fig_f1_CO0}
\end{subfigure}
\begin{subfigure}[t]{.33\linewidth}
  \centering
  \includegraphics[width=.99\linewidth]{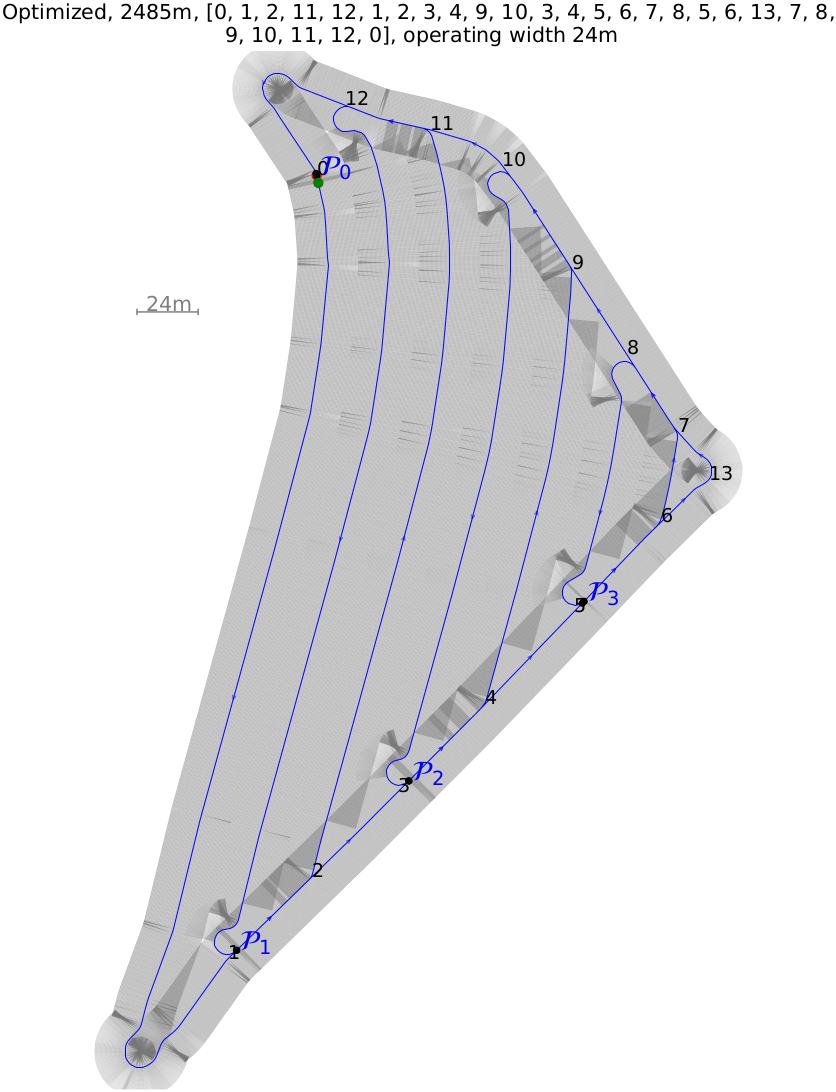}\\[-1pt]
\caption{$S_{\textsf{M}_2}^{2}$}
  \label{fig_f1_CO1}
\end{subfigure}
\begin{subfigure}[t]{.33\linewidth}
  \centering
  \includegraphics[width=.99\linewidth]{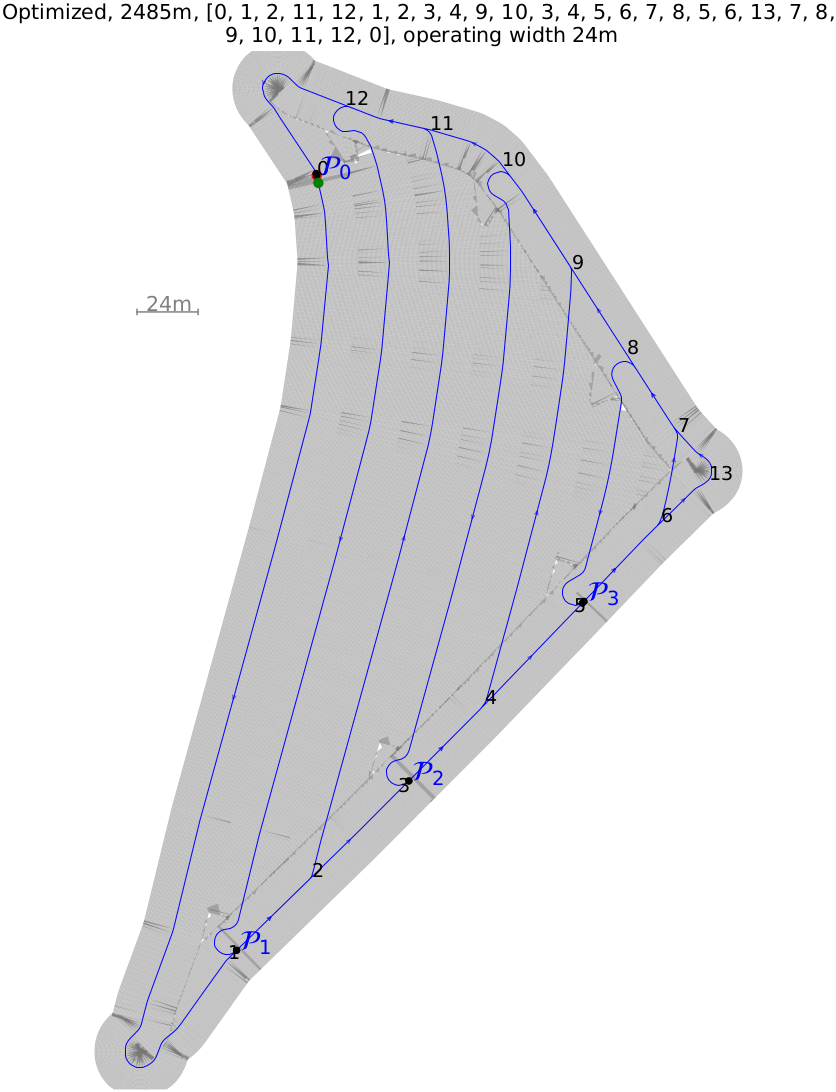}\\[-1pt]
\caption{$S_{\textsf{M}_2}^{48}$}
  \label{fig_f1_CO2}
\end{subfigure}
\caption{Field example 9: visual comparison of 6 different setups. See Table \ref{tab_s} for quantitative evaluation. }
\label{fig_f1}
\end{figure*}

\begin{figure*}
\centering
\begin{subfigure}[t]{.33\linewidth}
  \centering
  \includegraphics[width=.99\linewidth]{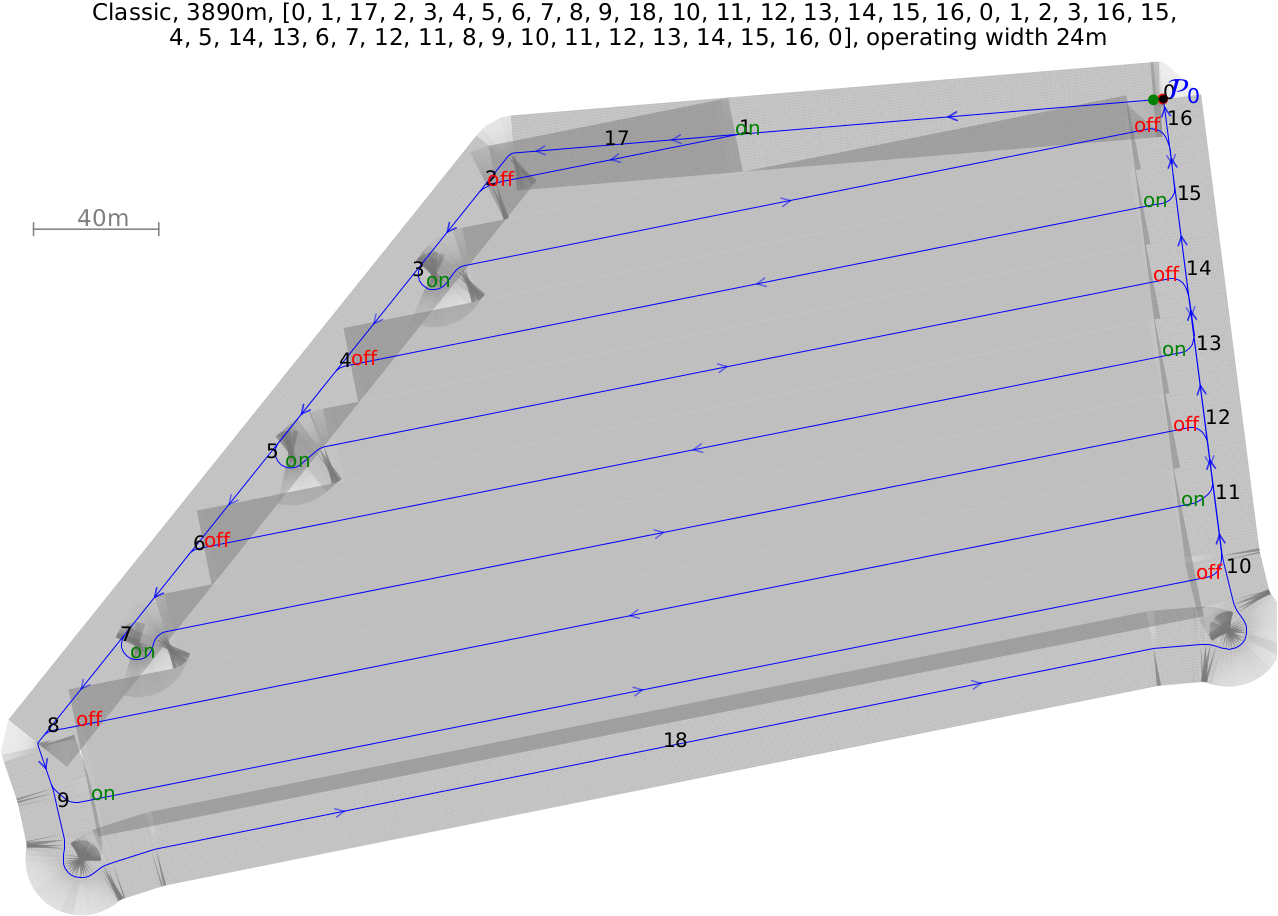}\\[-1pt]
\caption{$S_{\textsf{M}_1}^{1}$ \\[10pt]}
  \label{fig_f11_SC0}
\end{subfigure}
\begin{subfigure}[t]{.33\linewidth}
  \centering
  \includegraphics[width=.99\linewidth]{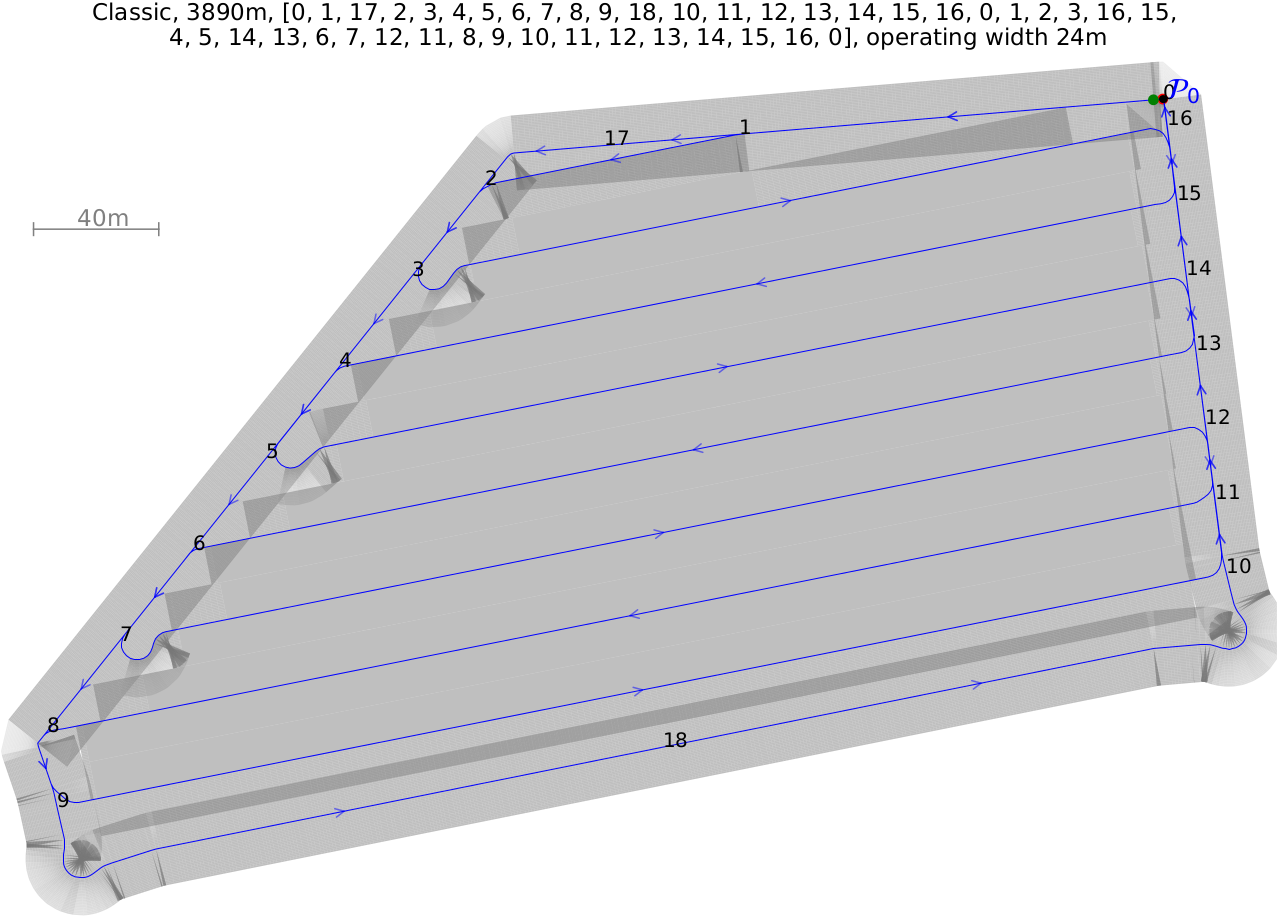}\\[-1pt]
\caption{$S_{\textsf{M}_1}^{2}$\\[10pt]}
  \label{fig_f11_SC1}
\end{subfigure}
\begin{subfigure}[t]{.33\linewidth}
  \centering
  \includegraphics[width=.99\linewidth]{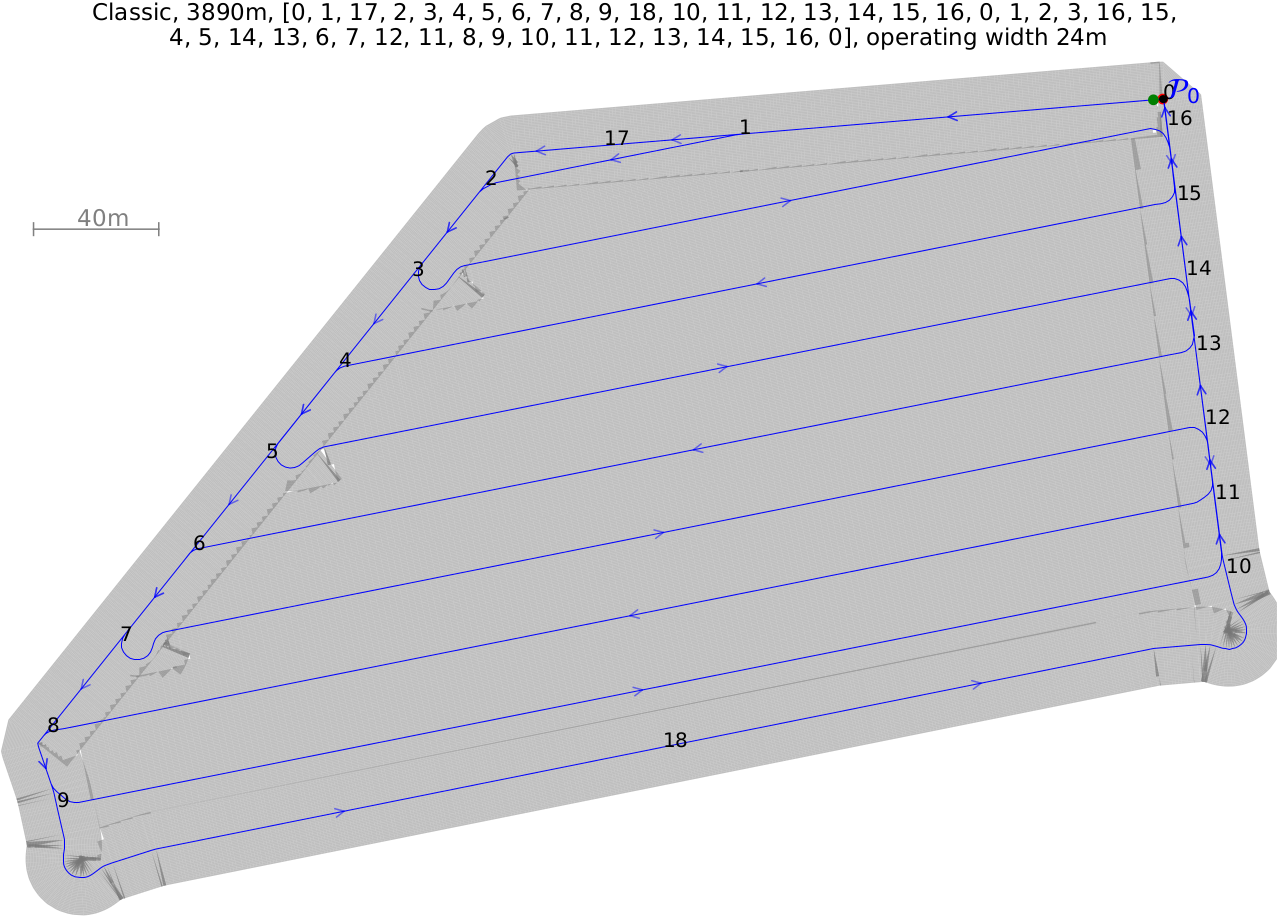}\\[-1pt]
\caption{$S_{\textsf{M}_1}^{48}$\\[10pt]}
  \label{fig_f11_SC2}
\end{subfigure}
\begin{subfigure}[t]{.33\linewidth}
  \centering
  \includegraphics[width=.99\linewidth]{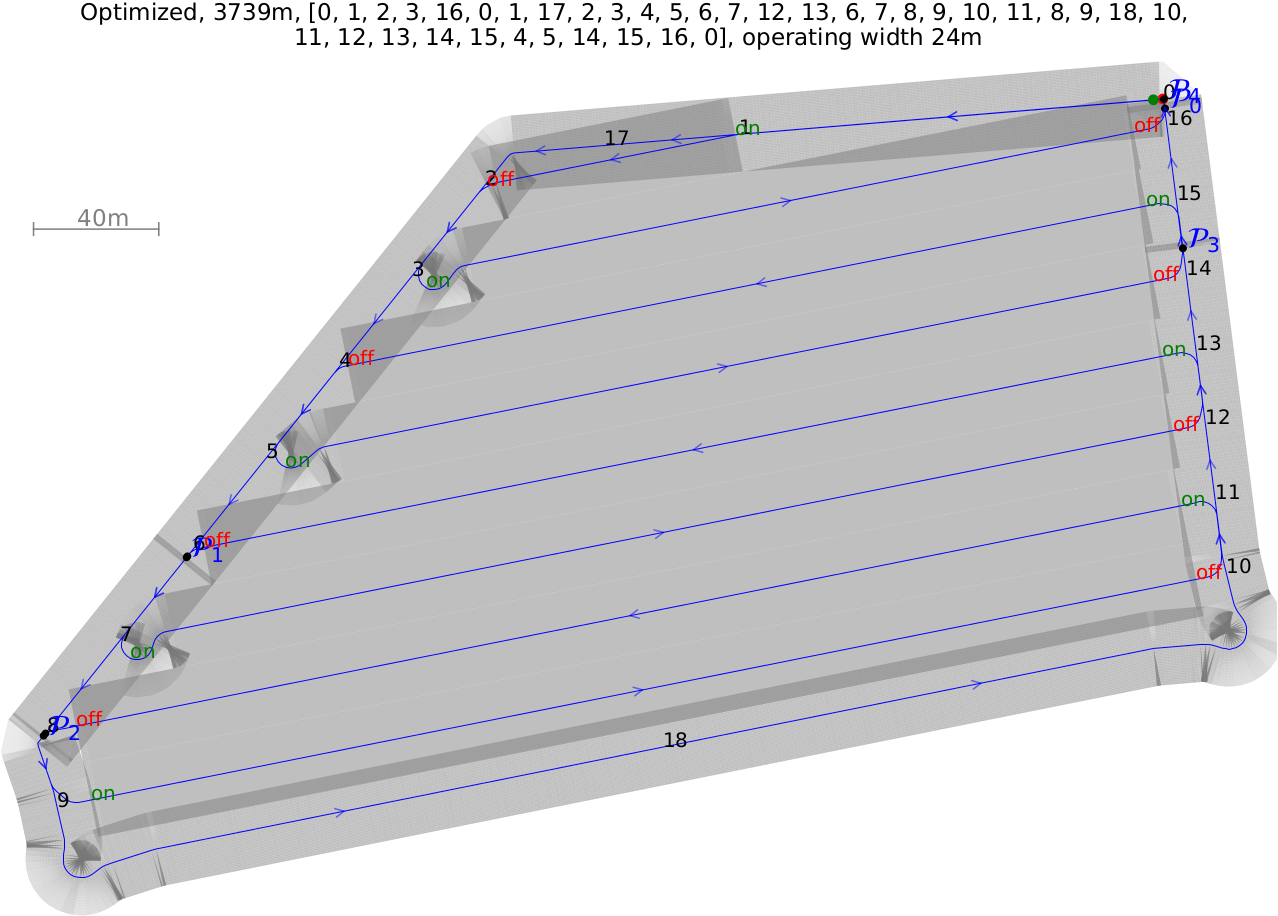}\\[-1pt]
\caption{$S_{\textsf{M}_2}^{1}$}
  \label{fig_f11_SO0}
\end{subfigure}
\begin{subfigure}[t]{.33\linewidth}
  \centering
  \includegraphics[width=.99\linewidth]{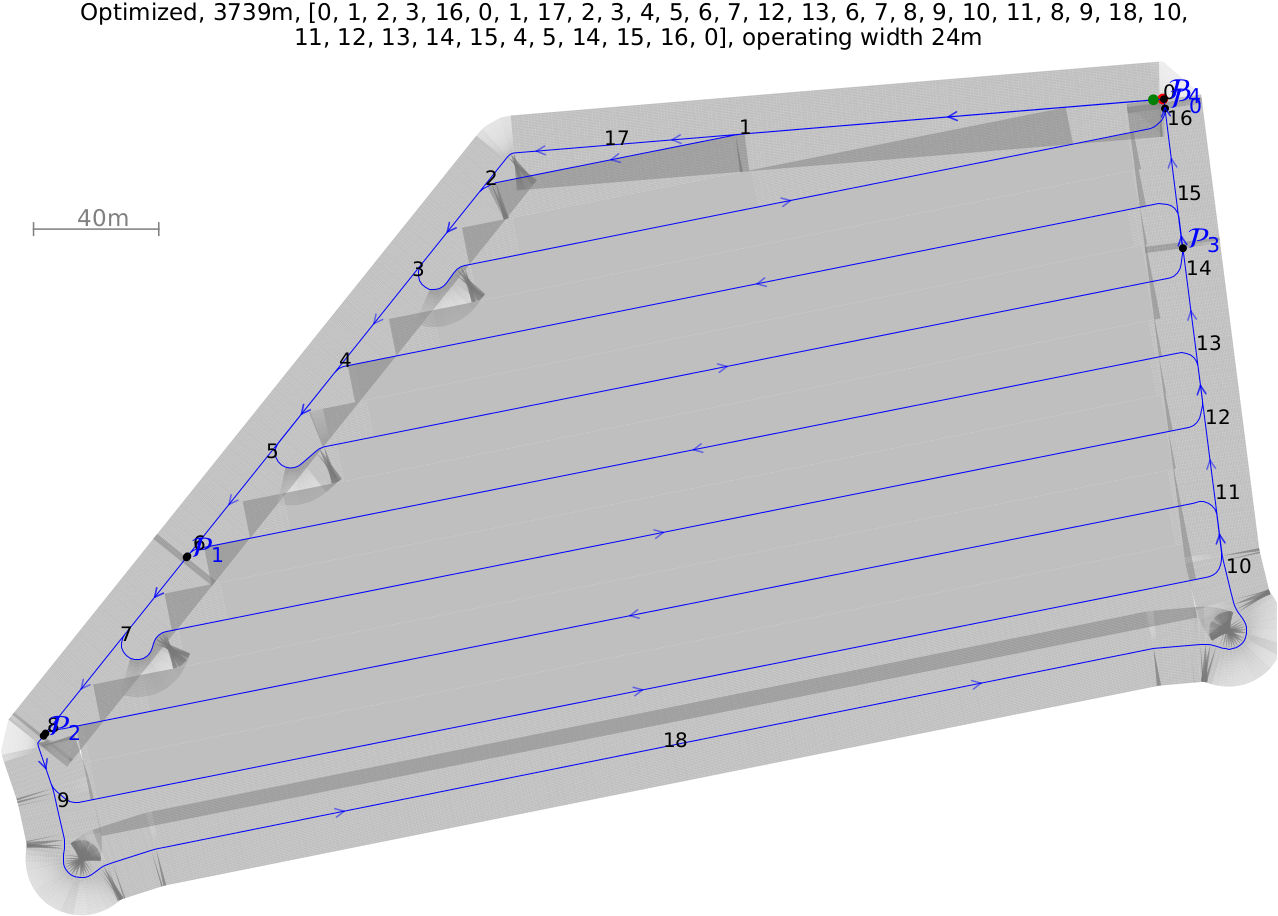}\\[-1pt]
\caption{$S_{\textsf{M}_2}^{2}$}
  \label{fig_f11_SO1}
\end{subfigure}
\begin{subfigure}[t]{.33\linewidth}
  \centering
  \includegraphics[width=.99\linewidth]{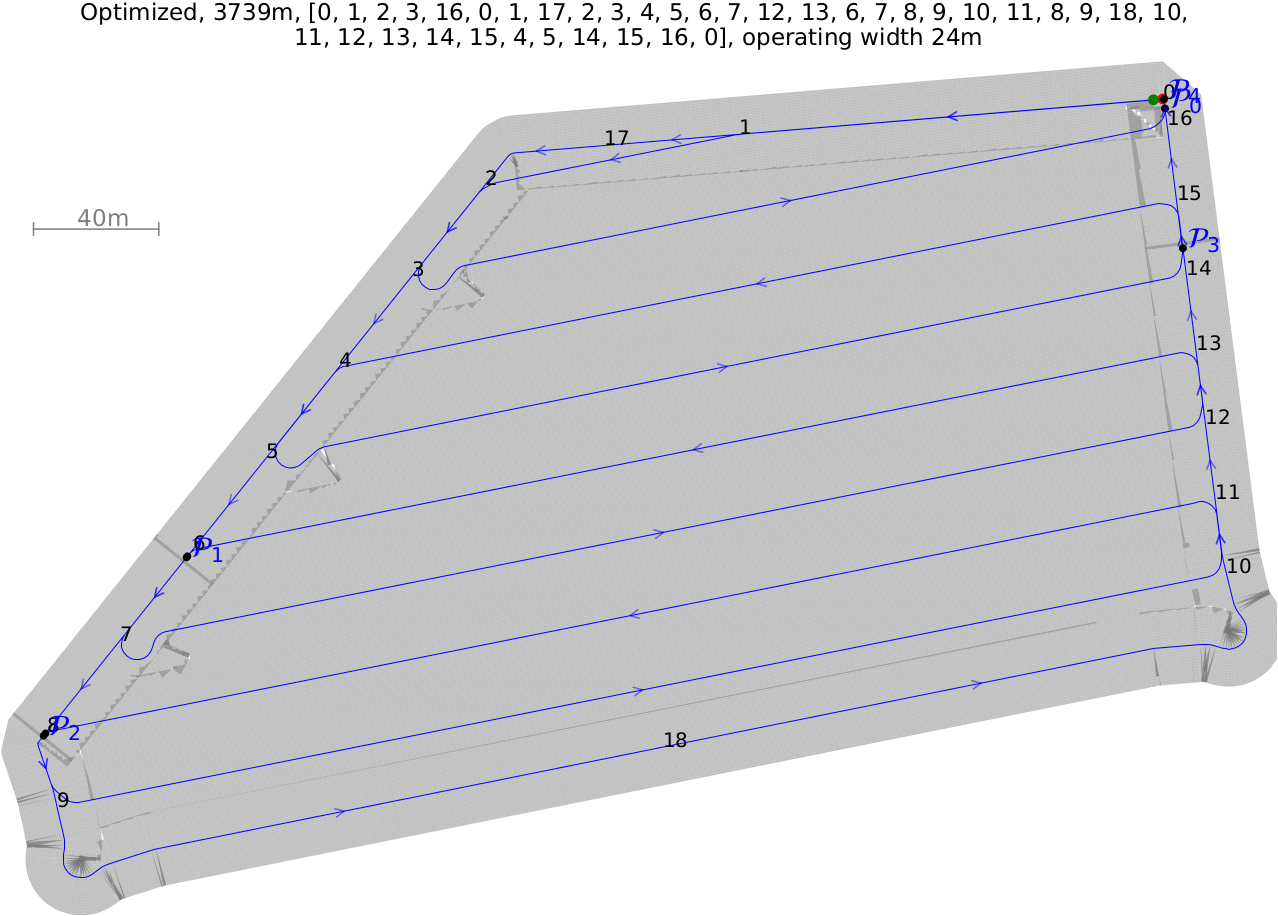}\\[-1pt]
\caption{$S_{\textsf{M}_2}^{48}$}
  \label{fig_f11_SO2}
\end{subfigure}
\caption{Field example 10: visual comparison of 6 different setups. See Table \ref{tab_s} for quantitative evaluation. }
\label{fig_f11}
\end{figure*}


\end{document}